\pdfoutput=1

\documentclass[11pt]{article}

\usepackage[final]{acl}

\usepackage{times}
\usepackage{latexsym}

\usepackage[T1]{fontenc}

\usepackage[utf8]{inputenc}

\usepackage{microtype}

\usepackage{inconsolata}

\usepackage{graphicx}

\title{Cross-Lingual Generalization and Compression:\\
From Language-Specific to Shared Neurons}

\author{
  Frederick Riemenschneider \and Anette Frank\\
  Department of Computational Linguistics\\
  Heidelberg University, Germany\\
  \texttt{\{riemenschneider|frank\}@cl.uni-heidelberg.de}
}

\usepackage{csquotes}
\usepackage{xspace}
\usepackage[noabbrev]{cleveref}
\usepackage{siunitx}
\usepackage{subcaption}
\usepackage{amssymb}
\usepackage{enumerate}
\usepackage{tabularx}
\usepackage{multirow}
\usepackage{booktabs} 
\usepackage{array}  
\usepackage{CJKutf8}
\usepackage{enumitem}
\usepackage{comment}

\newcommand{\bloom}{\textsc{BLOOM}\xspace}
\newcommand{\bloomm}{\textsc{BLOOM-560m}\xspace}
\newcommand{\bloomb}{\textsc{BLOOM-7b1}\xspace}

\newcommand{\xlmr}{\textsc{XLM-R}\xspace}
\newcommand{\mbert}{\textsc{mBERT}\xspace}
\newcommand{\onesec}{\textsc{OneSec}\xspace}
\newcommand{\wordnet}{\textsc{WordNet}\xspace}
\newcommand{\nllb}{\textsc{NLLB 1.3B}\xspace}
\newcommand{\lape}{\textsc{LAPE}\xspace}
\newcommand{\xglm}{\textsc{XGLM}\xspace}
\newcommand{\langdetect}{\textsc{langdetect}\xspace}
\newcommand{\oscar}{\textsc{OSCAR}\xspace}

\begin{document}

\maketitle
\begin{abstract}
 Multilingual language models (MLLMs) have demonstrated remarkable abilities to transfer knowledge across languages, despite being trained without explicit cross-lingual supervision. We analyze the parameter spaces of three MLLMs to study how their representations evolve during pre-training, observing patterns consistent with compression: models initially form language-specific representations, which gradually converge into cross-lingual abstractions as training progresses.
Through probing experiments, we observe a clear transition from uniform language identification capabilities across layers to more specialized layer functions.
For deeper analysis, we focus on neurons that encode distinct semantic concepts. By tracing their development during pre-training, we show how they gradually align across languages.
Notably, we identify specific neurons that emerge as increasingly reliable predictors for the same concepts across 
languages. 
This alignment manifests concretely in generation: 
once an MLLM exhibits cross-lingual generalization according to our measures, we can select concept-specific neurons identified from, e.g., Spanish text and manipulate them to guide token predictions. Remarkably, rather than generating Spanish text, the model produces semantically coherent English text. This demonstrates that cross-lingually aligned neurons encode generalized semantic representations, independent of the original language encoding.

\end{abstract}

\section{Introduction}
How do multilingual language models achieve cross-lingual generalization? This question has puzzled researchers for years--particularly since standard pre-training objectives do not explicitly encourage cross-lingual alignment.

Existing research has explored various potential explanations, ranging from linguistic similarity due to genetic or geographic relatedness \citep{lin-etal-2019-choosing,lauscher-etal-2020-zero} and word order properties \citep{dufter-schutze-2020-identifying,deshpande-etal-2022-bert} to architectural features like shared subwords \citep[][\textit{inter alia}]{pires-etal-2019-multilingual,wu-dredze-2019-beto}. However, researchers found conflicting evidence that challenges these explanations, especially regarding the role of shared subwords \citep[][\textit{inter alia}]{artetxe-etal-2020-cross,K2020Cross-Lingual}.

In this paper, we aim to derive explanations at a deeper level, by exploring the connection between cross-lingual generalization and signals of compression. We build on the theory that the limited capacity of language models forces them to discover efficient, shared representations encoded in specific neurons across languages, rather than maintaining separate, language-dependent encodings.

To gain deeper insights into the development of cross-lingual representations, we focus on the pre-training process itself rather than just analyzing the final model state. For our analysis, we use models from the \bloom family \citep{bigscience_workshop_2022} at different scales (\bloomm and \bloomb), which are state-of-the-art multilingual decoder models that provide open access to training checkpoints, though their number is limited. Additionally, we introduce our own decoder-only model where we collect checkpoints at much finer intervals and maintain precise control over the training data and languages, allowing us to build a detailed picture of how cross-lingual representations are shaped during training.

While much prior work has focused on zero-shot cross-lingual transfer performance, recent research has shown such transfer to be unreliable \citep{rajaee-monz-2024-analyzing}. 
We therefore take a more mechanistic approach and analyze the models themselves rather than their zero-shot transfer capabilities. We start our analysis by probing model-internal representations for language identity prediction,
to build initial intuitions, which reveals clear shifts in performance across pre-training checkpoints. We then examine how semantic concepts (like \texttt{house} or \texttt{earthquake}) are represented in \emph{individual neurons} across different languages, following the neuron analysis approach of \citet{suau2022selfcond}.

Our analysis reveals increasing cross-lingual alignment during pre-training, with specific neurons emerging as shared concept experts across languages. We complement these observations with an information-theoretic perspective on compression and demonstrate their practical implications through controlled text generation experiments, showing how MLLMs 
evolve from language-specific to generalized neural representations during the pre-training process.
To summarize, our contributions are: 
\begin{enumerate}[label=(\roman*.),noitemsep]
\item We provide empirical evidence for the compression hypothesis in MLLMs by tracking how representations evolve from language-specific to cross-lingual throughout training, using mechanistic interpretability methods and special probing tasks. 
\item To our knowledge, we are the first to analyze the development of cross-lingual semantic generalization during pre-training, by identifying specific neurons that encode the same concepts across different languages.  Our ana\-lysis reveals how semantic information concentrates in middle layers and evolves into generalized concept representations shared across languages at later training stages.
\item We demonstrate how our findings have a visible effect on text generation through controlled neuron manipulation experiments, illustrating that the model's internal representations encode shared conceptual knowledge beyond specific language boundaries.
\end{enumerate}
We release our MLLM
with comprehensive training checkpoints, along
with code and data.\footnote{\url{https://github.com/Heidelberg-NLP/cross-lingual-generalization}.}

\section{Related Work} 
\paragraph{Studying Cross-Lingual Generalization.} 
Since the early development of MLLMs, researchers have investigated their remarkable effectiveness, 
primarily through the lens of zero-shot cross-lingual transfer performance. 
Explanatory factors for the identified  cross-lingual generalization capabilities can roughly be divided into \emph{linguistic aspects}, such as genetic and geographic relatedness and word order \citep[][\textit{inter alia}]{lin-etal-2019-choosing,lauscher-etal-2020-zero,dufter-schutze-2020-identifying,deshpande-etal-2022-bert}, and \emph{architectural considerations} like model depth and number of parameters \citep[][\textit{inter alia}]{dufter-schutze-2020-identifying,K2020Cross-Lingual}. The role of \emph{lexical overlap between languages} has been a particular point of debate in the literature \citep[][\textit{inter alia}]{pires-etal-2019-multilingual,wu-dredze-2019-beto,artetxe-etal-2020-cross,dufter-schutze-2020-identifying,K2020Cross-Lingual}. For a comprehensive overview of these developments, we refer to \citet{philippy-etal-2023-towards}. 

\paragraph{Compression in Language Models.} 
The information bottleneck method, introduced by \citet{tishby99information}, provides a theoretical framework for analyzing information flow in neural networks. \citet{bottleneck2} apply this framework to deep learning, showing that neural networks must learn to efficiently represent task-relevant information while \enquote{forgetting} irrelevant input details. Building on these insights, \citet{voita-etal-2019-bottom} analyze how representations evolve bottom-up in Transformers. \citet{shwartz2017opening} identify two distinct phases in neural network training: an \emph{initial fitting phase}  followed by a \emph{compression phase}, the latter being causally linked to the network's generalization capabilities. 

In the multilingual context, the compression hypothesis has been acknowledged, but to date remains underexplored: \citet{chi-etal-2021-infoxlm} explicitly reference the information bottleneck method while deferring its investigation, and \citet{dufter-schutze-2020-identifying} observe that overparameterization may actually hinder multilingual performance. In our work, we systematically investigate  how compression manifests in multilingual models during pre-training, hypothesizing that restricted model capacity forces the development of shared cross-lingual representations, rather than maintaining separate language-specific ones.

\paragraph{Mechanistic Interpretability.} 
Mechanistic interpretability seeks to reverse engineer neural networks to understand their internal functioning. Even in work not focused on MLLMs, multilingual phenomena have emerged as peripheral findings: \citet{gurnee2023finding} identify neurons that respond to French texts through sparse probing, while \citet{bricken2023monosemanticity} discover Arabic script and Hebrew features via sparse autoencoders.

Current work has begun to explicitly address multilinguality:
\citet{wendler-etal-2024-llamas} show that \textsc{Llama 2} models \citep{touvron2023llama}, consistent with their English-dominated training data, process other languages using English as internal pivot.
Recently, research has developed specialized methods to identify language-specific neurons: 
\citet{tang-etal-2024-language} propose \lape (Language Activation Probability Entropy), while \citet{kojima-etal-2024-multilingual} build on \citeposs{suau2022selfcond} methodology, which we introduce below. However, these investigations focus solely on identifying neurons responsible for language-\emph{specific} processing, rather than examining, as we do, whether semantic concepts \emph{share} neural representations across languages.

Closest to our work is \citet{blevins-etal-2022-analyzing}, who analyze \xlmr \citep{conneau-etal-2020-unsupervised} checkpoints for its performance in linguistic tasks such as PoS tagging or dependency parsing across different languages. Yet, unlike their work, we examine decoder-only models and  investigate the pre-training process at a more fundamental level.

\section{Conceptualizing Cross-Lingual Generalization}

Most prior work studies MLLMs through zero-shot cross-lingual transfer. In this setting, \enquote{a model that is fine-tuned on one language can be applied to others without any further training} \citep{tunstall2022natural}. This approach has become the standard for evaluating multilingual models \citep{xtreme}.

Often, zero-shot cross-lingual transfer (often abbreviated to just \enquote{\emph{cross-lingual transfer}}) is treated as synonymous with \emph{cross-lingual generalization}. This conflation is problematic for two reasons. First, despite its name suggesting otherwise, in zero-shot cross-lingual transfer, models \emph{do} undergo fine-tuning, 
potentially obscuring more subtle phenomena \citep{papadimitriou-etal-2023-multilingual}. Second, these evaluations are vulnerable to dataset artifacts like word overlap and answer position bias \citep{rajaee-monz-2024-analyzing}, and may instead reflect surface-level patterns, rather than linguistic generalization.

We therefore argue for a clear distinction between \emph{zero-shot cross-lingual transfer} as a specific evaluation method and \emph{cross-lingual generalization} as the fundamental ability of models to form cross-lingual abstractions. Our work investigates the latter by analyzing internal representations directly, without any fine-tuning.
We hypothesize that \emph{cross-lingual generalization emerges through compression} during pre-training. We assume that once a model's capacity constraints prevent pure memorization, it develops more space-efficient representations by abstracting away language-specific features from the encoded content. Our experiments support this hypothesis: we observe the emergence of \emph{cross-lingual concept neurons} that respond to the same concepts across different languages.

\section{Model Details}

To study cross-lingual generalization during pre-training, we require access to model checkpoints throughout the pre-training process. We therefore focus our analysis on the \bloom family \citep{bigscience_workshop_2022}, the only recent collection of MLLMs to offer publicly available training checkpoints. Due to computational resource constraints, we conduct our most detailed analysis on \bloomm. However, we confirm our key findings on \bloomb, demonstrating that our results generalize to larger models.\footnote{Two checkpoints each from \bloomm and \bloomb were excluded from our analysis, as they appear to be corrupted (cf.~\url{https://huggingface.co/bigscience/bloom-560m-intermediate/discussions/2} and \url{https://huggingface.co/bigscience/bloom-7b1-intermediate/discussions}).}

In addition, to allow for a more fine-grained ana\-lysis of training dynamics than \bloom's checkpoint frequency allows, we pre-trained our own model based on the \xglm architecture \citep{lin-etal-2022-shot}, using a reduced dimension (\(\texttt{d\_model}=512\) instead of \(1024\)), resulting in approx.\ 257M parameters. We trained on 16 languages spanning diverse language families and scripts (Germanic, Italic, Bantoid, and Slavic), collecting checkpoints at powers of two and regular 5000-step intervals.

\begin{figure*}[t!]
    \centering
    \begin{subfigure}[t]{0.5\textwidth}
        \centering
        \includegraphics[width=\linewidth,trim=2cm 0cm 4cm 1cm,clip]{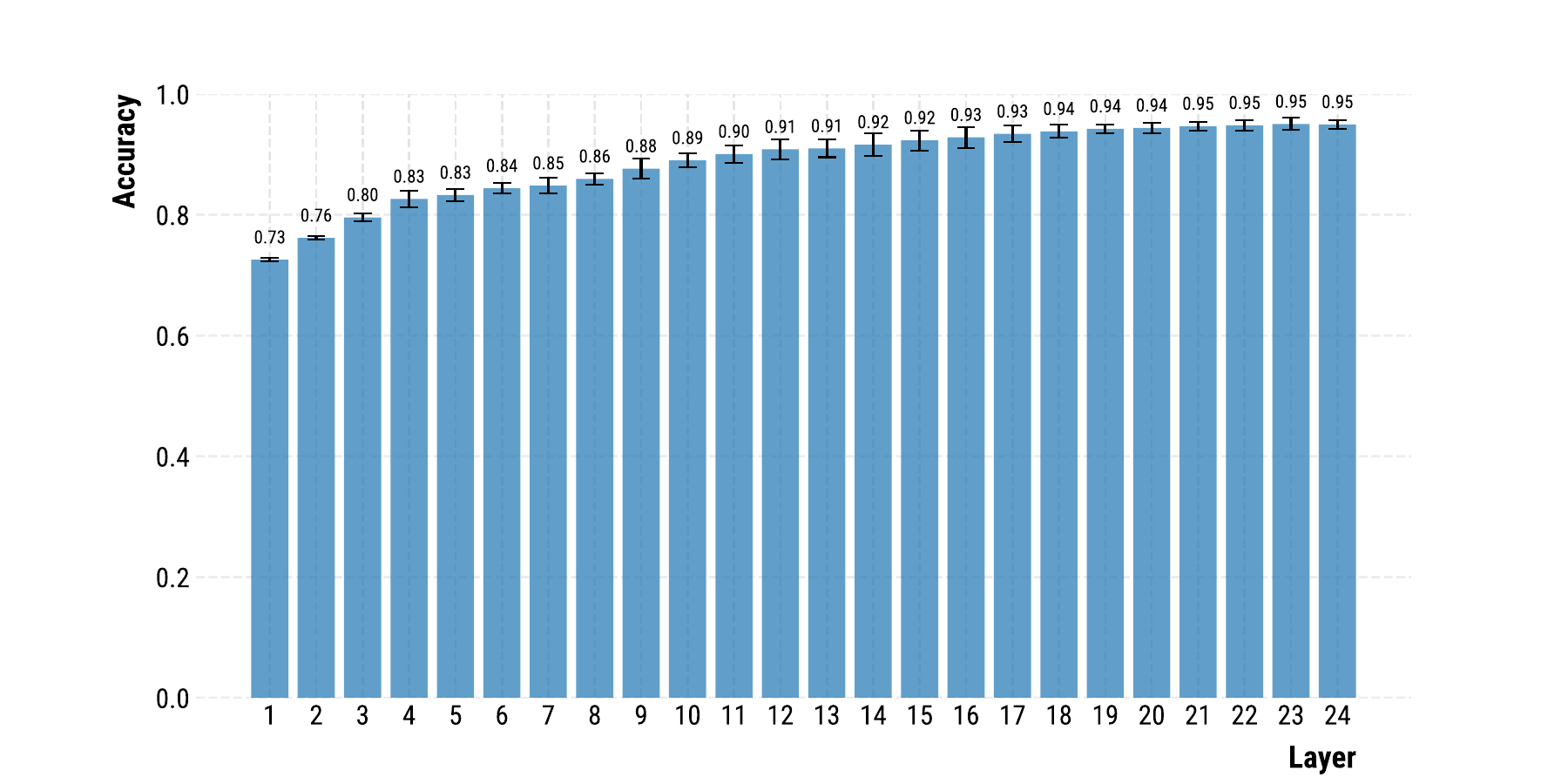}
        \caption{Early training stage (step \num{1000}).}
    \end{subfigure}%
    ~ 
    \begin{subfigure}[t]{0.5\textwidth}
        \centering
        \includegraphics[width=\linewidth,trim=2cm 0cm 4cm 1cm,clip]{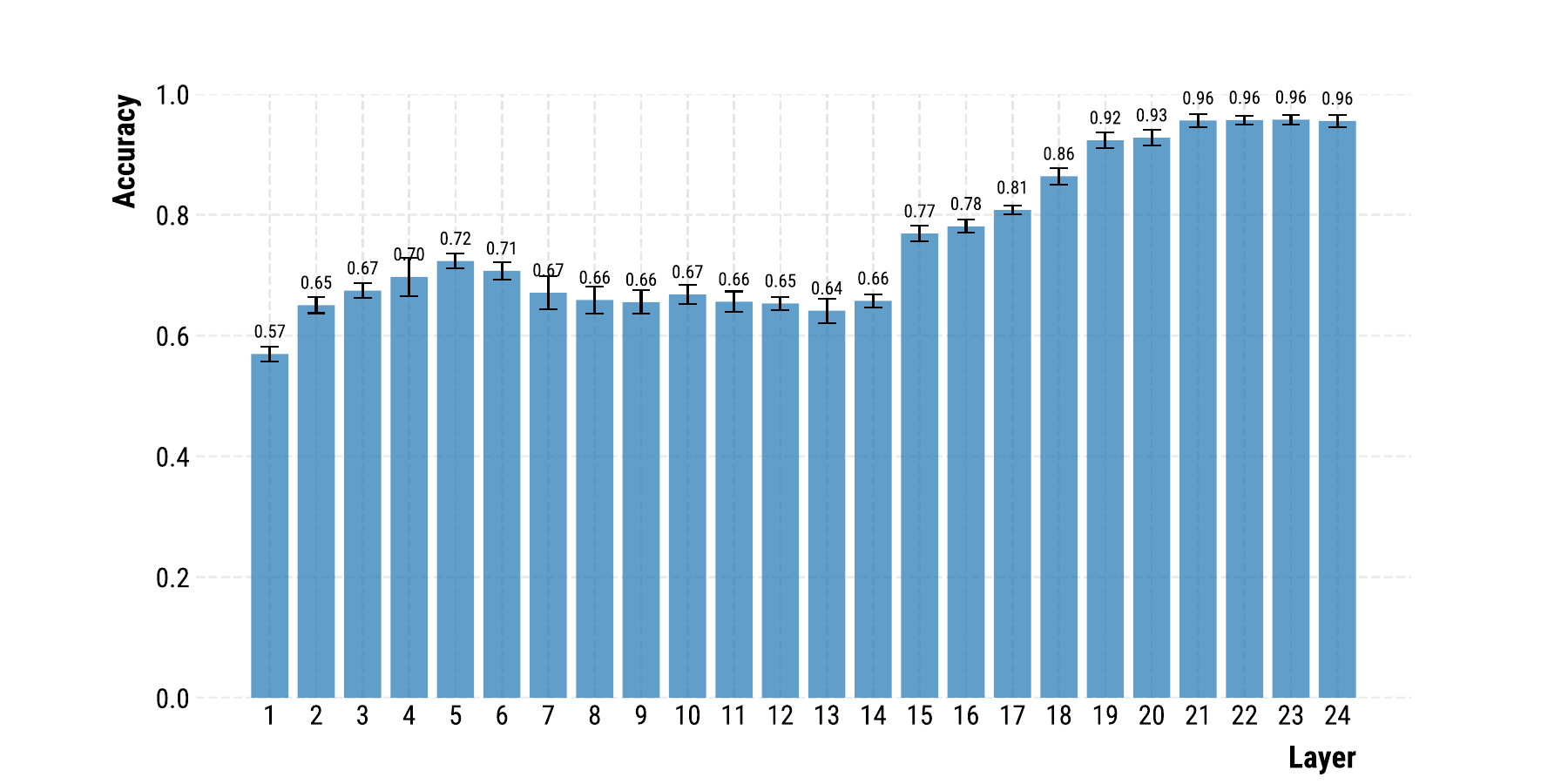}
        \caption{Late training stage (step \num{400000}).}
    \end{subfigure}
    \caption{Language identity probing classification accuracy across layers of the \bloomm model at different training stages. Higher accuracy indicates that language-specific information is more easily extractable from the hidden states at that layer. Error bars show standard deviation across three random seeds.}\vspace{-.3\baselineskip}
    \label{fig:probing_detailed}
\end{figure*}

\section{Probing Language Identity Across Layers and Checkpoints}
\label{probing}

\paragraph{Motivation.} To investigate the relationship between cross-lingual generalization and the model's representations, we examine to what extent language-specific information is encoded across layers and training stages, focusing on how the MLLM's
internal organization of languages develops during training. In a first step, we probe each model layer's ability to identify which language is being processed,  examining how language-specific information is distributed across the model's layers. This initial probing experiment serves as a foundation for understanding how language representations develop both through the model's layers and throughout its training process.

\paragraph{Experiment Setup.} For a given language \(l\), we sample \(\{s_0^l,s_1^l,...,s_n^l\}\) sentences from the \oscar corpus \citep{OrtizSuarezSagotRomary2019}. Each sentence \(s^l_i\) is tokenized into a sequence of tokens \([t_{i,0}^ l,t_{i,1}^l,...,t_{i,T_i-1}^l]\). For each tokenized sentence, the model \(\mathcal{M}\) produces hidden representations: \(h^l_{i,0}, h_{i, 1}^l,...,h_{i,T_i-1}^l = \mathcal{M}(t_{i,0}^l, t_{i,1}^l,...,t^l_{i,T_i-1}) \).

From these sequences of hidden states, we randomly sample one token position per sentence and extract the hidden representation at that position. For instance, for sentence \(s_i^l\), we might select position \(p_i^l\) to obtain \(h_{i, p_i^l}^l\). We then train a logistic regression classifier on these sampled hidden states, aiming to predict which language \(l\) the hidden state originated from. By analyzing classification performance across layers, we investigate how the representation of languages evolves throughout the MLLM's
architecture, and how languages are organized.  For implementation details see \Cref{app:probing}.
	
\paragraph{Results.} We present analysis results for \bloomm at pre-training steps \num{1000} and \num{400000} in \Cref{fig:probing_detailed}. At step \num{1000}, the model already demonstrates strong language identification capabilities, with a slight performance increase after the first layer followed by small, monotonic improvements across subsequent layers. At step \num{400000}, by contrast, we observe markedly different behavior: performance in earlier layers is substantially weaker, starting at 57\% accuracy in the first layer, increasing until layer 5, then declining until layer 14. From layer 15 onwards, performance recovers, eventually matching the levels observed in the earlier checkpoint. 

The precise layer-wise accuracy trajectory appears to be architecture-dependent, with decoder-only models showing this distinctive pattern (see results for \bloomb and our toy model, as well as comparisons with encoder-only models \xlmr \citep{conneau-etal-2020-unsupervised} and \mbert \citep{devlin-etal-2019-bert} in \Cref{app:probing}).
However, we observe a fundamental organization  that is shared across different model families: language-specific information diminishes in the middle layers, while the final layers maintain strong identification capabilities.

\begin{figure}
    \centering
    \includegraphics[width=\linewidth,clip,trim=2cm 0cm 2.5cm 0cm]{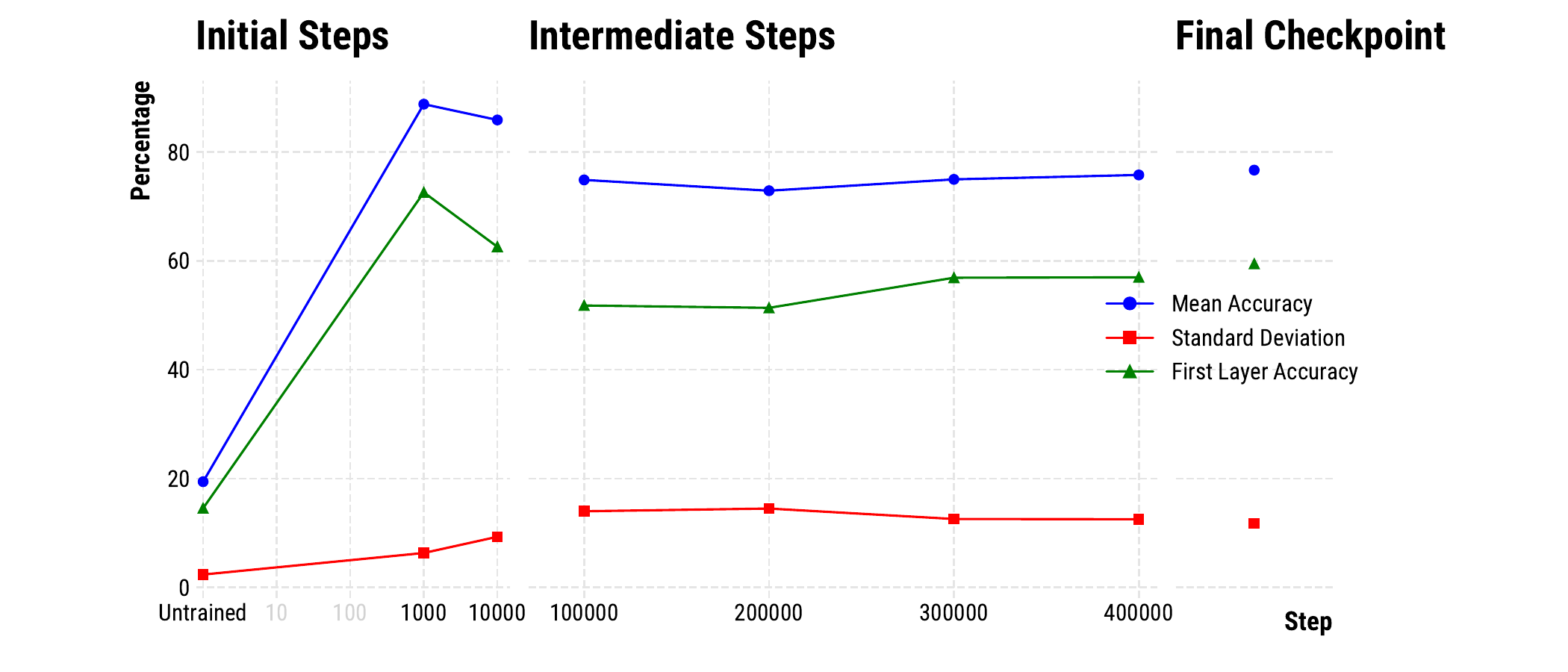}
    \caption{Language identification probing accuracy through\-out training of \bloomm. For each checkpoint we show: (1) mean accuracy across layers, (2) standard deviation across layers (indicating how much accuracy varies between layers), and (3) first layer accuracy, which exhibits the most significant changes during training. Results averaged over three random seeds.}
    \label{fig:full_training}\vspace{-.2\baselineskip}
\end{figure}

Complementing our detailed analysis of individual checkpoints in \Cref{fig:probing_detailed}, \Cref{fig:full_training} tracks three key statistics throughout training: the first layer accuracy, the mean probing accuracy averaged across all layers, and the corresponding standard deviation between layer-wise accuracies. During early training (steps \num{1000} to \num{10000}), we observe uniformly high language identification performance across layers, reflected in high mean accuracy and low between-layer variance. Beyond step \num{100000}, layer-averaged accuracy decreases (especially in the first layer), while standard deviation increases, indicating greater differentiation between layers.

These findings reveal a fundamental shift in how language information is processed throughout pre-training: the model initially develops strong language identification capabilities, but subsequently this ability diminishes. 
We hypothesize that different layers develop distinct functional roles during training: while final layers maintain high language identification accuracy necessary for next-token prediction,  middle layers develop representations that tend to be more language-agnostic. These observations can be argued to provide initial evidence for
a compression effect that manifests during pre-training, characterized by a shift from language-specific to more generalized representations.
	
In what follows, we build on this hypothesis and investigate whether cross-lingual generalization can be explained as a result of \emph{compression} that occurs in a multilingual model's representation space after an initial phase of memorization.

\section{Tracing Concepts in Neurons} %

\paragraph{Research Question.} We now examine how individual concepts are represented in MLLMs by tracing their evolution during pre-training, examining their relationships across languages.
We hypothesize that MLLMs first develop language-specific representations of concepts (e.g., separate encodings for \enquote{house}, \enquote{casa}, or \enquote{Haus}), which, as training progresses, merge into unified abstractions (e.g., the concept of a \texttt{dwelling}). We expect that these abstracted representations can be \enquote{projected} into language-specific instantiations during generation, offering a more space-efficient organization than separate representations for each language.

\paragraph{Experiment Setup.} To identify concept-specific \enquote{expert} neurons (e.g., those specialized in representing \texttt{dwelling}), we adopt the methodology of \citet{suau2022selfcond}. Each concept \(c\) in language \(l\) is represented by a dataset \(\{\mathbf{x}^{c, l}_i, b^{c, l}_i\}^N_{i=1}\), where the total of \(N_{c, l} = N^+_{c, l} + N^-_{c, l}\) sentences are divided into positive samples that contain the concept (\(b^{c, l}_i=1\)) and negative ones that do not 
(\(b^{c, l}_i=0\)). A neuron demonstrates 
\emph{expertise} for concept $c$ if it selectively activates for positive examples, while remaining inactive for negative ones. 

We evaluate how well an MLP neuron \(m\)'s activation pattern (excluding attention neurons) predicts concept \(c\) by analyzing the neuron's outputs \(\mathbf{z}^{c, l}_m = \{z^{c, l}_{m,i}\}^N_{i=1}\) 
in response to sentences \(\{\mathbf{x}^{c, l}_i\}\) and use these activations as \emph{concept prediction scores}, that indicate the presence of concept $c$ in a given input for language $l$.\footnote{A fixed-size sentence representation is obtained via max-pooling.} We measure a neuron's predictive power through Average Precision \(\text{AP}^{c, l}_m = \text{AP}(\mathbf{z}^{c, l}_m, \mathbf{b}^{c, l})\), which quantifies the area under the precision-recall curve.

\paragraph{Data.} We follow \citet{suau2022selfcond} in constructing our concept dataset from \onesec \citep{scarlini-etal-2019-just}, which provides Wikipedia sentences annotated with \wordnet senses \citep{miller-1994-wordnet}. From this corpus, we sample 200 \wordnet senses as target concepts, ensuring \(100 \leq N^+_{c,\text{eng}} \leq 1000\) positive samples and \(N^-_{c,\text{eng}} = 1000\)  negative samples per concept. We translate the resulting English dataset \(N_{\text{eng}}\) using the \nllb model \citep{costa2022no} to create parallel versions in the languages we use in our analysis.  Details on the dataset construction process are given in \Cref{app:data}.

\paragraph{General Neuron Alignment.}

\begin{figure*}[t!]
    \centering
    \begin{subfigure}[t]{0.5\textwidth}
        \centering
        \includegraphics[width=\linewidth,trim=1cm 0cm 0cm 0cm,clip]{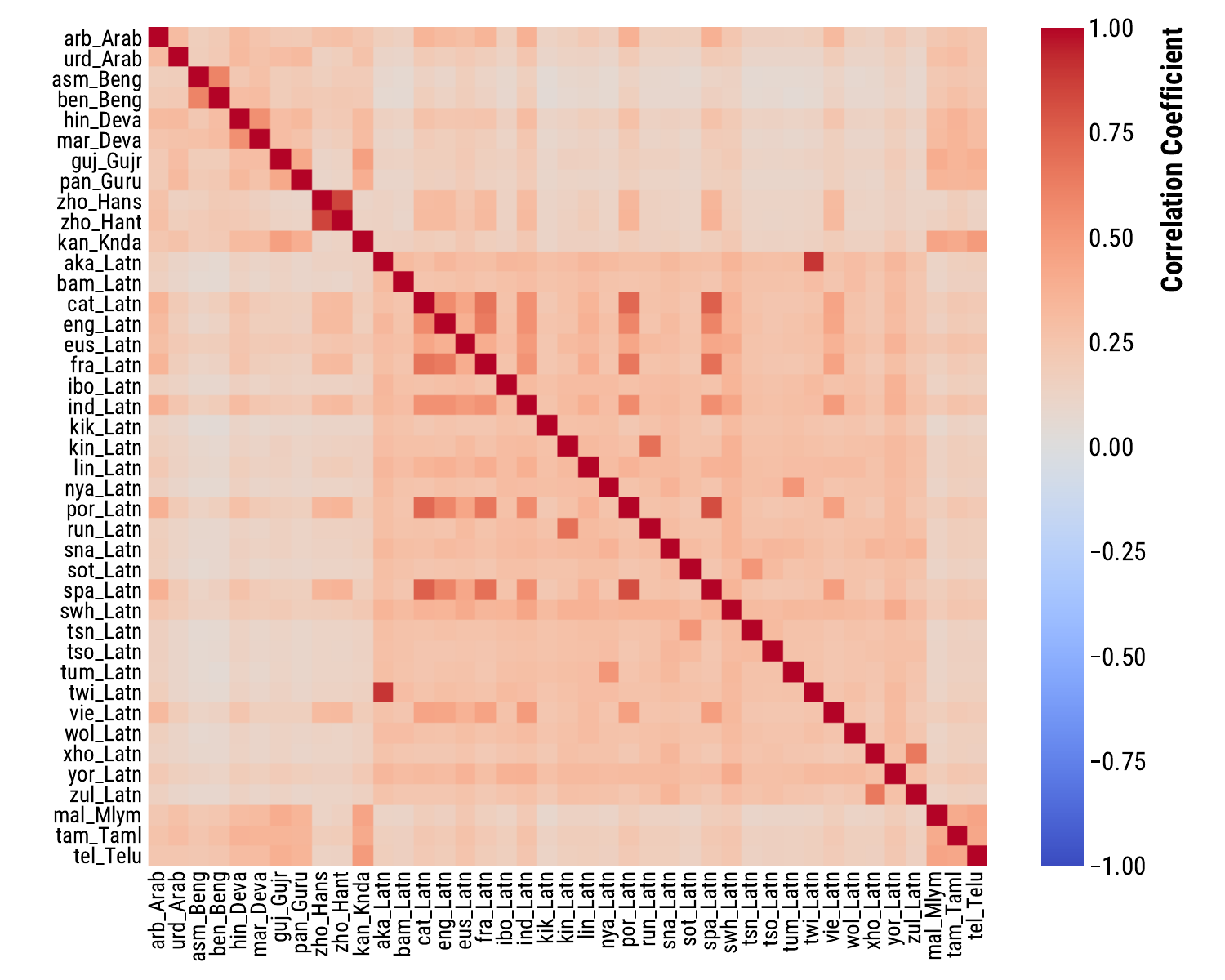}
        \caption{Early training stage (step \num{1000}).}
        \label{subfig:neuron_1000}
    \end{subfigure}%
    ~ 
    \begin{subfigure}[t]{0.5\textwidth}
        \centering
        \includegraphics[width=\linewidth,trim=1cm 0cm 0cm 0cm,clip]{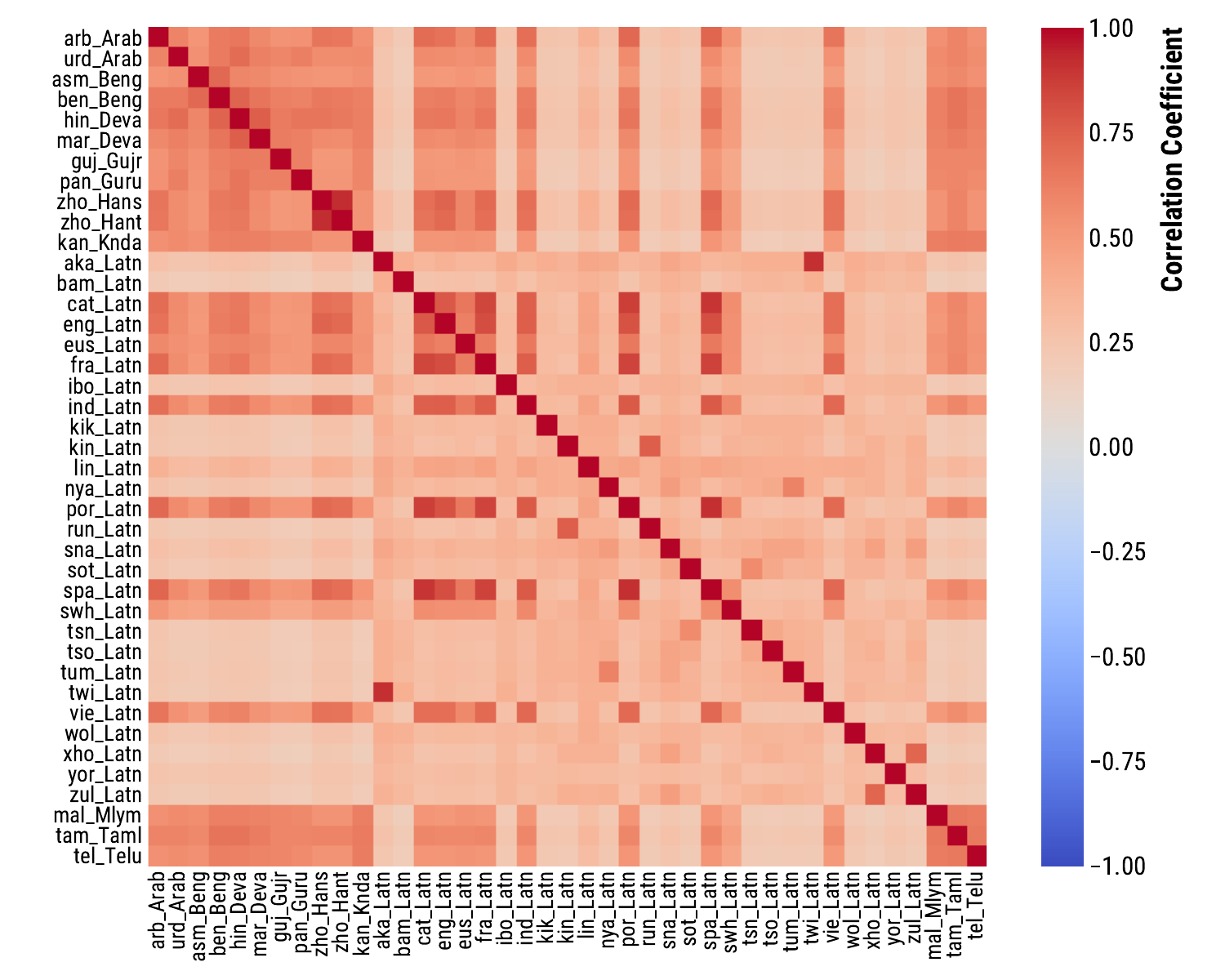}
        \caption{Late training stage (step \num{400000}).}
        \label{subfig:neuron_400000}
    \end{subfigure}
    \caption{Expert neuron alignment across languages in \bloomm at different training stages, measured by Pearson correlation coefficients averaged across concepts using Fisher's Z transformation.}
    \label{fig:neuron_alignment}
    \vspace{-.6\baselineskip}
\end{figure*}

\begin{figure}
    \centering
    \includegraphics[width=\linewidth,trim=1.25cm 0cm 1.25cm .6cm,clip]{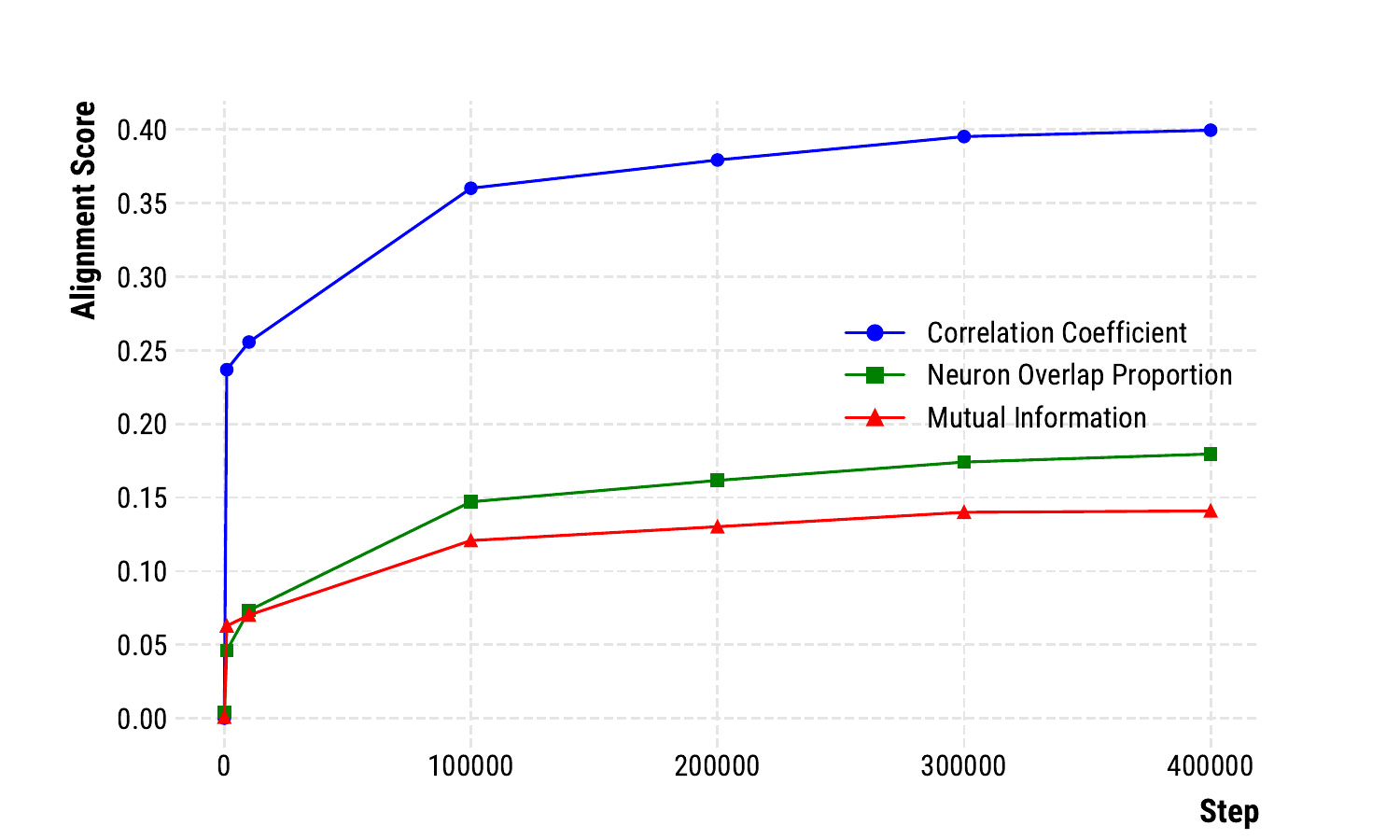} 
    \caption{(1) Correlation Coefficient, (2) Neuron Overlap Proportion of top 500 neurons, and (3) MI throughout training, averaged across concepts and languages for \bloomm.}\vspace{-.6\baselineskip}
    \label{fig:neuron_complete_training}
\end{figure}

Using our multilingual corpus derived from \onesec and the concept prediction score introduced above, we compute, 
for each language \(l\) and concept \(c\), an \emph{expert score vector} \(\mathbf{e}^{c, l} \in \mathbb{R}^M\), where \(M\) is the number of neurons and each element \(e^{c, l}_m = \text{AP}^{c, l}_m\) represents the expertise of neuron \(m\) for concept \(c\) in language \(l\). To investigate whether \emph{the same neurons} specialize in representing the same concepts across languages, we analyze the \emph{cross-lingual alignment} of these expert scores. Specifically, we compute the Pearson correlation coefficient between expert score vectors \(\mathbf{e}^{c, l_1}\) and \(\mathbf{e}^{c, l_2}\) for each concept across different language pairs \((l_1, l_2)\).

Given the large number of pairwise correlations across concepts and languages, we need a way to summarize these results concisely. We therefore apply Fisher's Z transformation to the correlation coefficients, compute their average, and transform the result back. We emphasize that this averaged score cannot be interpreted as a statistical correlation, but it still serves as a meaningful indicator of the degree of neuron alignment between languages.

The resulting matrices for \bloomm at training steps \num{1000} and \num{400000} are shown in \Cref{fig:neuron_alignment}, with values averaged across all concepts. \Cref{fig:neuron_complete_training} provides a view of this alignment throughout the training process, showing the averaged scores across both concepts and language pairs. For additional matrices and results for \bloomb and our toy model see \Cref{app:results}.

Early in training, the alignment between languages is relatively weak, but it strengthens substantially by step \num{400000}. The correlation matrix reveals dependencies that are partially attributable to script families. We observe small but distinct clusters of related languages sharing the same script (e.g., Assamese and Bengali, Hindi and Marathi), and a broader positive alignment across languages using the Latin alphabet. Most notably, there appears to be a strong distinction between Latin-script and non-Latin-script languages, though this pattern is not absolute. The Dravidian languages (Kannada, Malayalam, Tamil, Telugu) represent a case of a language family exhibiting high similarities despite using distinct scripts, reinforcing previous findings that subword overlap alone cannot explain cross-lingual generalization. A deeper analysis of these relationships remains for future investigation.

\paragraph{Information-Theoretic Perspective.} Beyond examining correlations across neurons, we adopt an information-theoretic approach by analyzing the Mutual Information (MI) between neural representations across languages. MI quantifies how much knowledge of a concept's representation in one language informs its representation in another language. Specifically, MI measures \emph{compression efficiency} by indicating to what degree a concept's representation in one language is redundant, and thus predictable, given its representation in another language. We compute MI for continuous data using entropy estimation based on \(k\)-nearest neighbors distances, following the methods of \citet{mi1} and \citet{mi2}, as implemented in \texttt{scikit-learn} \citep{scikit-learn}. The evolution of MI (\Cref{fig:neuron_complete_training}) closely mirrors the correlation-derived alignment scores, reinforcing our findings through an information-theoretic lens.

\paragraph{Neuron Overlap.} Finally, we analyze the concrete overlap between the most concept-selective neurons across languages. For each concept \(c\) and language \(l\), we identify the set  \(S^{c, l}\) of the top \(k\) neurons with the highest expertise scores \(\mathbf{e}^{c,l}\). We quantify the cross-lingual overlap between languages \(l_1\) and \(l_2\) using the overlap proportion \(O^c_{l_1, l_2} = \frac{|S^{c,l_1} \cap S^{c,l_2}|}{k}\). This directly measures the degree of neuron sharing between languages, suggesting compression, as shared neurons indicate a more compact representation of concepts.

\Cref{fig:neuron_complete_training} shows the resulting overlap for \(k = 500\), averaged across languages and concepts.
The evolution of the overlap aligns with both the cor\-rela\-tion\--derived and mutual information measures. Remarkably, among the more than 200 million MLP parameters analyzed, we find that approx.\ \(\frac{1}{6}\) of the top 500 concept-selective neurons are shared between any pair of languages. This substantial overlap, despite the model's vast capacity, suggests significant cross-lingual representation sharing.

\section{Revisiting Layer Distributions}

As shown in \Cref{probing}, early layers partially lose their language identity information during pre-training.
We now return to this observation 
and examine how it relates to the distribution of concepts across a model's layers. Specifically, we investigate where concept-specific neurons are located, and how their distribution evolves during training.

\paragraph{Layer-Wise Distribution of Expert Neurons.} 
First, we explore where the previously identified top \(k\) expert neurons are located across layers.
By examining the layer distribution of these neurons, averaged across all languages and concepts, we obtain a language-agnostic view of where concept-specific information is concentrated in the model.

\begin{figure}
    \centering
    \includegraphics[width=\linewidth]{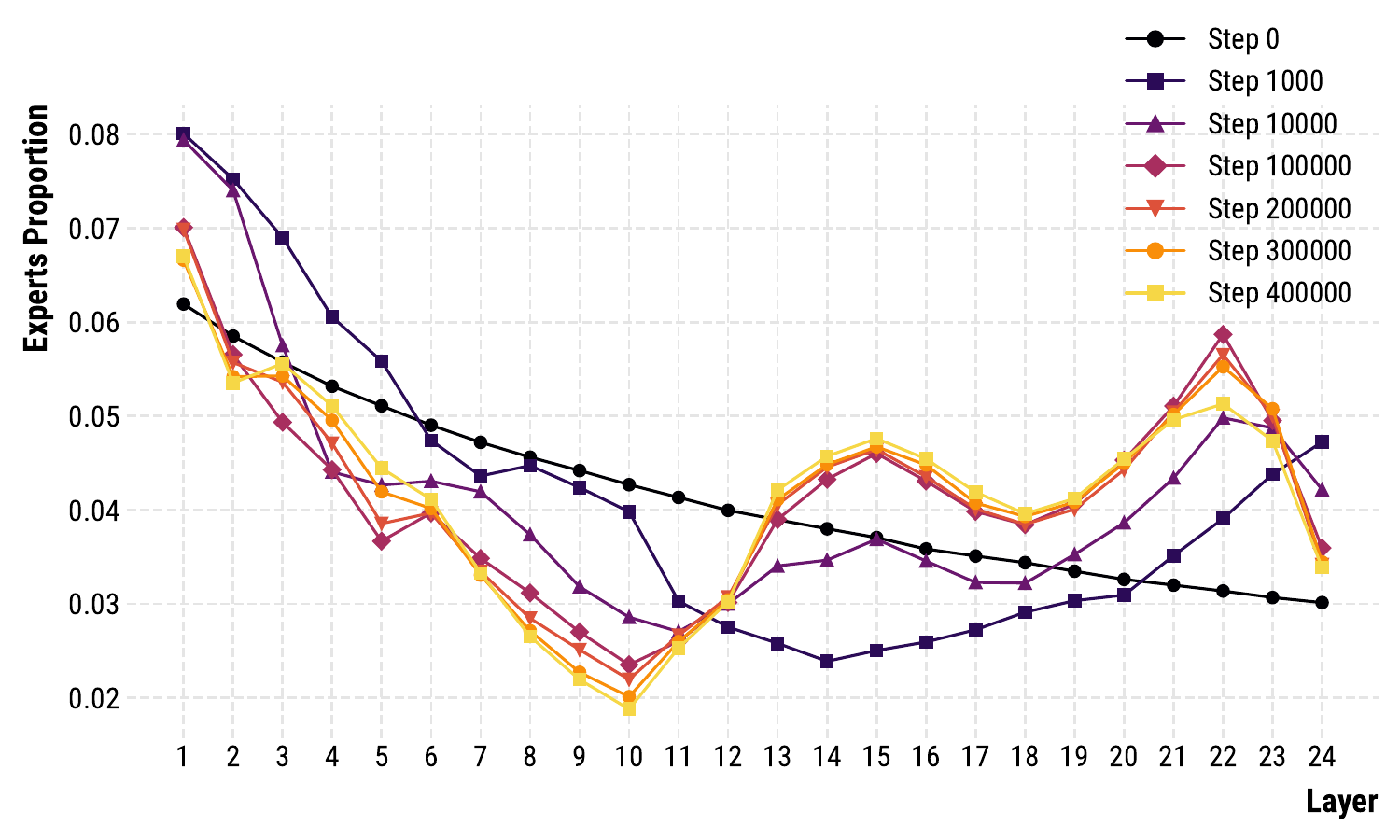}
    \caption{Layer-wise distribution of \textsc{BLOOM-560m}'s top 500 expert neurons, averaged across languages and concepts.}
    \label{fig:layer_distribution_comparison}%
\end{figure}

Our analysis in \Cref{fig:layer_distribution_comparison} reveals how the concentration of concept information across layers evolves throughout training.
In the randomly initialized model, the first layer contains the highest concentration of expert neurons. This is 
intuitive, 
as the untrained model can only use surface-level word overlap to \enquote{identify} concepts. This first-layer dominance intensifies during early training (step \num{1000}), suggesting that the model initially relies heavily on these lexical cues.

At step \num{100000} we observe a fundamental reordering of concept information across layers, which stabilizes and shows only marginal changes in later checkpoints. This new distribution reveals three distinct regions: After initial concentration in the first layers, there is a notable drop reaching its lowest point at layer 10.  This same layer marks the beginning of the first of two concentration peaks.  The final layer (24) shows a particularly low proportion, likely due to its role in token generation.

\paragraph{Layer-Wise Cross-Lingual Semantic Overlap.}
Building on these insights, we now analyze the 
\emph{cross-lingual alignment of concept representations for different layers}. 
For each layer \(\ell\), we compute the average pairwise overlap between  top \(k\) expert neurons by comparing the sets \(S^{c,l}_\ell\) across languages, measuring the proportion of shared neurons between languages \(l_1\) and \(l_2\) as \(\frac{|S^{c,l_1}_\ell \cap S^{c,l_2}_\ell|}{k}\). 

\begin{figure}
    \centering
    \includegraphics[width=\linewidth]{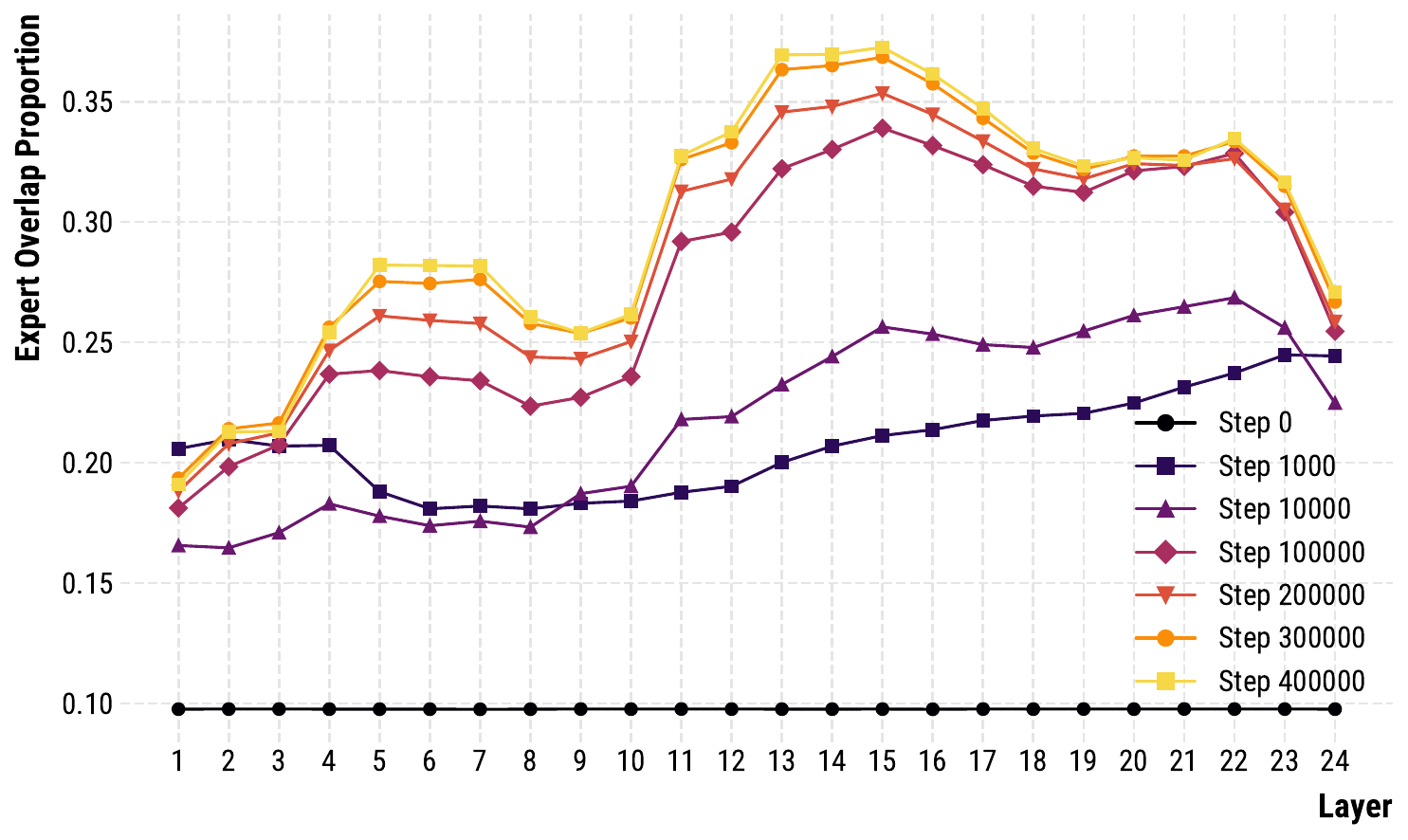}
    \caption{Cross-lingual overlap of \textsc{BLOOM-560m}'s top 500 expert neurons per layer, showing the averaged proportion of shared neurons between language pairs.}
    \label{fig:semantics-layers}\vspace{-.3\baselineskip}
\end{figure}

The results in \Cref{fig:semantics-layers} complement our previous findings (\Cref{fig:layer_distribution_comparison}). While the early layers showed high concept specificity, likely due to subword overlap, this does not lead to strong cross-lingual alignment. Instead, substantial cross-lingual overlap develops in the middle layers (10-17), particularly in later checkpoints, suggesting that genuine semantic generalization occurs in this region. This indicates that while subword similarity provides a useful initial bias for the model, actual cross-lingual semantic representations emerge in the middle layers. The increasing alignment in later checkpoints supports the compression hypothesis, suggesting that the model learns to abstract away from language-specific features. The decrease in overlap in the final layers aligns with their specialization for language-specific token generation. By comparing \Cref{fig:semantics-layers} and \Cref{fig:layer_distribution_comparison}, we can disentangle the effect of subword overlap from true cross-lingual generalization.
We confirm the same trends for \bloomb in \Cref{fig:layer_distribution_comparison_big,fig:semantics-layers_big}.

\begin{table*}[htb!]
\centering
\resizebox{0.85\linewidth}{!}{
\begin{tabular*}{\linewidth}{lp{2cm}p{2cm}p{8cm}}
\hline
\textbf{Checkpoint} & \textbf{Concept} & \textbf{Language} & \textbf{Generation} \\ \hline
\multirow{12}{*}{\num{10000}} 
    & \multirow{6}{*}{\texttt{earthquake}} 
        & Spanish & Posteriormente se quem se hizo sentir en el segundo momento una intensidad máxima de unos 40 minutos y la presencia de varios volcanes. [...] \\ \cline{3-4}
    &  & Simplified Chinese  & \begin{CJK*}{UTF8}{gbsn} 去年,大连市高岭土场镇发现30余处安全隐患。经市安全气象台和地质灾害防御站队员检查,发现大量高空存在安全隐患, \end{CJK*} [...] \\ \cline{3-4}\cline{2-4}
    & \multirow{4}{*}{\texttt{joy}} 
        & Spanish & Por todo lo que he leído sobre este nuevo reto, me ha encantado y he querido brotar las historias [...]
 \\ \cline{3-4}
    &  & Simplified Chinese  & \begin{CJK*}{UTF8}{gbsn}
        你越长大越幸福,幸福带给你的就是一生的幸福。你越长大越幸福,幸福带给你的就是一生的幸福。\end{CJK*} [...]  \\ \cline{3-4}
 \hline

 \multirow{12}{*}{\num{400000}} 
    & \multirow{6}{*}{\texttt{earthquake}} 
        & Spanish & Strong earthquakes occurred in Japan on Saturday. Five large earthquakes occurred in central Japan on Saturday, and the epicities affected areas [...] \\ \cline{3-4}
    &  & Simplified Chinese  & There is no obvious risk to the city and infrastructure in the past 12 hours. Numerary records for GTC were occurring at 8.4 degrees (58.8, 18.3) [...]  \\ \cline{3-4}\cline{2-4}
    & \multirow{6}{*}{\texttt{joy}} 
        & Spanish & The queer, introvert, positive and wonderful ever. In a happy, happy way for you. Embrace this away from the point of. Simple joy to everybody. [...] \\ \cline{3-4}
    &  & Simplified Chinese  & Been on the monitor for an hour now, absolutely amazing work. The photo imagery, the quality of the work and the POCs are over an hour and we are pleased with our work. [...] \\ \cline{3-4}
 \hline

\end{tabular*}
}
\caption{Example text generations from \bloomm when activating top 500 expert neurons derived from Spanish and Simplified Chinese concepts, shown at training checkpoints \num{10000} and \num{400000}.}\vspace{-.25\baselineskip}
\label{tab:generation}
\end{table*}

\section{Steering Text Generation}

Until now, we have examined probing performance and neuron behavior to understand neuron alignment  in MLLMs. We now investigate whether these findings are reflected in \textsc{BLOOM-560m}'s text generation capabilities. To test the cross-lingual semantic properties of concept-specific expert neurons, we adapt the neuron manipulation technique from \citet{suau2022selfcond}. For a given concept (e.g., \texttt{earthquake}), we identify its top 500 expert neurons using data from one language (e.g., Spanish) and manipulate their activations. Specifically, we compute the median activation value of these neurons across all samples containing the concept (\(b_i^{c,l} = 1\)), and set their activations to these values.
We then generate text by prompting the model with only a beginning-of-sequence token, using nucleus sampling (\(p=0.9\)) and temperature (\(t=0.8\)) across 100 random seeds. Full details are in \Cref{app:generation}.

Setting neurons to their median values biases the model's representations toward the target concept. This allows us to examine whether concept-specific neurons, identified in one language, encode semantic information that generalizes across languages (or whether such neurons remain language-specific, such that neurons derived from the concept \texttt{earthquake} in Spanish texts lead to earthquake-related content in Spanish). 
Importantly, our manipulation is limited to modifying expert neurons. We provide neither language-specific nor concept-related tokens, thus giving the model freedom in choosing language and content of its generations.

\begin{figure*}[htb!]
    \centering
    \begin{subfigure}[t]{0.475\textwidth}
        \centering
        \includegraphics[width=\linewidth,trim=0cm 0cm 0cm 3.6cm,clip]{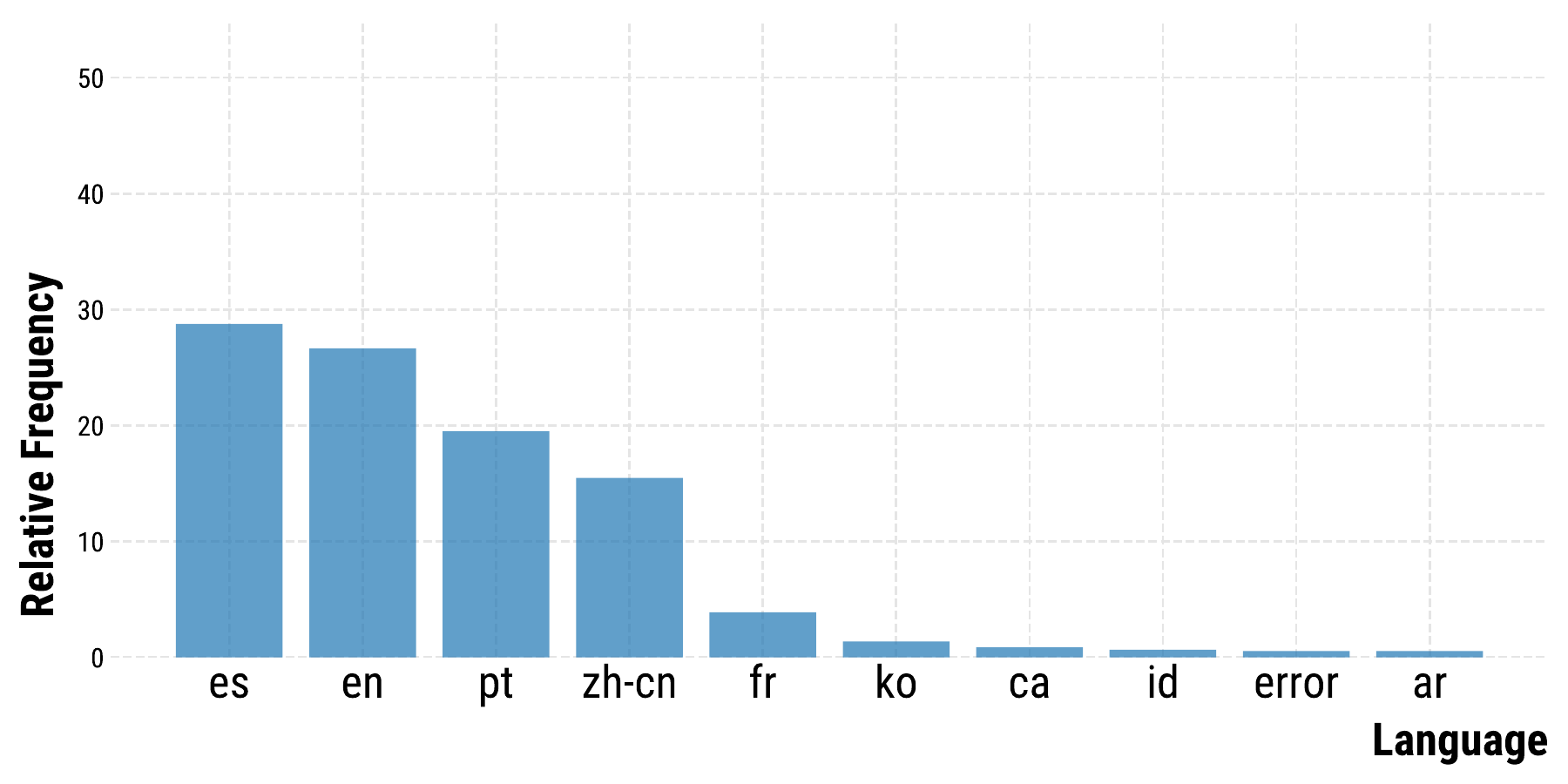}
        \caption{Early training stage (step \num{10000}).}
    \end{subfigure}
    ~
    \begin{subfigure}[t]{0.475\textwidth}
        \centering
        \includegraphics[width=\linewidth,trim=0cm 0cm 0cm 3.6cm,clip]{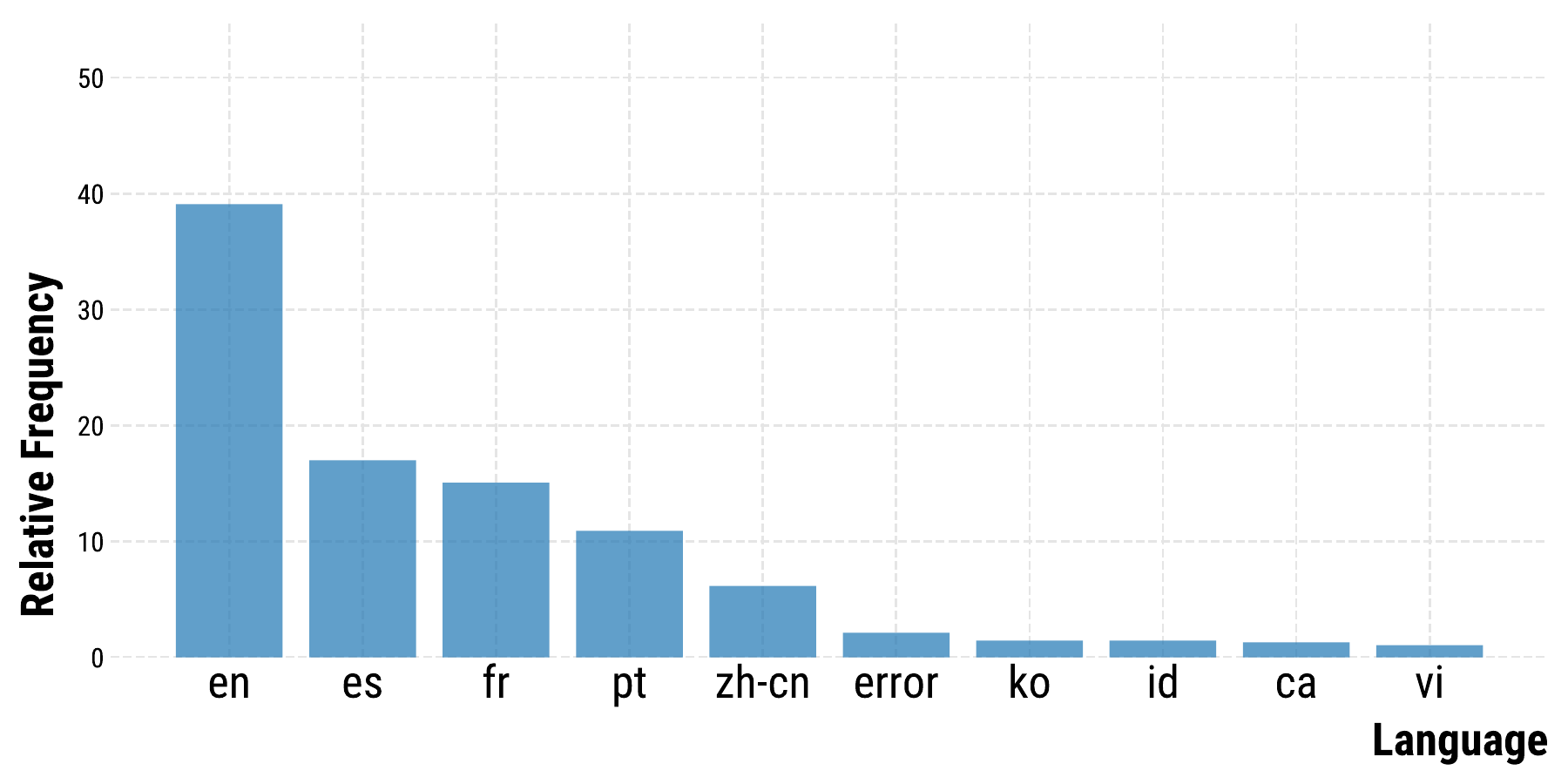}
        \caption{Late training stage (step \num{400000}).}
    \end{subfigure}
    \caption{Relative frequency distribution of the top 10 detected languages when manipulating neurons derived from Spanish data, as classified using \langdetect.}
    \label{fig:langdetect-detailed}\vspace{-.25\baselineskip}
\end{figure*}

Example generations are shown in \Cref{tab:generation}. Initially, the model produces incoherent text with excessive punctuation. By step \num{10000}, it generates concept-relevant text in the language from which the expert neurons were derived (e.g., Spanish text about earthquakes). However, at step \num{400000}, while the generated text remains concept-relevant, most generations are in English, even though the manipulated expert neurons were determined using exclusively non-English data.

To quantify these observations, we analyze the language of the generated text using \langdetect \citep{nakatani2010langdetect}. We classify 100 generations per checkpoint for all 200 concepts. For this analysis, we focus on neurons derived from Spanish data. The language distributions for steps \num{10000} and \num{400000} are presented in \Cref{fig:langdetect-detailed}. We show distributions across all steps alongside results for neurons derived from Chinese and Swahili data in \Cref{app:generation}.

At step \num{1000}, the model is too underdeveloped to generate meaningful text, producing mainly punctuation marks that \langdetect fails to classify. By step \num{10000}, language-specific representations become most prominent--the model primarily generates Spanish text, though with a substantial presence of English. Interestingly, we also observe significant Portuguese generation, likely due to its 
proximity to Spanish. The notable presence of Chinese text can be attributed to its prominence in \textsc{BLOOM}'s pre-training corpus, where it represents the second most common language after English.

This shift from generating text in the concept's source language to producing content in other
languages supports our core hypothesis about the model's learning trajectory: Early training builds language-specific representations that gradually 
transform into compressed cross-lingual representations.
Our generation experiments present direct evidence of this  generalization effect in MLLMs.

While cross-lingual generalization succeeds in our experiments, its nature raises important questions. Concept knowledge successfully transfers across languages, but this transfer is biased toward high-resource languages: 
We observe that the model tends to express concepts in English and Chinese, regardless of the language from which these concepts were learned. This bias is particularly pronounced for neurons derived from low-resource languages like Swahili (\Cref{fig:langdetect-detailed-swahili-full}), which never generate in their source language. We also observe spillover within language families, such as Portuguese generation from Spanish-derived neurons. This suggests that the model uses shared neurons to form a common understanding of concepts that can be accessed across languages. 
However, the key question that remains is 
whether models can reliably draw upon 
such shared representations when  they generate text
in specific languages--especially those underrepresented in the training data.

\section{Conclusion}
We investigate cross-lingual generalization from a compression perspective, complementing prior and concurrent work by analyzing the pre-training process of MLLMs. 
Our linear probing experiments reveal a decrease in language identification performance in certain layers during pre-training, pointing to changes in how the model utilizes its parameter space. By identifying and comparing expert neurons across languages, we demonstrate that multilingual models progressively align representations across languages, ultimately sharing a substantial portion of expert neurons.

Our analysis of expert neuron distributions reveals a systematic processing pattern: The model combines 
token-level features from 
early layers with abstracted semantic content in middle layers. Notably, the proportion of shared neurons increases significantly in middle layers, indicating this is where 
semantic generalization primarily occurs.
Generation experiments provide behavioral evidence of this phenomenon, 
showing the evolution from language-specific to abstracted concepts, as demonstrated by English generation from Spanish-derived concept neurons.

Future work could build on our insights to improve multilingual models. While we focused on shared representations, examining where languages maintain distinct encodings could provide supplementary understanding. Beyond obvious language-specific elements, culturally embedded concepts may require protection from the high-resource language bias we uncover. Our research offers insights for developing models that appropriately balance cross-lingual generalization with the preservation of linguistic and cultural diversity.

\section*{Limitations}

Our analysis spans multiple model scales (from our small toy model to \bloomb) but does not include the largest MLLMs due to the computational demands of calculating expert neuron scores across multiple languages, concepts, and training checkpoints. Nevertheless, the observed trends appear consistent and suggest broader applicability.

We analyze the \bloom family, which is currently the only state-of-the-art MLLM family with publicly available checkpoints. However, in both \bloomm and \bloomb, some checkpoints appear to be corrupted and were excluded from our analysis. To validate our findings despite these limitations, we conduct parallel experiments with our custom toy model.

Our analysis focuses specifically on individual semantic concepts, leaving other phenomena for future work: relationships between concepts (e.g., hierarchical categories or attribute sharing), syntactic phenomena shared across languages (such as agreement and word order), and specific patterns between individual language pairs.

\section*{Ethics Statement}
We do not foresee immediate ethical concerns for our research, as we primarily conduct analytical studies. \bloom is a considerably diverse language model family with a relatively high number of underrepresented languages. While we demonstrate biases toward high-resource languages in MLLMs, potentially disadvantaging speakers of lower-resourced languages, our analysis aims to make these biases transparent. Our toy model, though potentially inheriting biases from the \textsc{mC4} corpus, serves exclusively for controlled observation of MLLMs and has minimal dual-use potential.

\bibliography{anthology,custom}

\appendix
\clearpage

\section{Pre-training of Our Toy Model}
\label{app:pretraining}

We randomly initialize a model from the \textsc{XGLM-564M}  architecture \citep{lin-etal-2022-shot} and change \texttt{d\_model} from \(1024\) to \(512\), resulting in a model size of approximately 257M parameters (configuration details in \Cref{tab:config}). We pre-train this model using the \texttt{Transformers} library \citep{wolf-etal-2020-transformers} with default \texttt{Trainer} parameters and a batch size of \(32\) for \(2^{17}\) (= 131072) steps on the \textsc{mC4} corpus \citep{2020t5}, which is released  under the terms of ODC-BY. Pre-training language models is the intended use of this corpus. We uniformly sample data from the partitions of the following languages: \texttt{it}, \texttt{es}, \texttt{fr}, \texttt{pt}, \texttt{de}, \texttt{en}, \texttt{nl}, \texttt{af}, \texttt{zu}, \texttt{sn}, \texttt{sw}, \texttt{xh}, \texttt{ru}, \texttt{uk}, \texttt{bg}, \texttt{sr}. This sampling ensures balanced representation of all languages.
We take checkpoints at powers of two \(\{1, 2, 4,...,131072\}\) and regular 5000-step intervals. Training took 72 hours on a single NVIDIA A100-SXM4-80GB.

\section{Linear Probing for Language Identity}
\label{app:probing}

For every language that appears in both the model's pre-training data and the \oscar \citep{OrtizSuarezSagotRomary2019} corpus, we sample 100 sentences, splitting them into 80 for training and 20 for testing. Each sentence \(s^l_i\) is tokenized into a sequence of tokens \([t_{i,0}^ l,t_{i,1}^l,...,t_{i,T_i-1}^l]\). For each tokenized sentence, the model \(\mathcal{M}\) produces hidden representations: \(h^l_{i,0}, h_{i, 1}^l,...,h_{i,T_1-1}^l = \mathcal{M}(t_{i,0}^l, t_{i,1}^l,...,t^l_{i,T_i-1}) \).
From these sequences of hidden states, we randomly sample one token position per sentence and extract the hidden representation at that position. We then train a logistic regression classifier for each layer to predict the language of origin for each hidden state. To ensure robustness, we repeat this experiment with three different random seeds.

As the \bloom models have different available checkpoint intervals and our toy model is pre-trained on a different set of languages, the results are not directly comparable across models. However, the observed trends are consistent between all models regarding the evolution of representations across training checkpoints.

\begin{figure}[htb!]
    \centering
    \includegraphics[width=\linewidth,clip,trim=1cm 0cm 2.5cm 0cm]{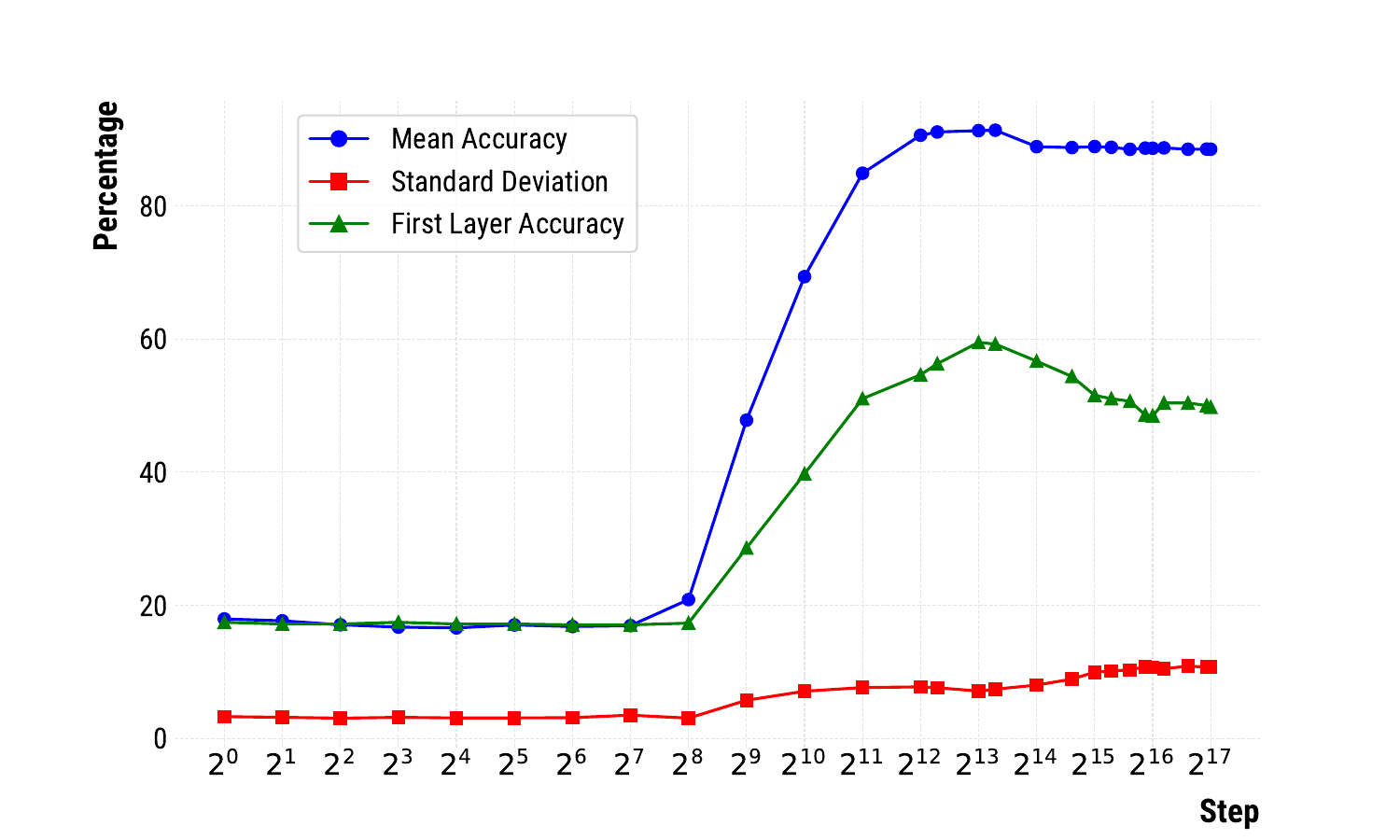}
    \caption{Language identification probing accuracy throughout training of our toy model. For each checkpoint, we show: (1) the mean accuracy across layers, (2) the standard deviation across layers (indicating how much accuracy varies between layers), and (3) the first layer's accuracy, which exhibits the most significant changes during training. Results averaged over three random seeds.}
    \label{fig:toy-model-probing}
\end{figure}

\begin{figure}[htb!]
    \centering
    \includegraphics[width=\linewidth,clip,trim=2cm 0cm 2.5cm 0cm]{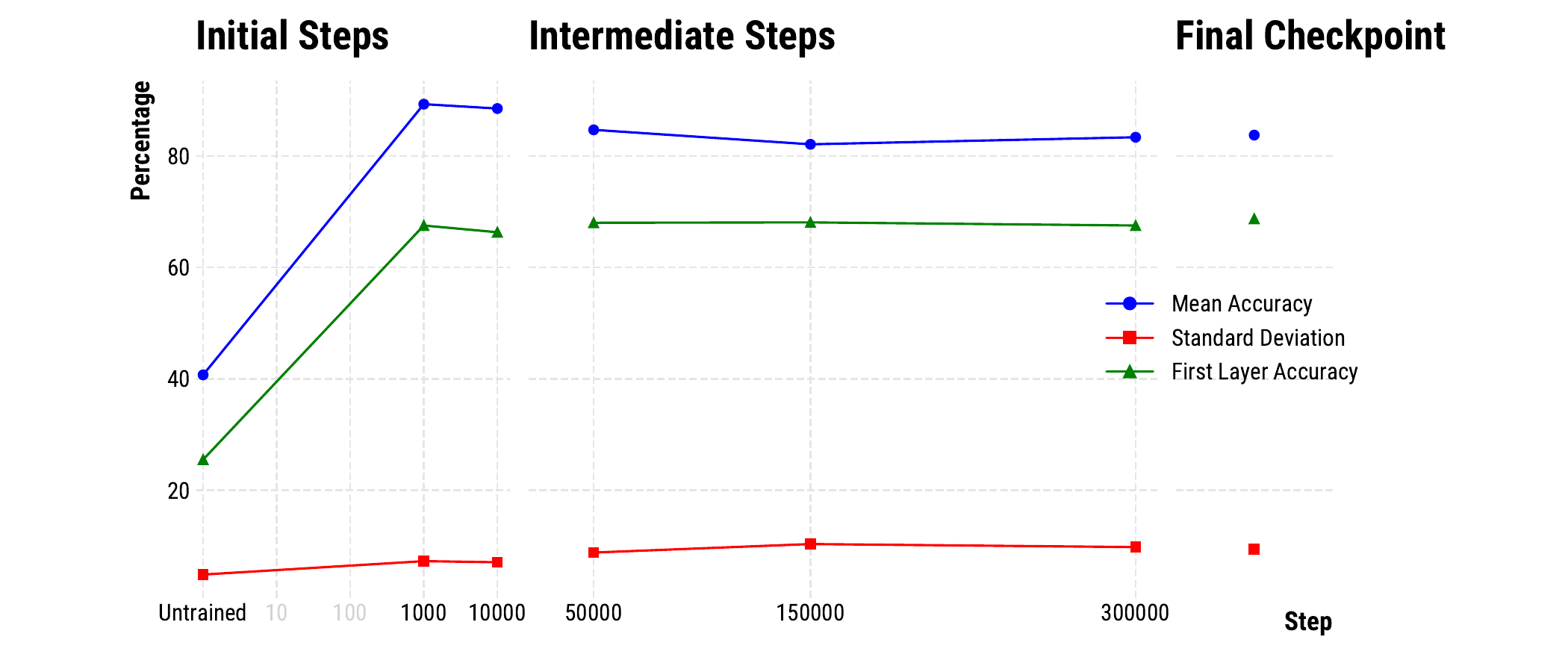}
    \caption{Language identification probing accuracy
throughout training of \bloomb. For each checkpoint, we show: (1) mean accuracy across layers, (2)
standard deviation across layers (indicating how much
accuracy varies between layers), and (3) first layer accu-
racy, which exhibits the most significant changes during
training. Results averaged over three random seeds.}
    \label{fig:full-training-7b1}
\end{figure}

\begin{figure*}[t!]
    \centering
    \begin{subfigure}[t]{0.5\textwidth}
        \centering
        \includegraphics[width=\linewidth,trim=2cm 0cm 4cm 1cm,clip]{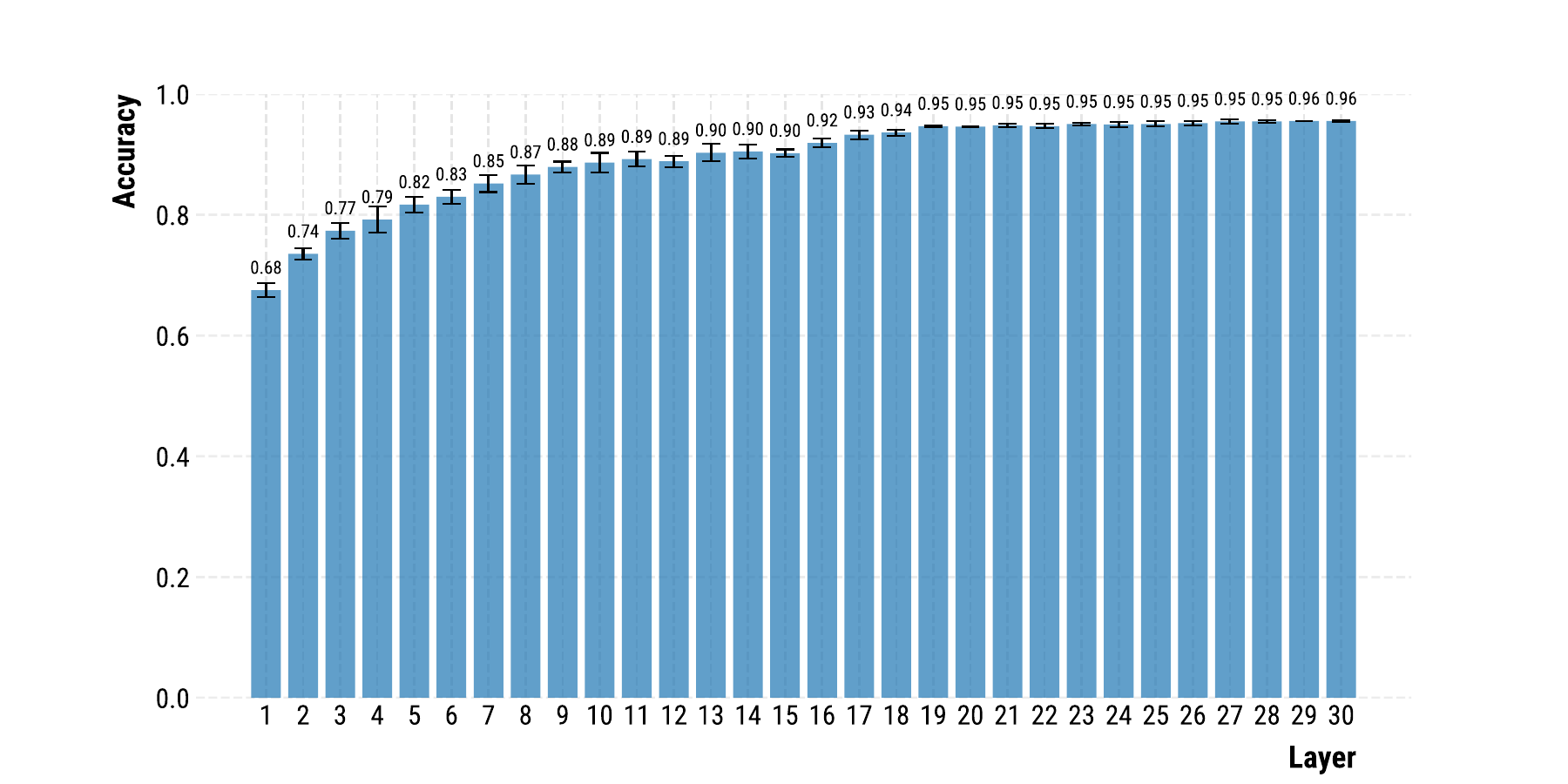}
        \caption{Early training stage (step \num{1000}).}
    \end{subfigure}%
    ~ 
    \begin{subfigure}[t]{0.5\textwidth}
        \centering
        \includegraphics[width=\linewidth,trim=2cm 0cm 4cm 1cm,clip]{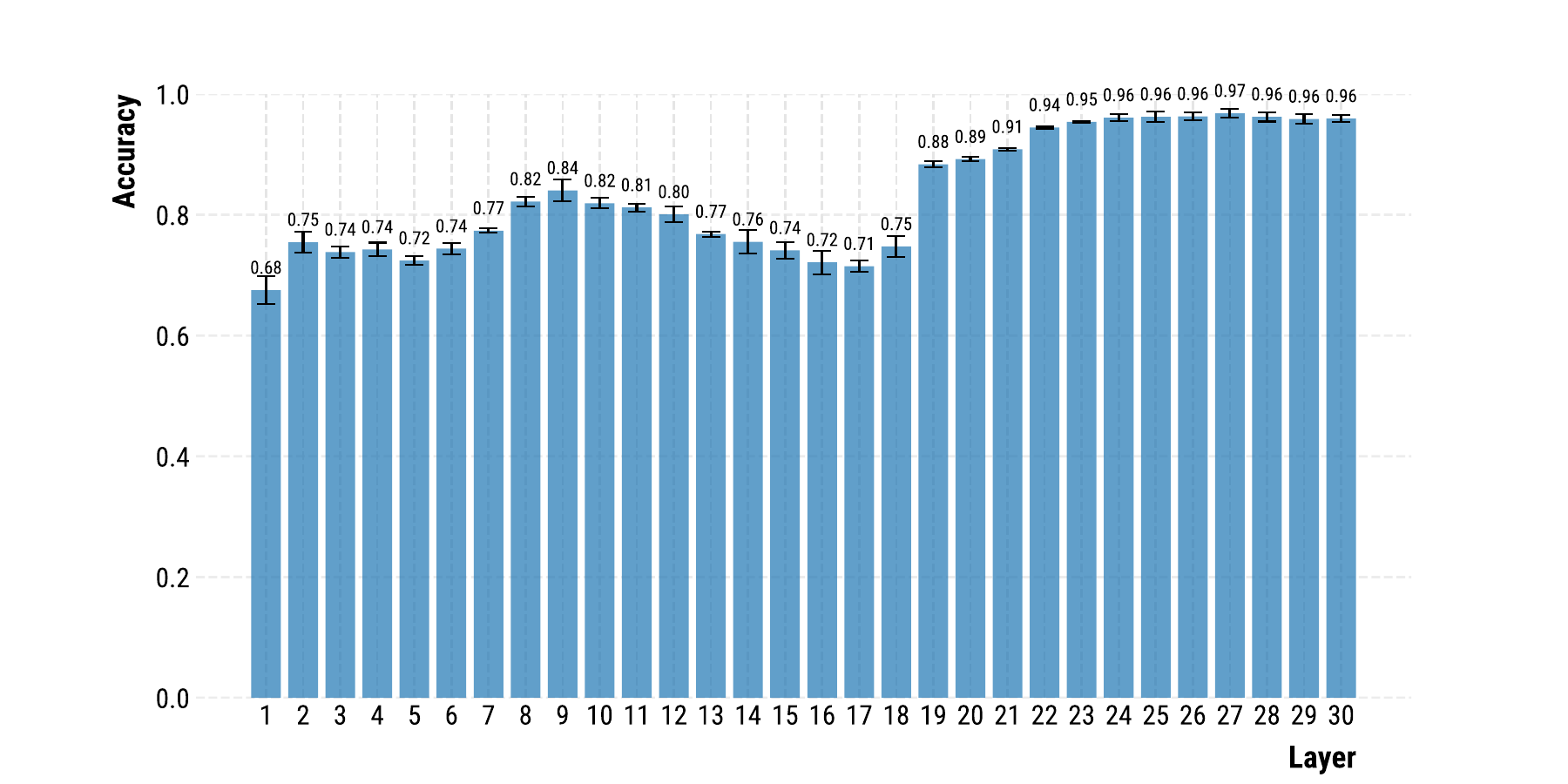}
        \caption{Late training stage (step \num{300000}).}
    \end{subfigure}
    \caption{Language identity probing classification accuracy across layers of the \bloomb model at different training stages. Higher accuracy indicates that language-specific information is more easily extractable from the hidden states at that layer. Error bars show standard deviation across three random seeds.}
    \label{fig:probing_detailed_7b1}
\end{figure*}

We present training progression results for our toy model in \Cref{fig:toy-model-probing} and for \textsc{BLOOM-7b1} in \Cref{fig:full-training-7b1}, with a detailed layer-wise comparison for \bloomb shown in \Cref{fig:probing_detailed_7b1}.

\begin{figure}
    \centering
    \includegraphics[width=\linewidth,trim=2cm 0cm 4cm 1cm,clip]{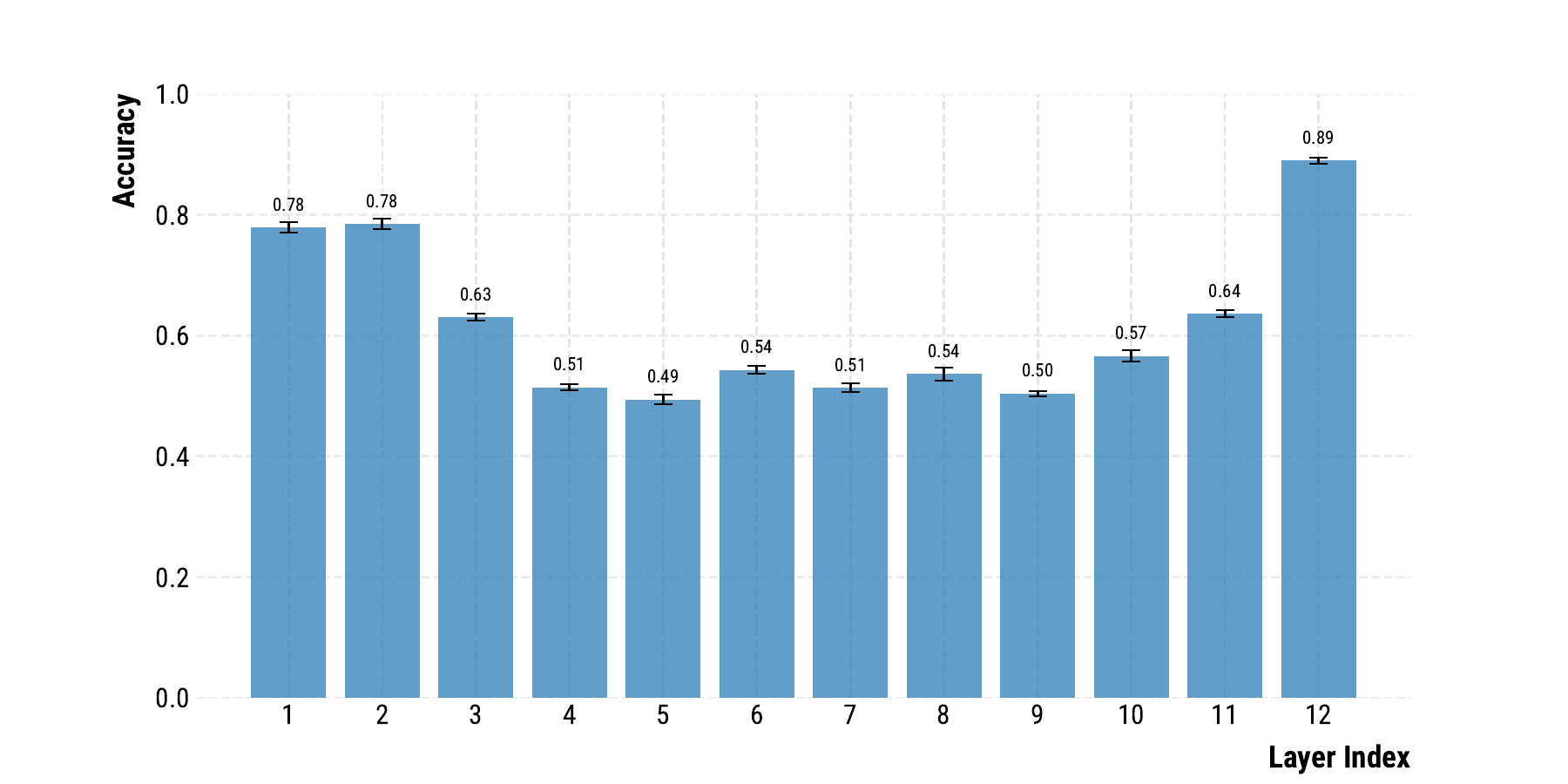}
    \caption{Layer-wise language identity probing on \xlmr \textsc{base}. Higher accuracy indicates that language-specific information is more easily extractable from the hidden states at that layer. Error bars show standard deviation across three random seeds.}
    \label{fig:xlmr-probing}
\end{figure}

\begin{figure}
        \centering
        \includegraphics[width=\linewidth,trim=2cm 0cm 4cm 1cm,clip]{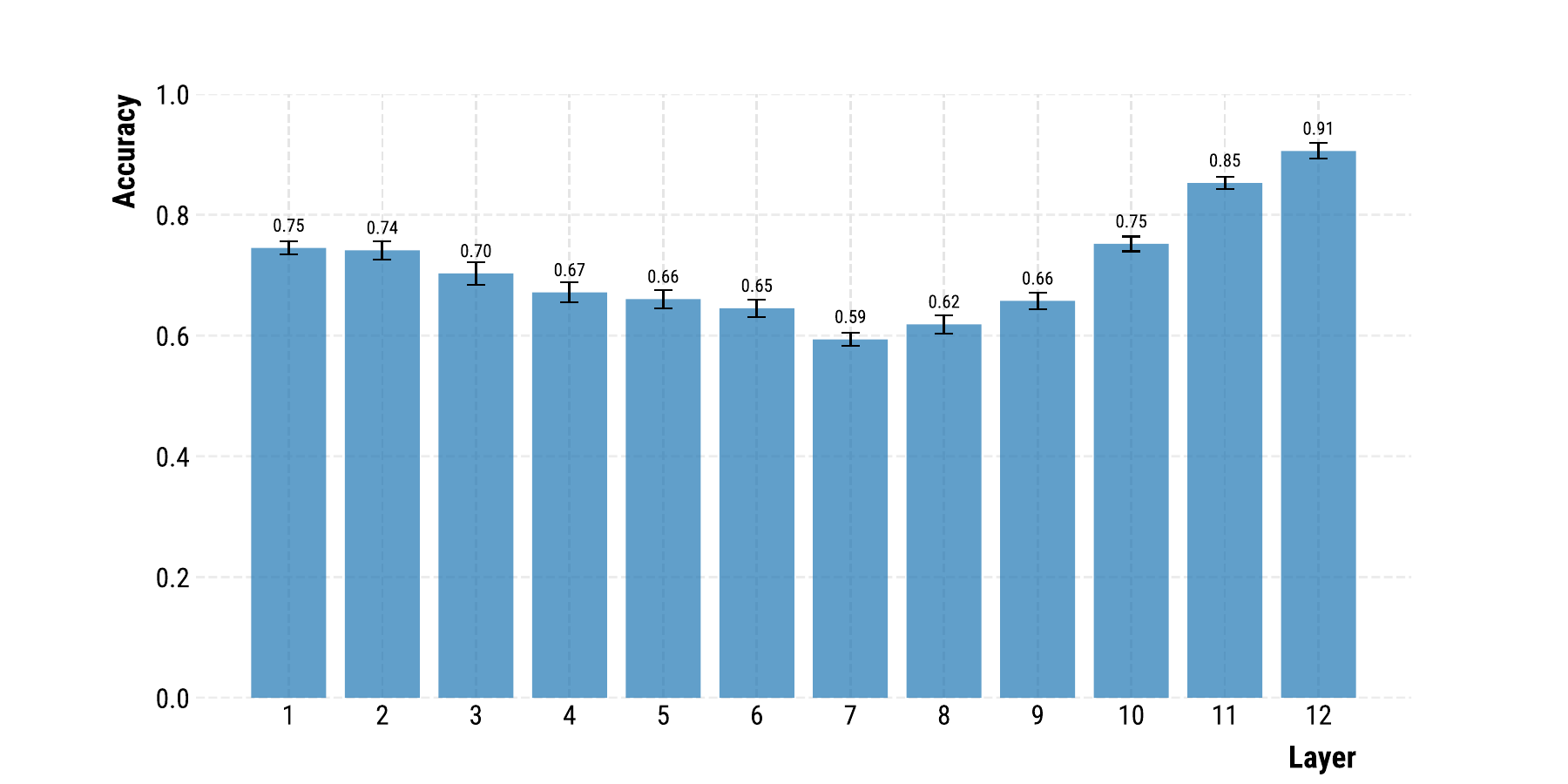}
        \caption{Layer-wise language identity probing on \mbert \textsc{base cased}. Higher accuracy indicates that language-specific information is more easily extractable from the hidden states at that layer. Error bars show standard deviation across three random seeds.}
        \label{fig:mbert-probing}
\end{figure}

For comparison, we include results from the encoder-only models \xlmr \textsc{base} (\Cref{fig:xlmr-probing}) and \mbert \textsc{base cased} (\Cref{fig:mbert-probing}). While decoder-only models exhibit relatively weak performance in early layers, which then increases, encoder-only models display a u-shaped pattern for language identification.

\begin{table}[htb!]
    \centering
    \begin{tabular}{ll}
    \texttt{attention\_dropout} & \(0.1\) \\
        \texttt{attention\_heads} & \(8\)\\
        \texttt{d\_model} & \(512\) \\
        \texttt{dropout} & \(0.1\)\\
        \texttt{ffn\_dim} & \(4096\) \\
        \texttt{num\_layers} & \(24\) \\
        \texttt{vocab\_size} & \(256008\)
    \end{tabular}
    \caption{Configuration details of our toy model.}
    \label{tab:config}
\end{table}

\clearpage

\section{Expert Neuron Data}
\label{app:data}

We construct binary concept identification datasets using \onesec \citep{scarlini-etal-2019-just}, which provides sentences where one word per sentence is annotated with its \wordnet sense. The data is licensed under a Creative Commons Attribution-Noncommercial-Share Alike 4.0 License for research purposes. From this data, we sample 200 concepts, ensuring each has at least 100 sentences containing a word annotated with that concept. For each concept, we create negative examples by randomly sampling sentences from other concepts. 

We then use \nllb \citep{costa2022no} with greedy decoding to translate all datasets into all of our target languages, thereby creating parallel corpora. For computational efficiency, our toy model and \bloomb experiments use a random subset of 100 concepts. The full concept lists are available in \Cref{tab:200concepts} (\bloomm) and \Cref{tab:100concepts} (\bloomb and toy model).

\begin{table*}
    \centering
    \begin{tabular}{p{.9\linewidth}}
        \texttt{acceptance-1\_09\_00}, \texttt{account-1\_10\_00}, \texttt{accumulation-1\_22\_00}, \texttt{action-1\_04\_02}, \texttt{adaptation-1\_22\_00}, \texttt{adviser-1\_18\_00}, \texttt{aftershock-1\_11\_00}, \texttt{agent-1\_17\_00}, \texttt{american-1\_18\_00}, \texttt{amount-1\_07\_00}, \texttt{amount-1\_21\_00}, \texttt{appearance-1\_11\_00}, \texttt{area-1\_15\_01}, \texttt{assumption-1\_09\_00}, \texttt{attendance-1\_28\_00}, \texttt{attention-1\_04\_01}, \texttt{authority-1\_18\_01}, \texttt{backing-1\_06\_00}, \texttt{bacterium-1\_05\_00}, \texttt{band-1\_14\_00}, \texttt{bank-1\_14\_00}, \texttt{bar-1\_06\_04}, \texttt{barrel-1\_06\_01}, \texttt{bay-1\_17\_00}, \texttt{beat-1\_15\_00}, \texttt{bell-1\_06\_02}, \texttt{bill-1\_10\_01}, \texttt{body-1\_14\_00}, \texttt{boost-1\_04\_00}, \texttt{bottom-1\_15\_00}, \texttt{bourbon-1\_18\_01}, \texttt{box-1\_06\_02}, \texttt{capital-1\_21\_01}, \texttt{cent-1\_21\_00}, \texttt{center-1\_15\_01}, \texttt{ceo-1\_18\_00}, \texttt{childhood-1\_26\_00}, \texttt{church-1\_06\_00}, \texttt{circle-1\_14\_00}, \texttt{cleric-1\_18\_00}, \texttt{client-1\_18\_01}, \texttt{commitment-1\_07\_01}, \texttt{companion-1\_18\_02}, \texttt{compensation-1\_22\_00}, \texttt{conservative-1\_18\_00}, \texttt{contract-1\_10\_01}, \texttt{contractor-1\_18\_00}, \texttt{copy-1\_10\_00}, \texttt{crystal-1\_27\_00}, \texttt{cycle-1\_14\_00}, \texttt{deposit-1\_19\_00}, \texttt{desk-1\_06\_00}, \texttt{duty-1\_04\_00}, \texttt{e-mail-1\_10\_00}, \texttt{earthquake-1\_11\_00}, \texttt{economy-1\_09\_01}, \texttt{edition-1\_14\_00}, \texttt{election-1\_04\_01}, \texttt{emotion-1\_12\_00}, \texttt{end-1\_15\_00}, \texttt{enterprise-1\_04\_00}, \texttt{equity-1\_21\_00}, \texttt{equity-1\_21\_01}, \texttt{excess-1\_07\_02}, \texttt{execution-1\_04\_00}, \texttt{expulsion-1\_04\_01}, \texttt{eyebrow-1\_08\_00}, \texttt{face-1\_08\_00}, \texttt{faithful-1\_14\_00}, \texttt{family-1\_14\_00}, \texttt{favor-1\_04\_00}, \texttt{fee-1\_21\_00}, \texttt{feeling-1\_03\_00}, \texttt{find-1\_04\_00}, \texttt{forehead-1\_08\_00}, \texttt{foreigner-1\_18\_00}, \texttt{game-1\_04\_03}, \texttt{genesis-1\_10\_00}, \texttt{germany-1\_15\_00}, \texttt{goal-1\_15\_00}, \texttt{gold-1\_21\_00}, \texttt{governance-1\_04\_00}, \texttt{grievance-1\_10\_01}, \texttt{hall-1\_06\_03}, \texttt{height-1\_07\_00}, \texttt{house-1\_14\_01}, \texttt{hydrogen-1\_27\_00}, \texttt{information-1\_09\_00}, \texttt{infrastructure-1\_06\_00}, \texttt{initial-1\_10\_00}, \texttt{injection-1\_27\_00}, \texttt{insight-1\_12\_00}, \texttt{inspiration-1\_06\_00}, \texttt{interference-1\_10\_00}, \texttt{involvement-1\_24\_00}, \texttt{job-1\_04\_00}, \texttt{joy-1\_12\_00}, \texttt{judge-1\_18\_00}, \texttt{kid-1\_18\_00}, \texttt{killer-1\_18\_00}, \texttt{kind-1\_09\_00}, \texttt{kitchen-1\_06\_00}, \texttt{lack-1\_26\_00}, \texttt{lady-1\_18\_02}, \texttt{length-1\_07\_00}, \texttt{level-1\_26\_01}, \texttt{library-1\_14\_00}, \texttt{lifetime-1\_28\_00}, \texttt{machine-1\_06\_00}, \texttt{march-1\_04\_00}, \texttt{march-1\_28\_00}, \texttt{margin-1\_07\_00}, \texttt{master-1\_18\_00}, \texttt{math-1\_09\_00}, \texttt{member-1\_18\_00}, \texttt{memory-1\_09\_01}, \texttt{message-1\_10\_00}, \texttt{minister-1\_18\_00}, \texttt{ministry-1\_06\_00}, \texttt{minute-1\_28\_01}, \texttt{money-1\_21\_00}, \texttt{money-1\_21\_02}, \texttt{morale-1\_07\_00}, \texttt{move-1\_04\_01}, \texttt{mr-1\_10\_00}, \texttt{mystery-1\_09\_00}, \texttt{need-1\_17\_00}, \texttt{negotiation-1\_10\_00}, \texttt{news-1\_10\_01}, \texttt{nickname-1\_10\_01}, \texttt{nobility-1\_14\_00}, \texttt{notion-1\_09\_00}, \texttt{one-1\_23\_00}, \texttt{order-1\_07\_01}, \texttt{paint-1\_06\_00}, \texttt{paradox-1\_10\_00}, \texttt{participant-1\_18\_00}, \texttt{participant-1\_18\_01}, \texttt{pattern-1\_09\_00}, \texttt{percent-1\_24\_00}, \texttt{percentage-1\_24\_00}, \texttt{perimeter-1\_25\_00}, \texttt{person-1\_03\_00}, \texttt{personality-1\_18\_00}, \texttt{pet-1\_05\_00}, \texttt{phase-1\_26\_00}, \texttt{phosphorus-1\_27\_00}, \texttt{pier-1\_06\_00}, \texttt{place-1\_15\_04}, \texttt{politician-1\_18\_01}, \texttt{poster-1\_18\_00}, \texttt{premonition-1\_12\_00}, \texttt{president-1\_18\_01}, \texttt{president-1\_18\_04}, \texttt{process-1\_04\_00}, \texttt{program-1\_09\_00}, \texttt{programme-1\_10\_00}, \texttt{pub-1\_06\_00}, \texttt{race-1\_11\_00}, \texttt{rank-1\_14\_00}, \texttt{recovery-1\_11\_00}, \texttt{refinery-1\_06\_00}, \texttt{regard-1\_09\_01}, \texttt{release-1\_04\_01}, \texttt{release-1\_06\_00}, \texttt{role-1\_04\_00}, \texttt{schoolteacher-1\_18\_00}, \texttt{score-1\_10\_00}, \texttt{scourge-1\_26\_00}, \texttt{senator-1\_18\_00}, \texttt{september-1\_28\_00}, \texttt{signal-1\_16\_00}, \texttt{situation-1\_26\_01}, \texttt{skepticism-1\_09\_01}, \texttt{solution-1\_27\_00}, \texttt{someone-1\_03\_00}, \texttt{space-1\_03\_00}, \texttt{spain-1\_15\_00}, \texttt{spite-1\_12\_00}, \texttt{statement-1\_10\_00}, \texttt{step-1\_04\_02}, \texttt{striker-1\_18\_02}, \texttt{suicide-1\_04\_00}, \texttt{suspension-1\_28\_00}, \texttt{system-1\_06\_00}, \texttt{tax-1\_21\_00}, \texttt{thing-1\_04\_00}, \texttt{thinking-1\_09\_00}, \texttt{times-1\_04\_00}, \texttt{triage-1\_04\_00}, \texttt{trial-1\_04\_00}, \texttt{type-1\_18\_00}, \texttt{unemployment-1\_26\_00}, \texttt{verdict-1\_04\_00}, \texttt{vicar-1\_18\_00}, \texttt{wealth-1\_26\_00}, \texttt{wednesday-1\_28\_00}, \texttt{will-1\_09\_00}, \texttt{yield-1\_04\_00}, \texttt{zip-1\_10\_00}
    \end{tabular}
    \caption{Complete set of 200 randomly sampled \wordnet senses (alphabetically ordered) used in \bloomm experiments.}
    \label{tab:200concepts}
\end{table*}

\begin{table*}
    \centering
    \begin{tabular}{p{.9\linewidth}}
    \texttt{accumulation-1\_22\_00}, \texttt{action-1\_04\_02}, \texttt{adviser-1\_18\_00}, \texttt{aftershock-1\_11\_00}, \texttt{american-1\_18\_00}, \texttt{amount-1\_07\_00}, \texttt{appearance-1\_11\_00}, \texttt{attention-1\_04\_01}, \texttt{authority-1\_18\_01}, \texttt{bar-1\_06\_04}, \texttt{boost-1\_04\_00}, \texttt{bourbon-1\_18\_01}, \texttt{center-1\_15\_01}, \texttt{ceo-1\_18\_00}, \texttt{circle-1\_14\_00}, \texttt{client-1\_18\_01}, \texttt{companion-1\_18\_02}, \texttt{conservative-1\_18\_00}, \texttt{contractor-1\_18\_00}, \texttt{cycle-1\_14\_00}, \texttt{deposit-1\_19\_00}, \texttt{e-mail-1\_10\_00}, \texttt{economy-1\_09\_01}, \texttt{edition-1\_14\_00}, \texttt{equity-1\_21\_00}, \texttt{excess-1\_07\_02}, \texttt{execution-1\_04\_00}, \texttt{eyebrow-1\_08\_00}, \texttt{faithful-1\_14\_00}, \texttt{family-1\_14\_00}, \texttt{favor-1\_04\_00}, \texttt{forehead-1\_08\_00}, \texttt{foreigner-1\_18\_00}, \texttt{genesis-1\_10\_00}, \texttt{germany-1\_15\_00}, \texttt{goal-1\_15\_00}, \texttt{gold-1\_21\_00}, \texttt{governance-1\_04\_00}, \texttt{height-1\_07\_00}, \texttt{house-1\_14\_01}, \texttt{information-1\_09\_00}, \texttt{infrastructure-1\_06\_00}, \texttt{insight-1\_12\_00}, \texttt{inspiration-1\_06\_00}, \texttt{interference-1\_10\_00}, \texttt{job-1\_04\_00}, \texttt{joy-1\_12\_00}, \texttt{kid-1\_18\_00}, \texttt{killer-1\_18\_00}, \texttt{lady-1\_18\_02}, \texttt{length-1\_07\_00}, \texttt{library-1\_14\_00}, \texttt{lifetime-1\_28\_00}, \texttt{march-1\_04\_00}, \texttt{margin-1\_07\_00}, \texttt{master-1\_18\_00}, \texttt{message-1\_10\_00}, \texttt{money-1\_21\_00}, \texttt{morale-1\_07\_00}, \texttt{move-1\_04\_01}, \texttt{mystery-1\_09\_00}, \texttt{negotiation-1\_10\_00}, \texttt{news-1\_10\_01}, \texttt{nobility-1\_14\_00}, \texttt{notion-1\_09\_00}, \texttt{paint-1\_06\_00}, \texttt{participant-1\_18\_00}, \texttt{participant-1\_18\_01}, \texttt{pattern-1\_09\_00}, \texttt{percent-1\_24\_00}, \texttt{perimeter-1\_25\_00}, \texttt{personality-1\_18\_00}, \texttt{pet-1\_05\_00}, \texttt{phosphorus-1\_27\_00}, \texttt{pier-1\_06\_00}, \texttt{place-1\_15\_04}, \texttt{premonition-1\_12\_00}, \texttt{president-1\_18\_01}, \texttt{president-1\_18\_04}, \texttt{program-1\_09\_00}, \texttt{pub-1\_06\_00}, \texttt{race-1\_11\_00}, \texttt{rank-1\_14\_00}, \texttt{refinery-1\_06\_00}, \texttt{release-1\_06\_00}, \texttt{september-1\_28\_00}, \texttt{skepticism-1\_09\_01}, \texttt{someone-1\_03\_00}, \texttt{spite-1\_12\_00}, \texttt{striker-1\_18\_02}, \texttt{system-1\_06\_00}, \texttt{tax-1\_21\_00}, \texttt{thing-1\_04\_00}, \texttt{thinking-1\_09\_00}, \texttt{type-1\_18\_00}, \texttt{unemployment-1\_26\_00}, \texttt{verdict-1\_04\_00}, \texttt{vicar-1\_18\_00}, \texttt{wealth-1\_26\_00}, \texttt{yield-1\_04\_00}
    \end{tabular}
    \caption{Subset of 100 \wordnet senses randomly selected from \Cref{tab:200concepts} (alphabetically ordered), used in \bloomb and toy model experiments for faster experimentation.}
    \label{tab:100concepts}
\end{table*}

\section{Additional Results}
\label{app:results}

\begin{figure*}[t!]
    \centering
    \begin{subfigure}[t]{0.5\textwidth}
        \centering
        \includegraphics[width=\linewidth,trim=1cm 0cm 0cm 0cm,clip]{latex/figures/correlation_matrix_step_1000_per_concept_average_edited.pdf}
        \caption{Step \num{1000}.}
        \label{subfig:neuron_1000_app}
    \end{subfigure}%
    ~ 
    \begin{subfigure}[t]{0.5\textwidth}
        \centering
        \includegraphics[width=\linewidth,trim=1cm 0cm 0cm 0cm,clip]{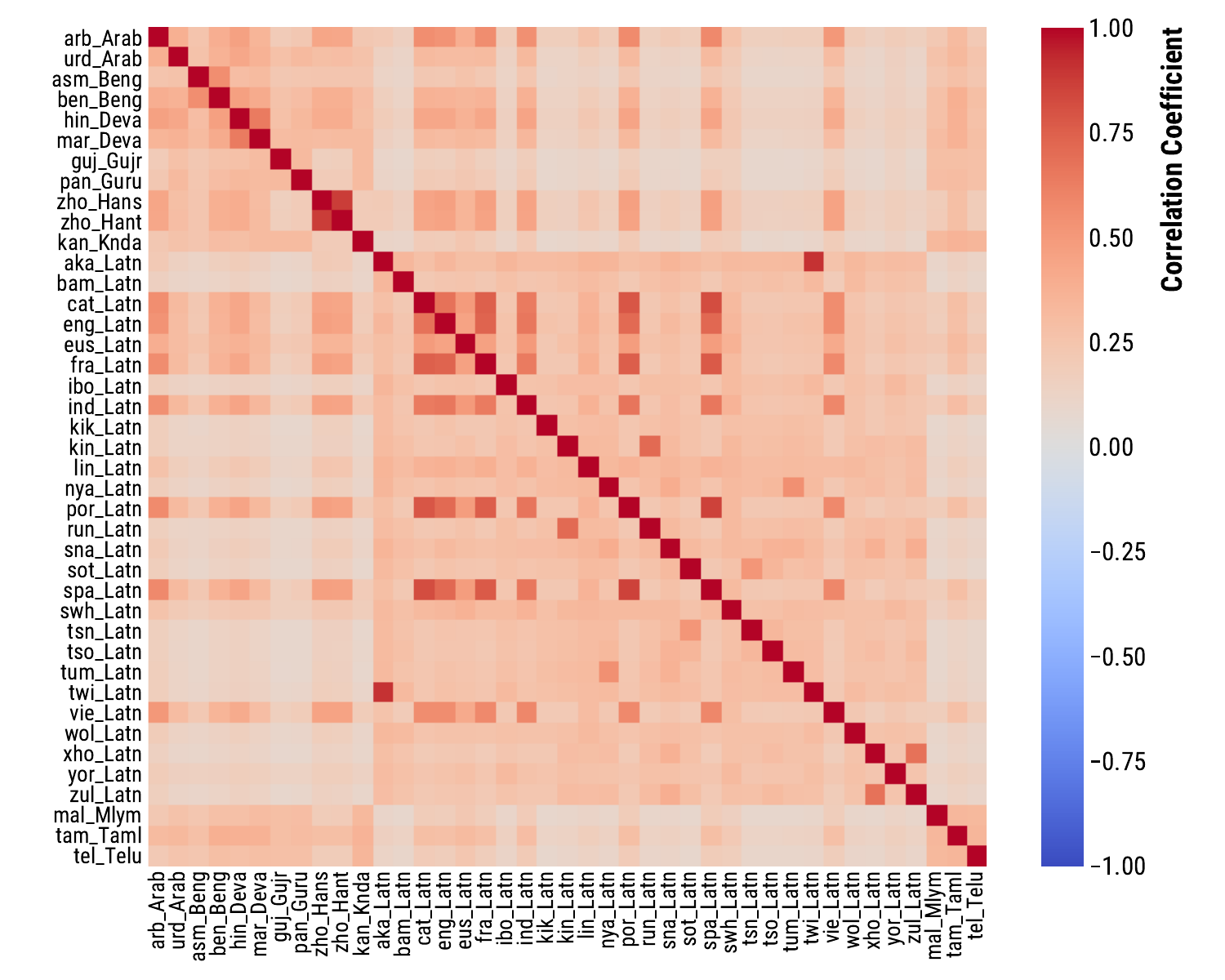}
        \caption{Early training stage (step \num{10000}).}
        \label{subfig:neuron_10000_app}
    \end{subfigure}
    \begin{subfigure}[t]{0.5\textwidth}
        \centering
        \includegraphics[width=\linewidth,trim=1cm 0cm 0cm 0cm,clip]{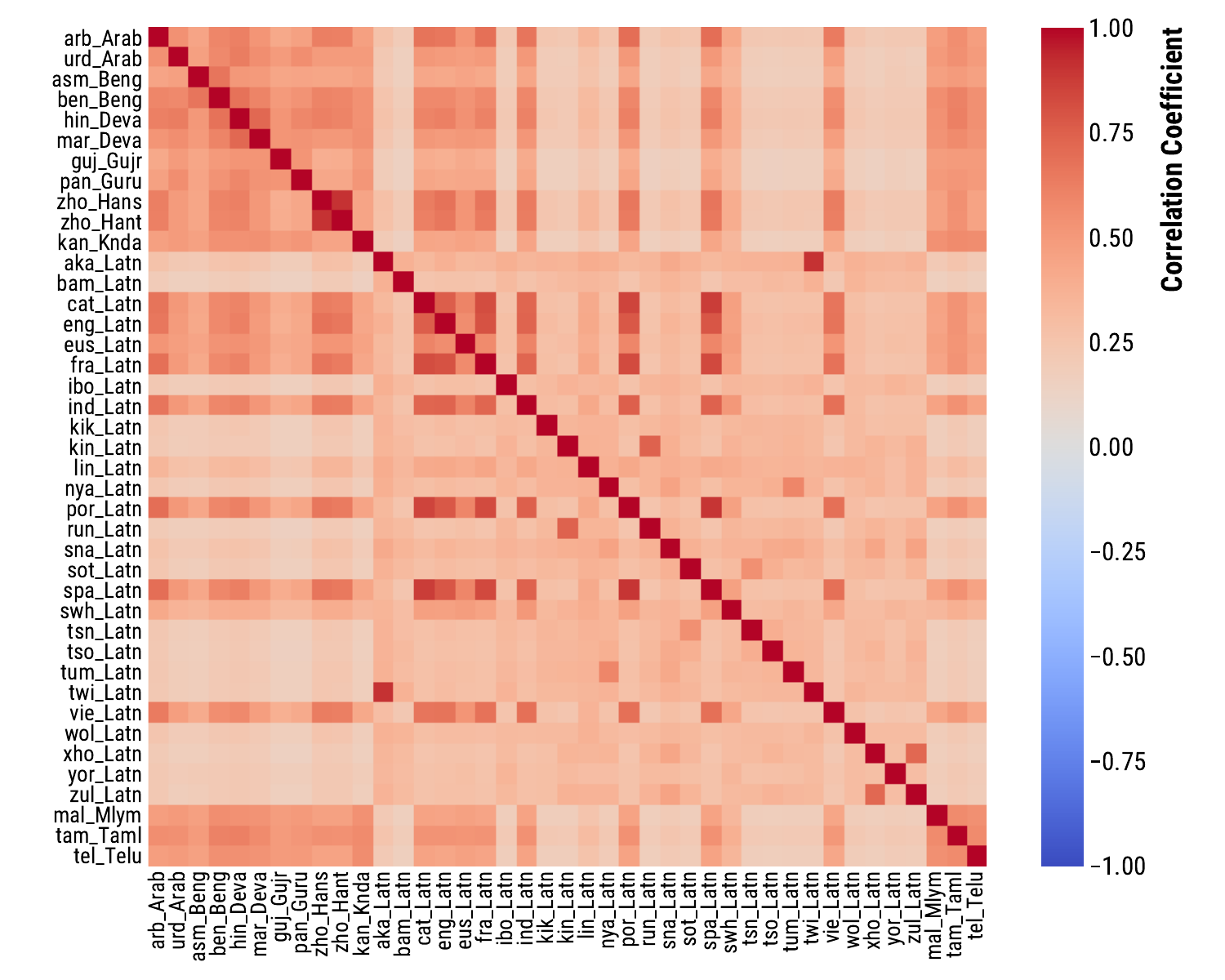}
        \caption{Step \num{100000}.}
        \label{subfig:neuron_100000_app}
    \end{subfigure}%
    ~ 
    \begin{subfigure}[t]{0.5\textwidth}
        \centering
        \includegraphics[width=\linewidth,trim=1cm 0cm 0cm 0cm,clip]{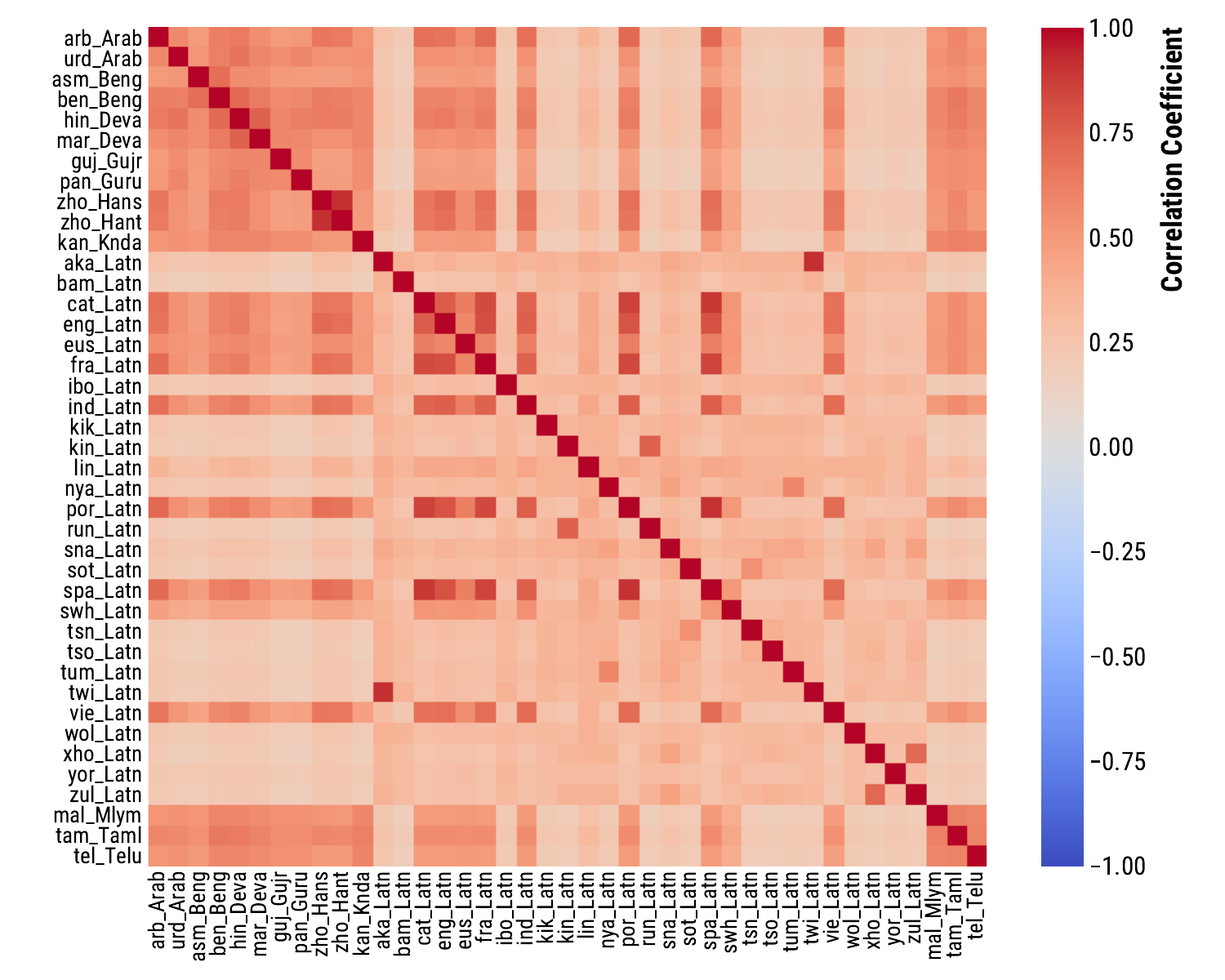}
        \caption{Step \num{200000}.}
        \label{subfig:neuron_200000_app}
    \end{subfigure}
    ~ 
    \begin{subfigure}[t]{0.5\textwidth}
        \centering
        \includegraphics[width=\linewidth,trim=1cm 0cm 0cm 0cm,clip]{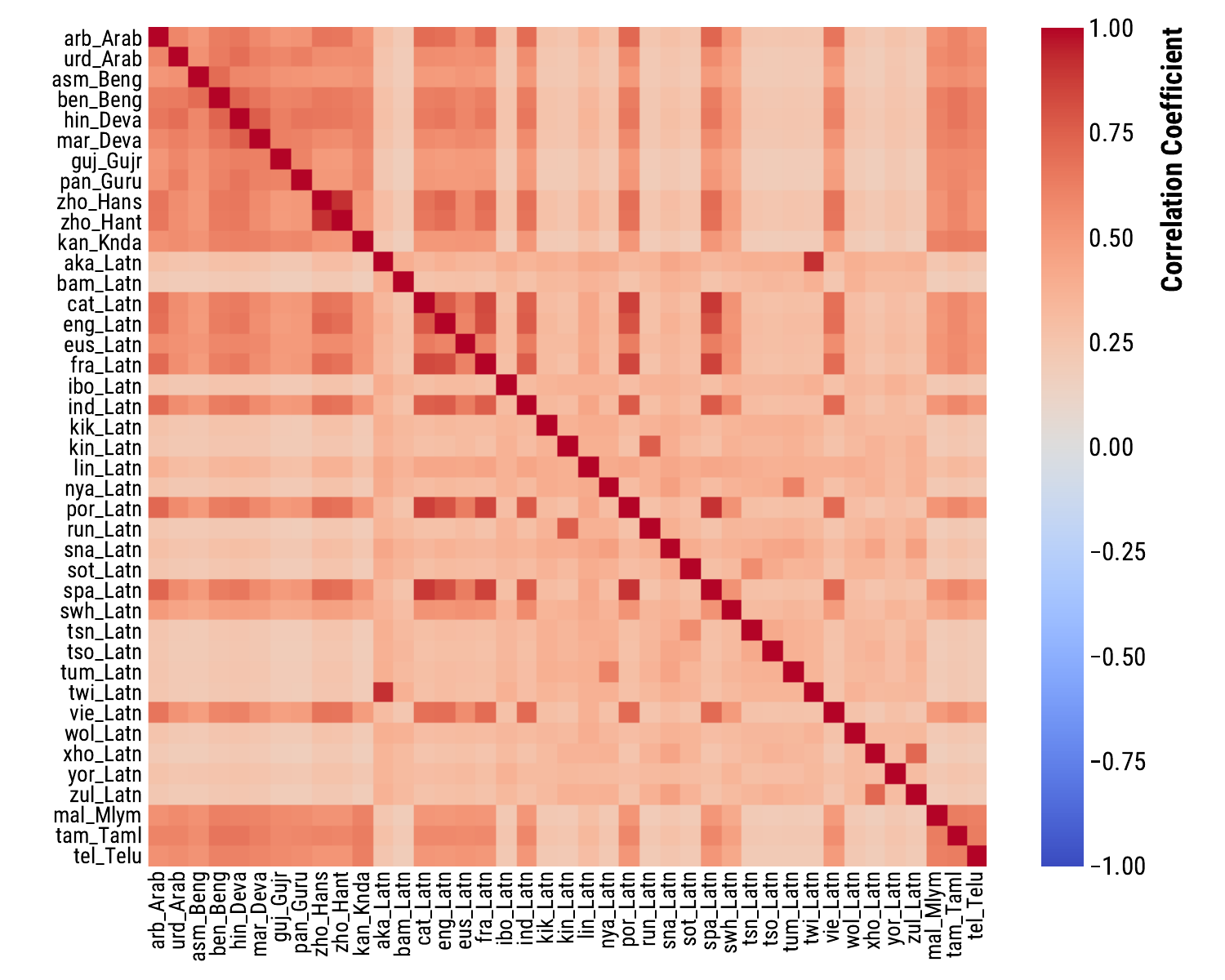}
        \caption{Step \num{300000}.}
        \label{subfig:neuron_300000_app}
    \end{subfigure}%
    ~ 
    \begin{subfigure}[t]{0.5\textwidth}
        \centering
        \includegraphics[width=\linewidth,trim=1cm 0cm 0cm 0cm,clip]{latex/figures/correlation_matrix_step_400000_per_concept_average_edited.pdf}
        \caption{Step \num{400000}.}
        \label{subfig:neuron_400000_app}
    \end{subfigure}
    \caption{Expert neuron alignment of \bloomm at different training stages.}
    \label{fig:neuron_alignment_bloom_full}
\end{figure*}

We show alignment matrices for all checkpoints of \bloomm in \Cref{fig:neuron_alignment_bloom_full}. While there is a clear increase in alignment from step \num{1000} to step \num{100000}, later checkpoints show minimal differences, becoming almost indistinguishable from each other.

\begin{figure*}[t!]
    \centering 
    \begin{subfigure}[t]{0.33\textwidth}
        \centering
        \includegraphics[width=\linewidth,trim=1cm 0cm 0cm 0cm,clip]{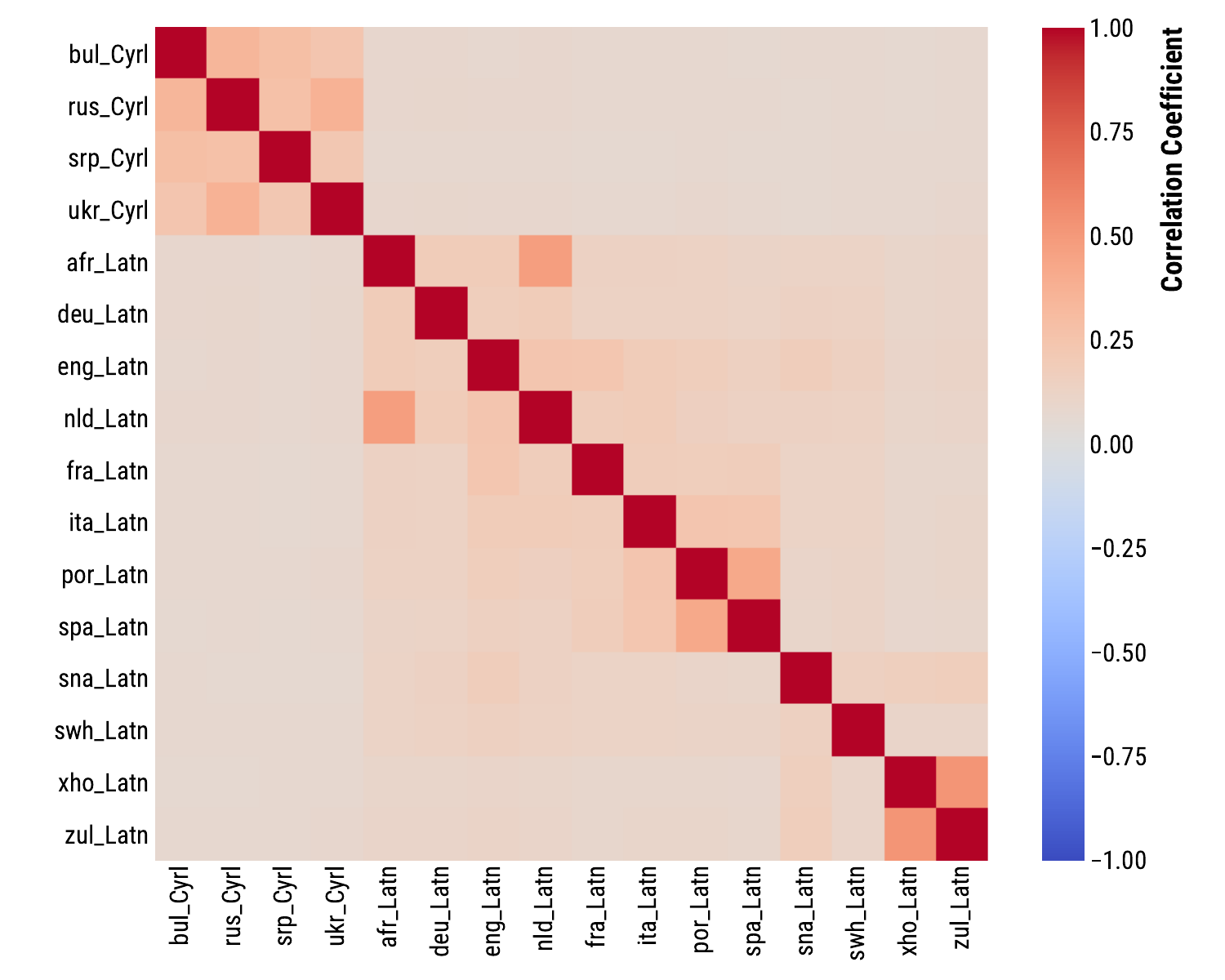}
        \caption{Step \num{64}.}
        \label{subfig:neuron_64_app_toy}
    \end{subfigure}%
    ~ 
    \begin{subfigure}[t]{0.33\textwidth}
        \centering
        \includegraphics[width=\linewidth,trim=1cm 0cm 0cm 0cm,clip]{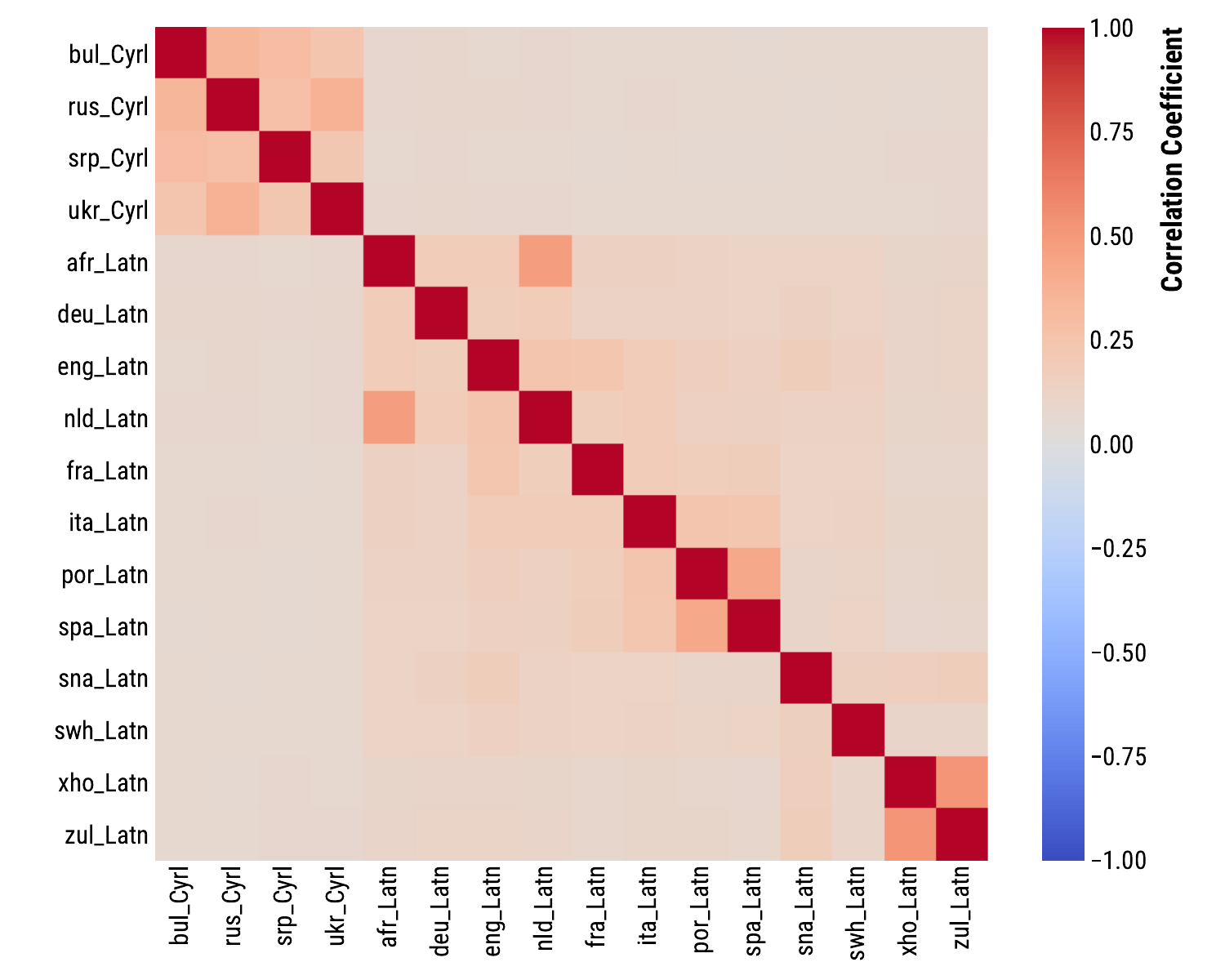}
        \caption{Step \num{128}.}
        \label{subfig:neuron_128_app_toy}
    \end{subfigure}%
    ~ 
    \begin{subfigure}[t]{0.33\textwidth}
        \centering
        \includegraphics[width=\linewidth,trim=1cm 0cm 0cm 0cm,clip]{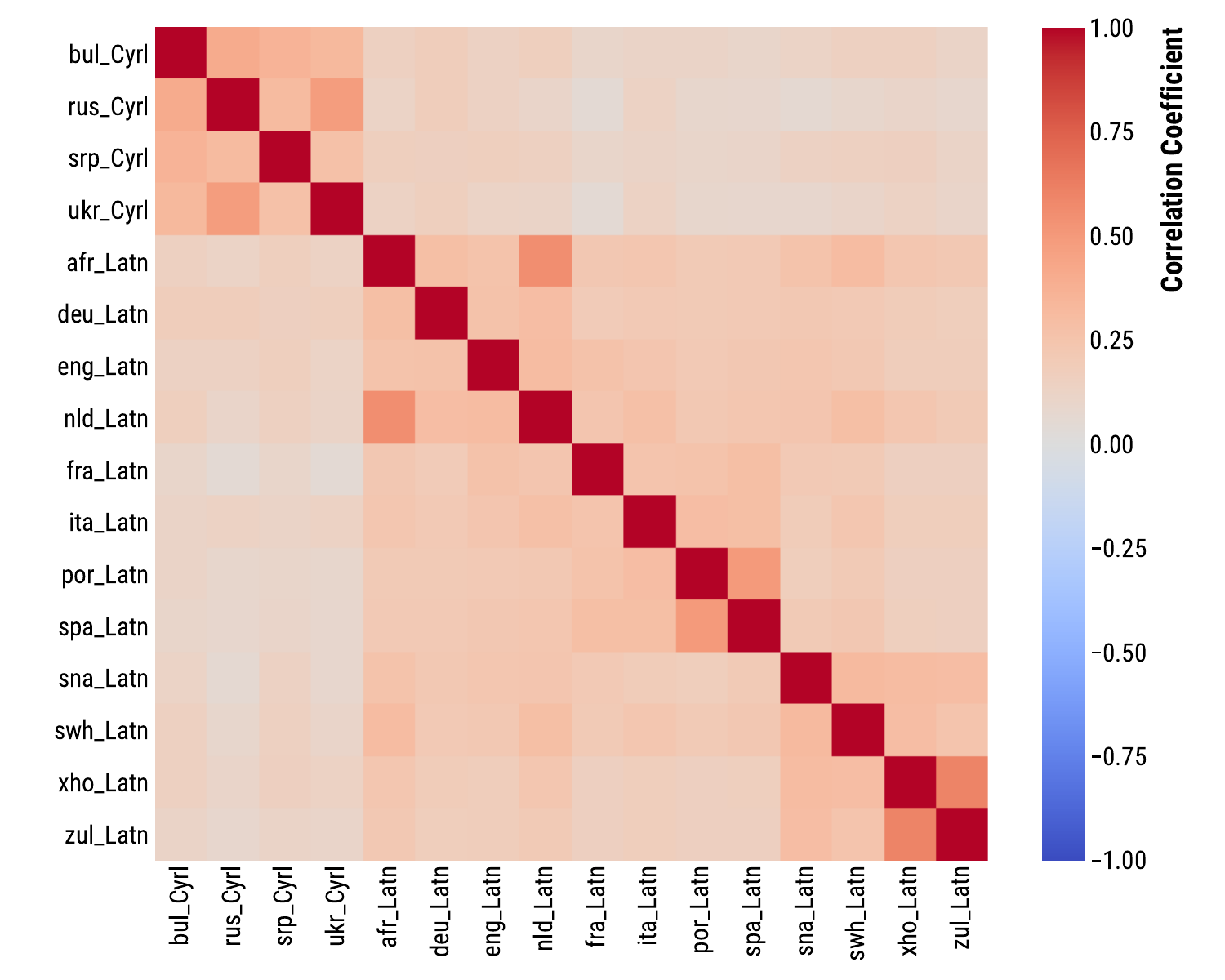}
        \caption{Step \num{256}.}
        \label{subfig:neuron_256_app_toy}
    \end{subfigure}
    ~ 
    \begin{subfigure}[t]{0.33\textwidth}
        \centering
        \includegraphics[width=\linewidth,trim=1cm 0cm 0cm 0cm,clip]{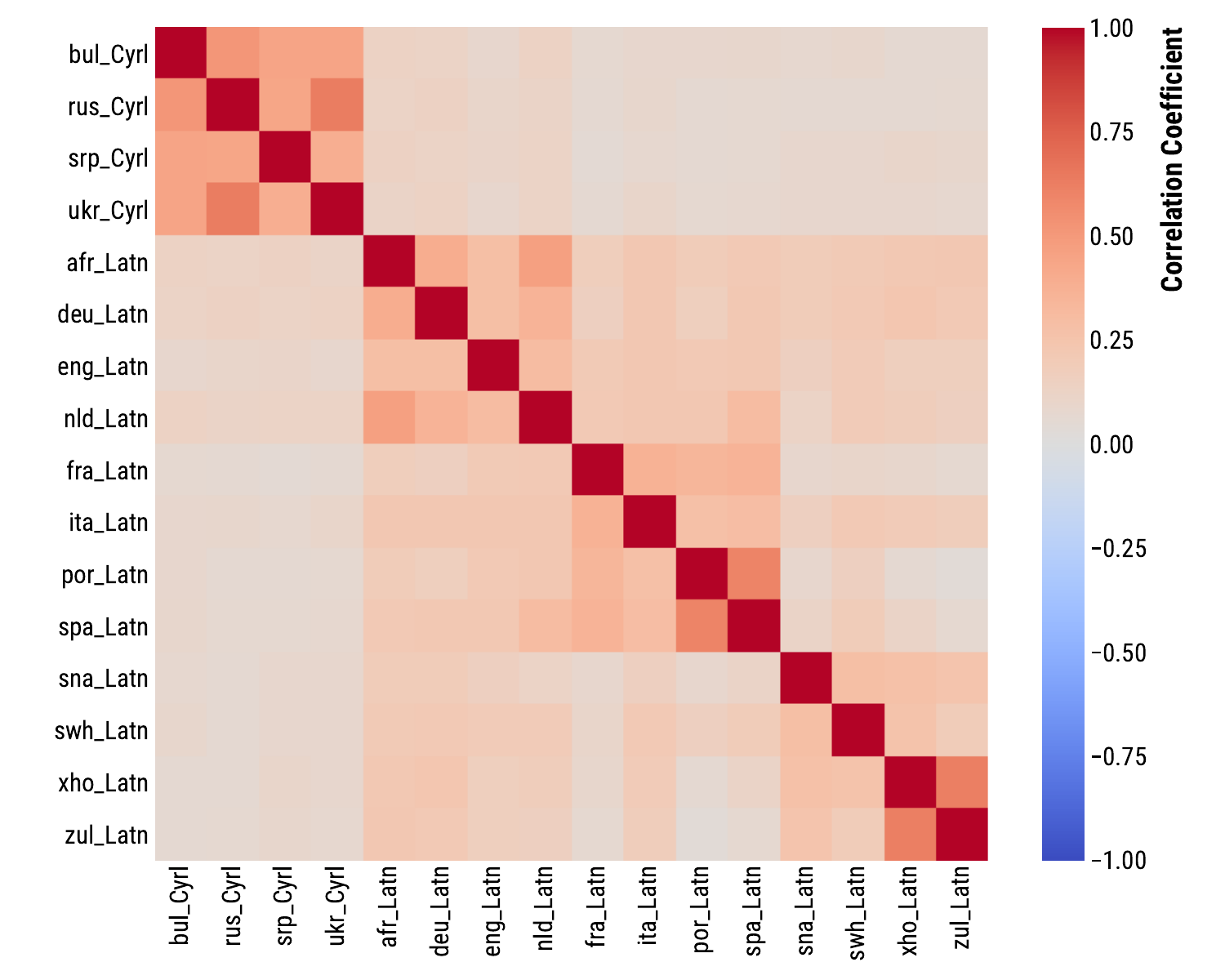}
        \caption{Step \num{512}.}
        \label{subfig:neuron_512_app_toy}
    \end{subfigure}%
    ~ 
    \begin{subfigure}[t]{0.33\textwidth}
        \centering
        \includegraphics[width=\linewidth,trim=1cm 0cm 0cm 0cm,clip]{latex/figures/toy_model/correlation_matrix_step_512_per_concept_average.pdf}
        \caption{Step \num{1024}.}
        \label{subfig:neuron_1024_app_toy}
    \end{subfigure}%
    ~ 
    \begin{subfigure}[t]{0.33\textwidth}
        \centering
        \includegraphics[width=\linewidth,trim=1cm 0cm 0cm 0cm,clip]{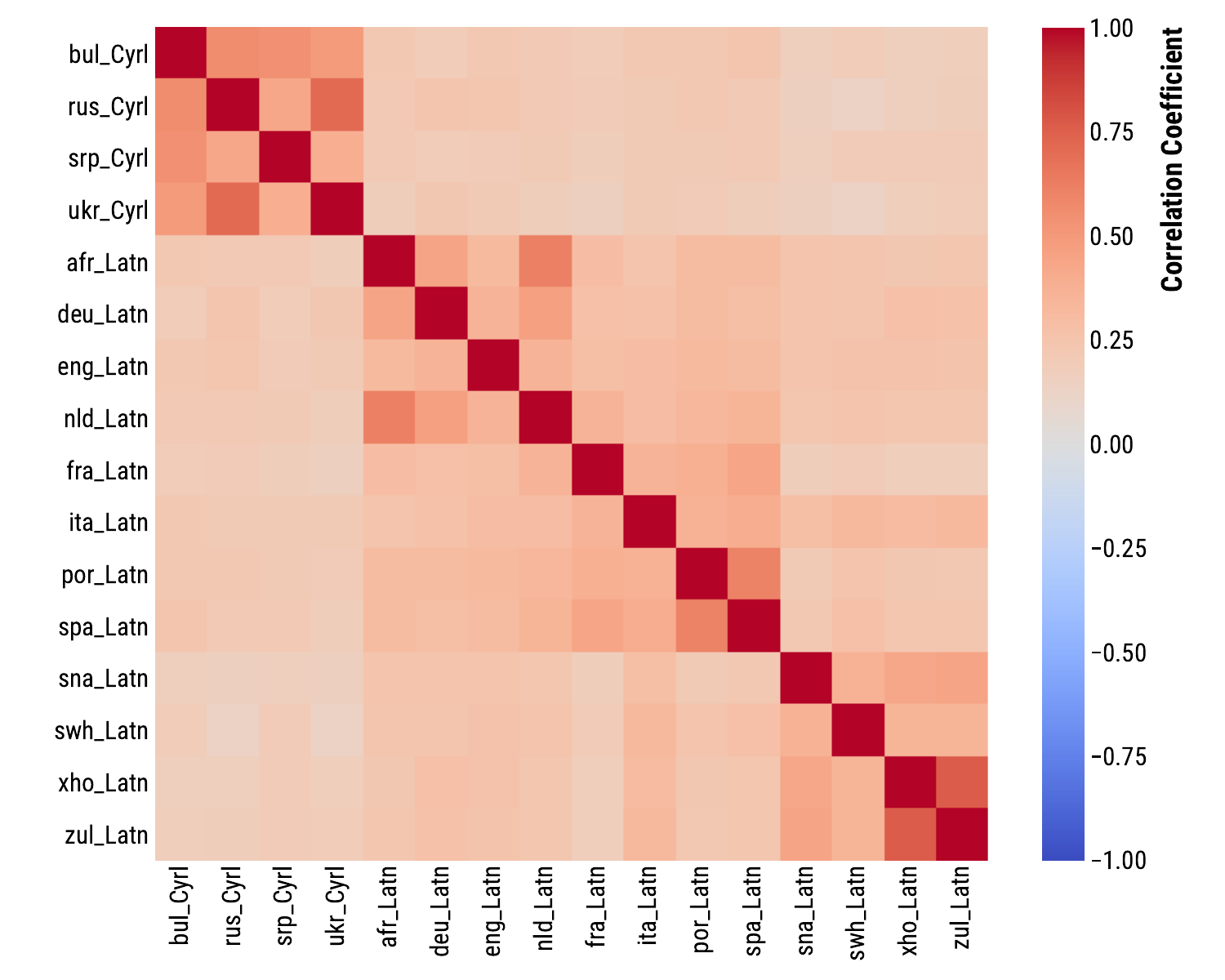}
        \caption{Step \num{2048}.}
        \label{subfig:neuron_2048_app_toy}
    \end{subfigure}
    ~
    \begin{subfigure}[t]{0.33\textwidth}
        \centering
        \includegraphics[width=\linewidth,trim=1cm 0cm 0cm 0cm,clip]{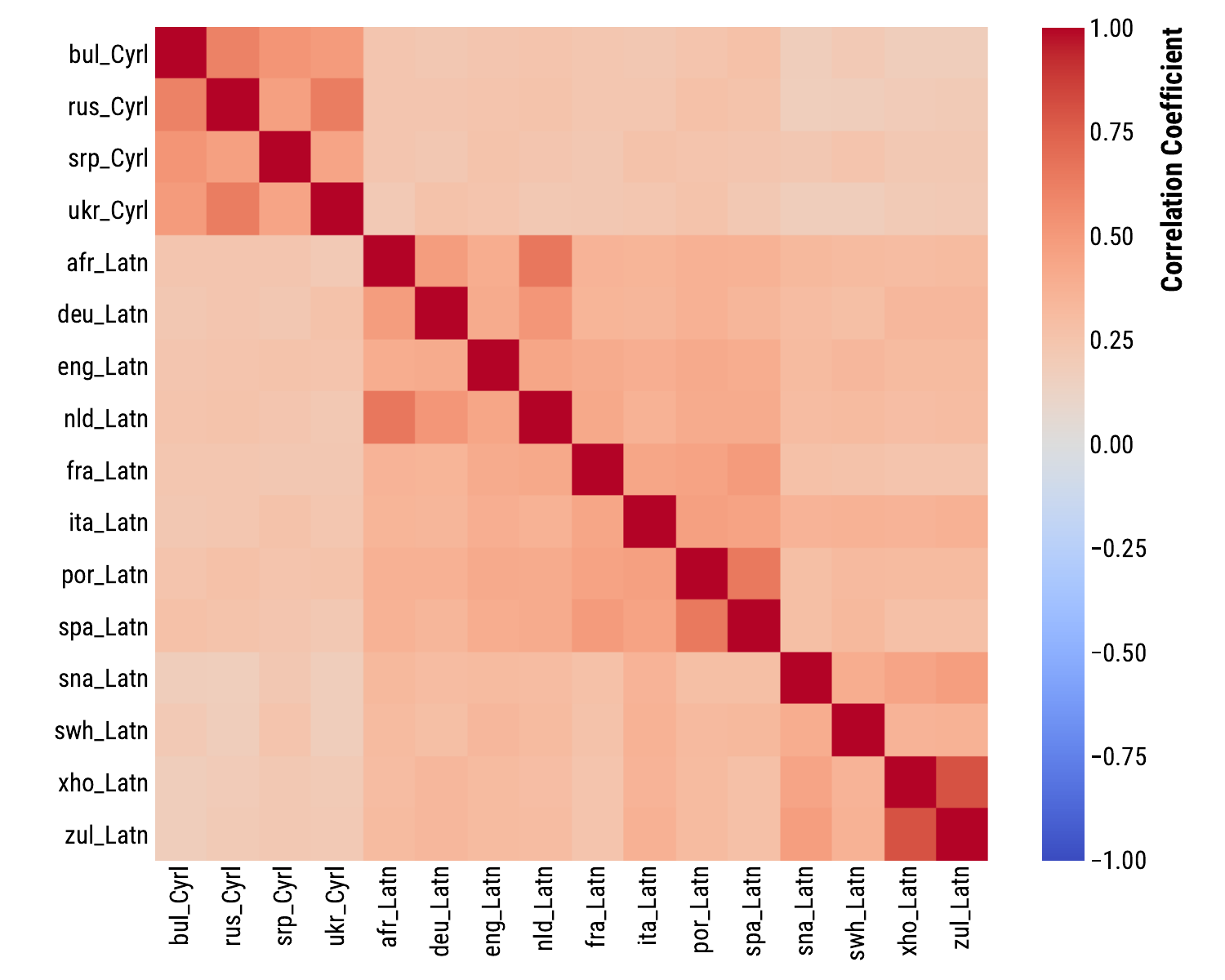}
        \caption{Step \num{4096}.}
        \label{subfig:neuron_4096_app_toy}
    \end{subfigure}%
    ~
    \begin{subfigure}[t]{0.33\textwidth}
        \centering
        \includegraphics[width=\linewidth,trim=1cm 0cm 0cm 0cm,clip]{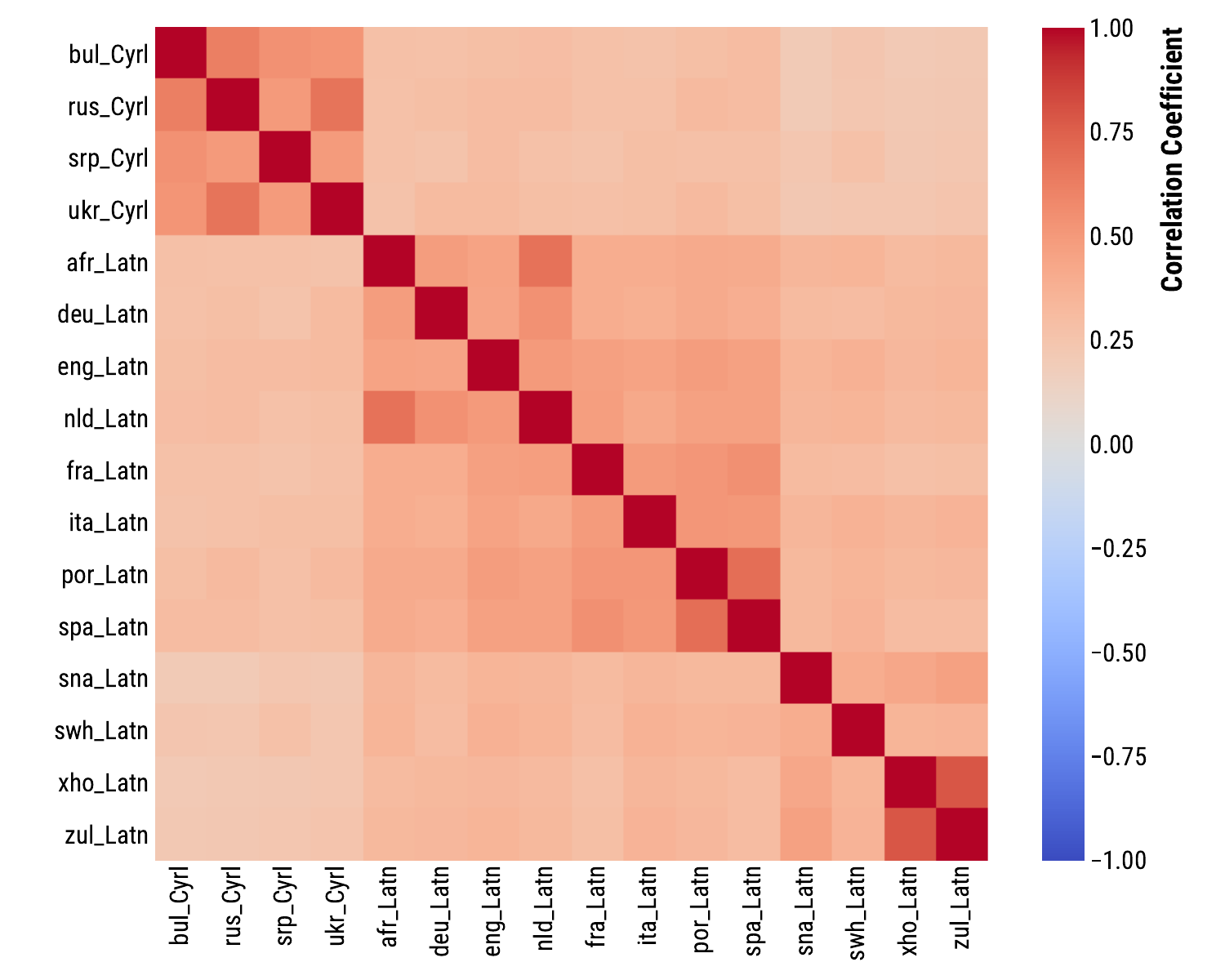}
        \caption{Step \num{8192}.}
        \label{subfig:neuron_8192_app}
    \end{subfigure}%
    ~
    \begin{subfigure}[t]{0.33\textwidth}
        \centering
        \includegraphics[width=\linewidth,trim=1cm 0cm 0cm 0cm,clip]{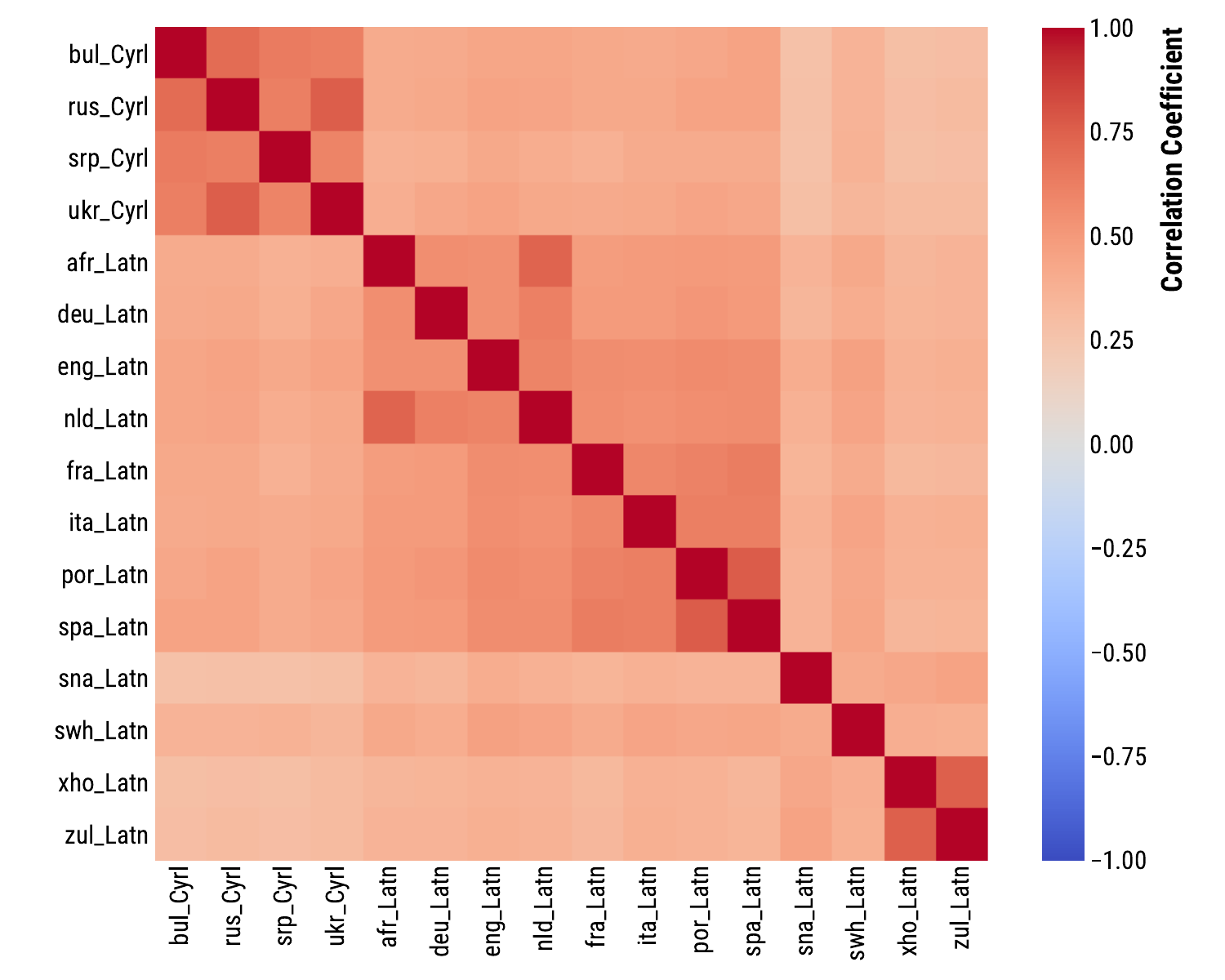}
        \caption{Step \num{16384}.}
        \label{subfig:neuron_16384_app}
    \end{subfigure}
    ~
    \begin{subfigure}[t]{0.33\textwidth}
        \centering
        \includegraphics[width=\linewidth,trim=1cm 0cm 0cm 0cm,clip]{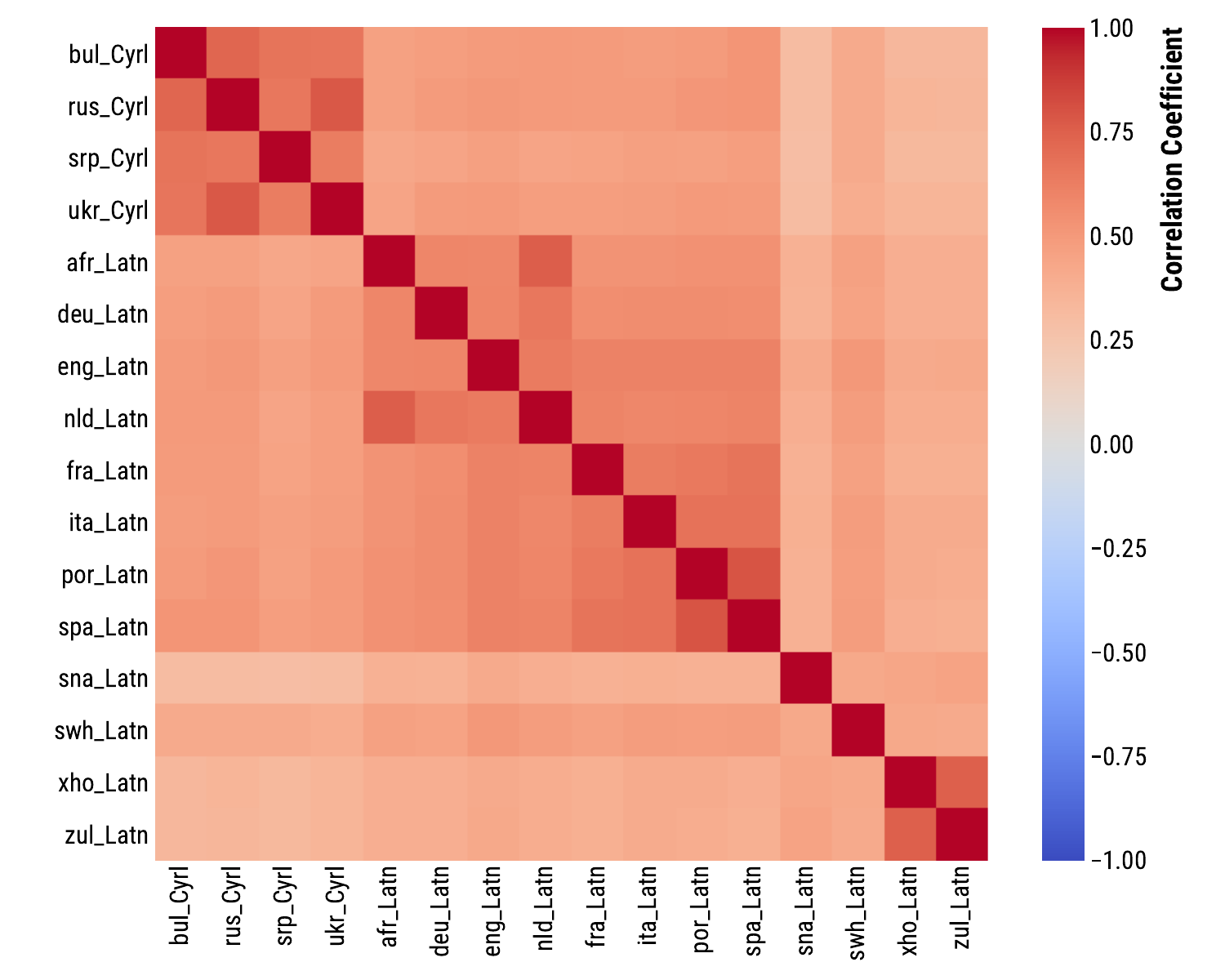}
        \caption{Step \num{32768}.}
        \label{subfig:neuron_32768_app}
    \end{subfigure}%
    ~
    \begin{subfigure}[t]{0.33\textwidth}
        \centering
        \includegraphics[width=\linewidth,trim=1cm 0cm 0cm 0cm,clip]{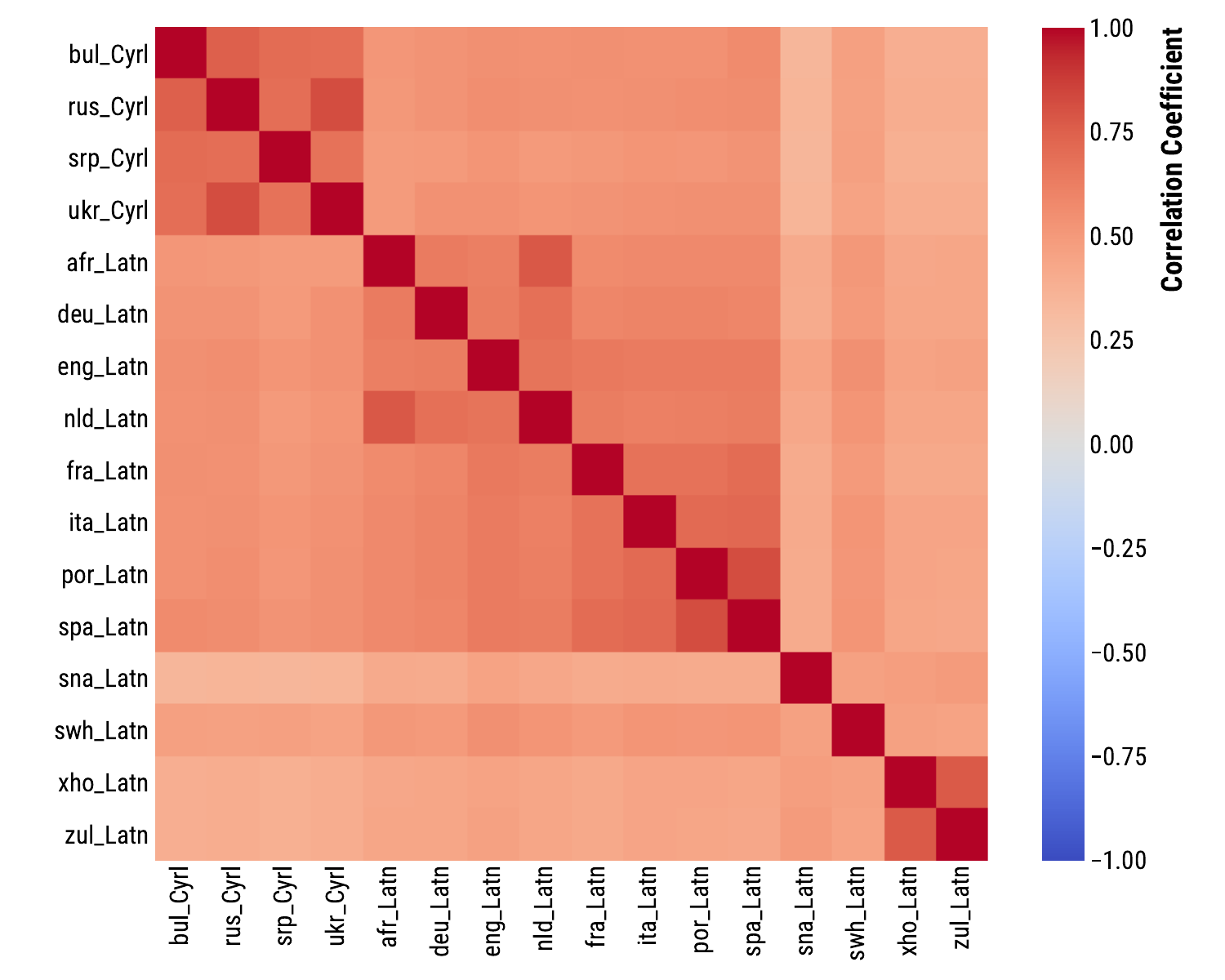}
        \caption{Step \num{65536}.}
        \label{subfig:neuron_65536_app}
    \end{subfigure}%
    ~
    \begin{subfigure}[t]{0.33\textwidth}
        \centering
        \includegraphics[width=\linewidth,trim=1cm 0cm 0cm 0cm,clip]{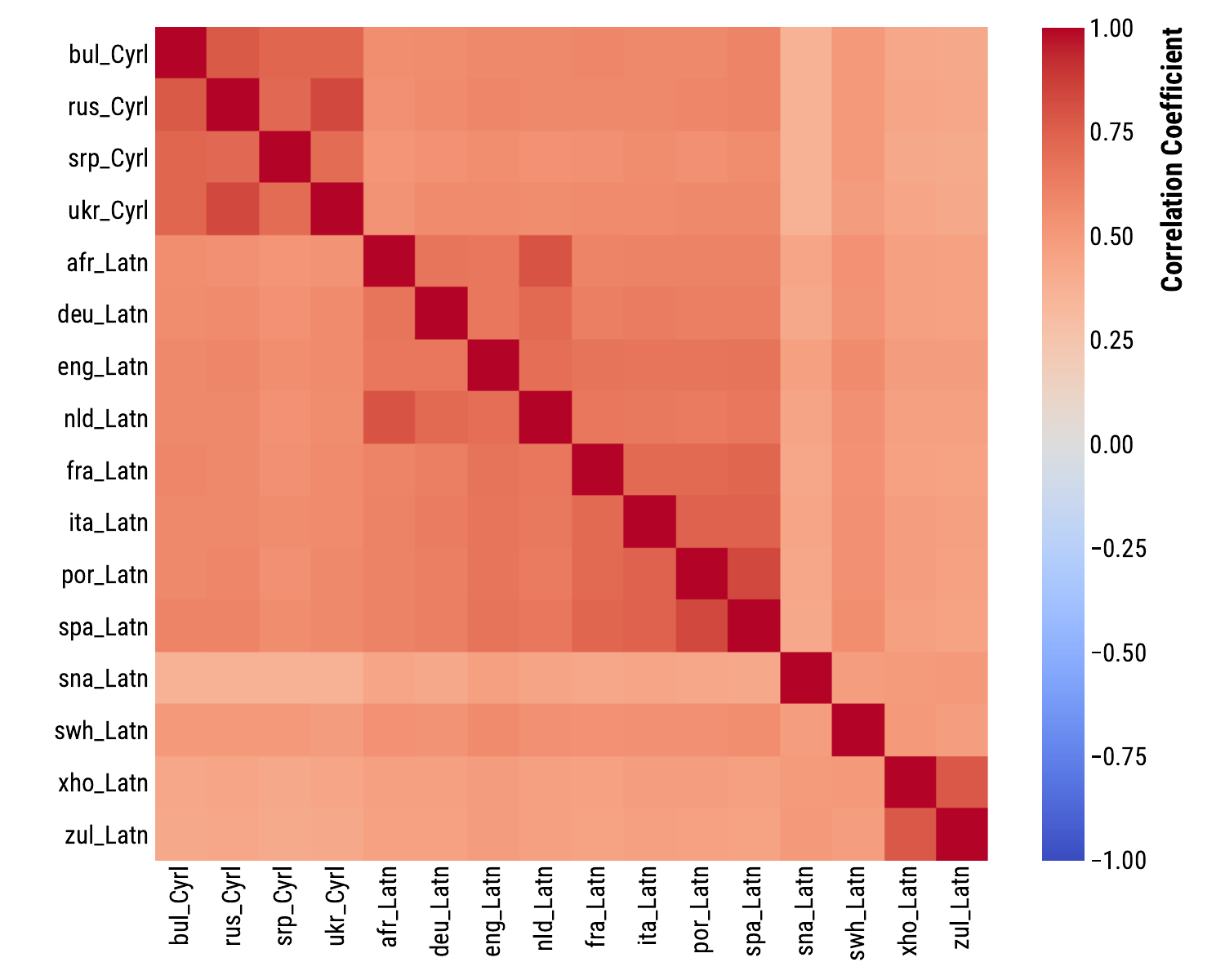}
        \caption{Step \num{131072}.}
        \label{subfig:neuron_131072_app}
    \end{subfigure}%
    \caption{Expert neuron alignment of our toy model at different training stages.}
    \label{fig:neuron_alignment_toy_full}
\end{figure*}

Alignment matrices for the toy model are displayed in \Cref{fig:neuron_alignment_toy_full}. Early in training, a division emerges between the four Cyrillic-script languages and those using the Latin script. By step 512, while this script-based division strengthens, language families develop distinct internal alignments, visible as red squares in the matrix. Notably, script alone does not determine alignment patterns: Germanic and Romance languages (middle of matrix) show stronger mutual alignment than either does with the Bantoid languages, although all use the Latin script.

\begin{figure}
    \centering
    \includegraphics[width=\linewidth,trim=1.25cm 0cm 1.25cm .6cm,clip]{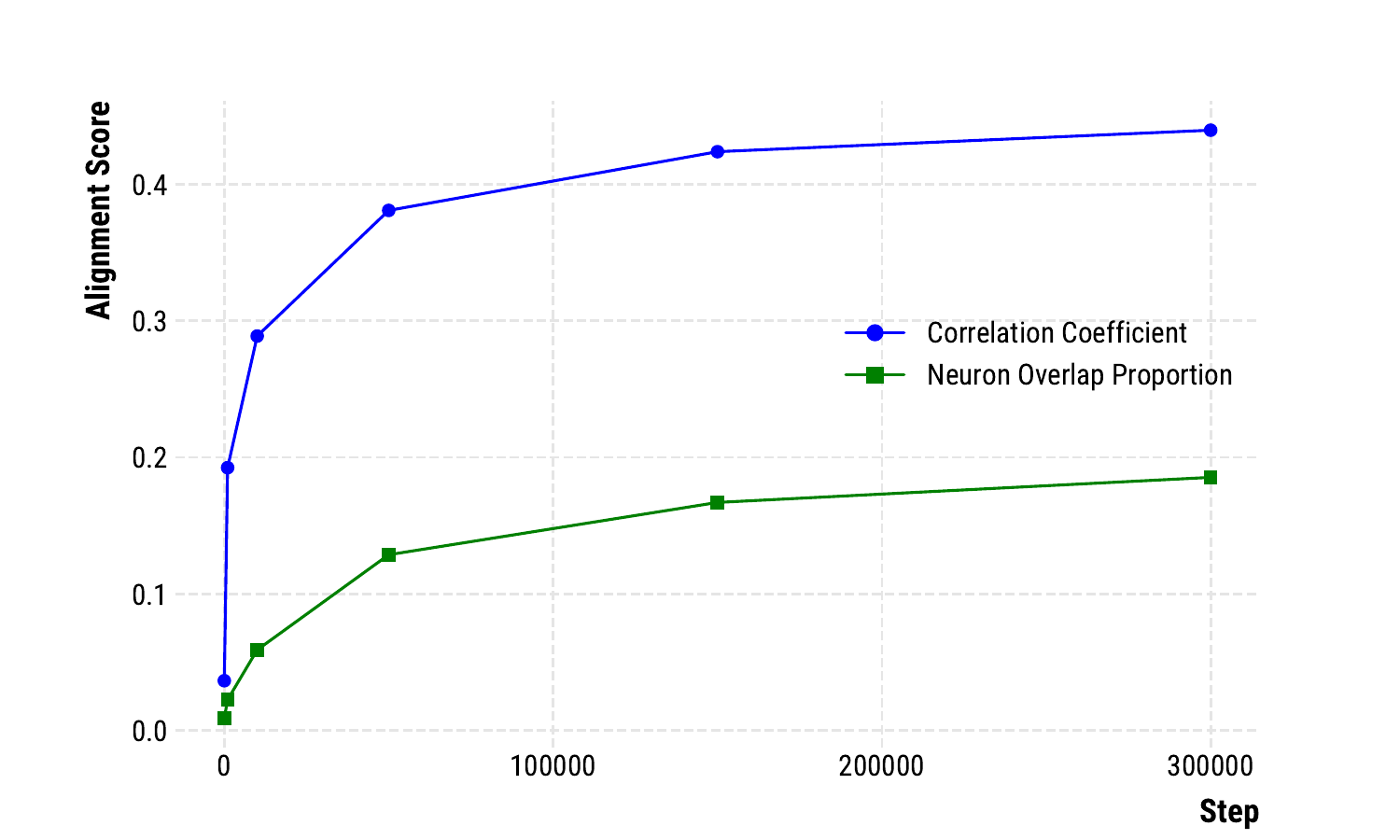} 
    \caption{Correlation Coefficient and Neuron Overlap Proportion of top 500 neurons throughout training, averaged across concepts and languages for \bloomb.}
    \label{fig:neuron_complete_training_big}
\end{figure}

\begin{figure}
    \centering
    \includegraphics[width=\linewidth]{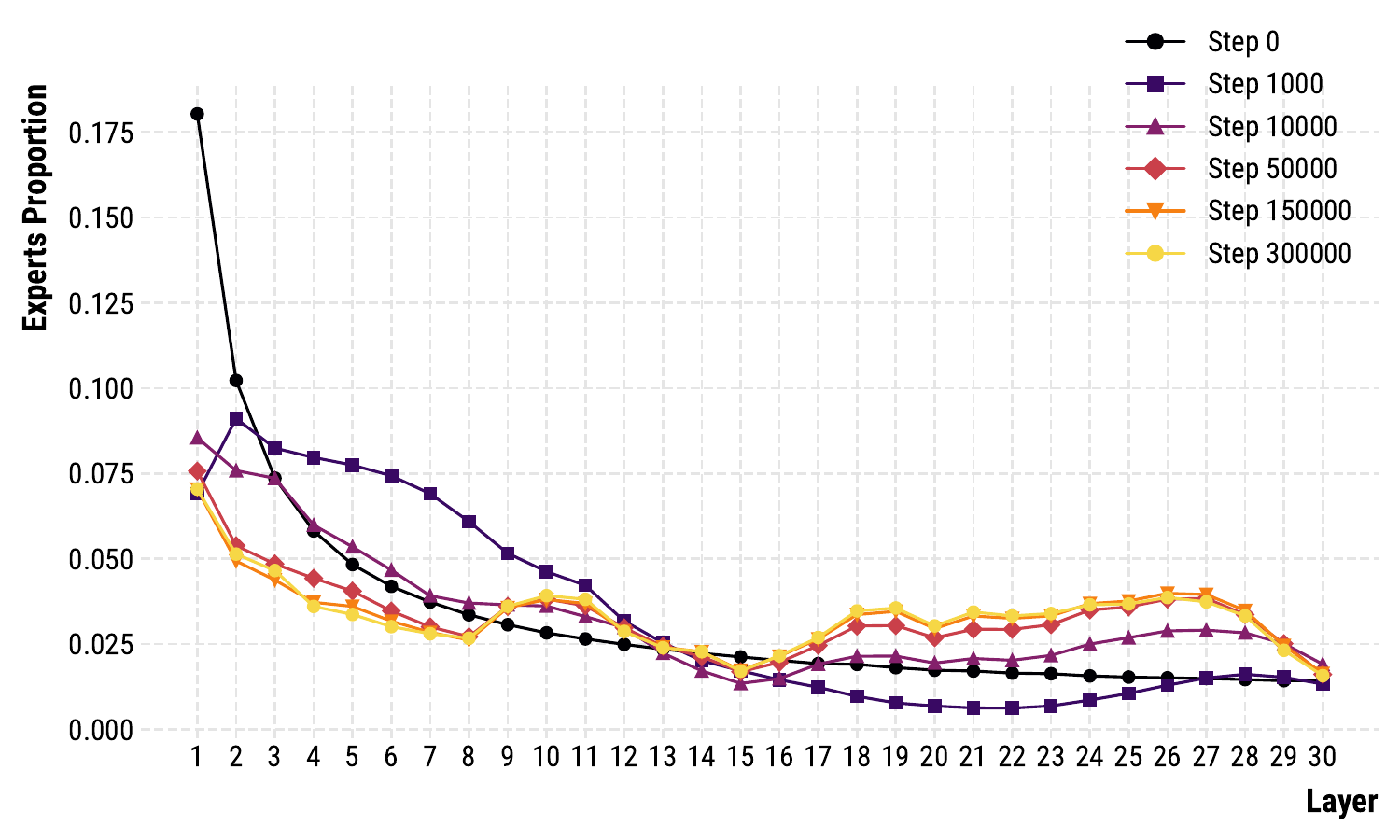}
    \caption{Layer-wise distribution of \textsc{BLOOM-7b1}'s top 500 expert neurons, averaged across languages and concepts.}
    \label{fig:layer_distribution_comparison_big}
\end{figure}

\begin{figure}
    \centering
    \includegraphics[width=\linewidth]{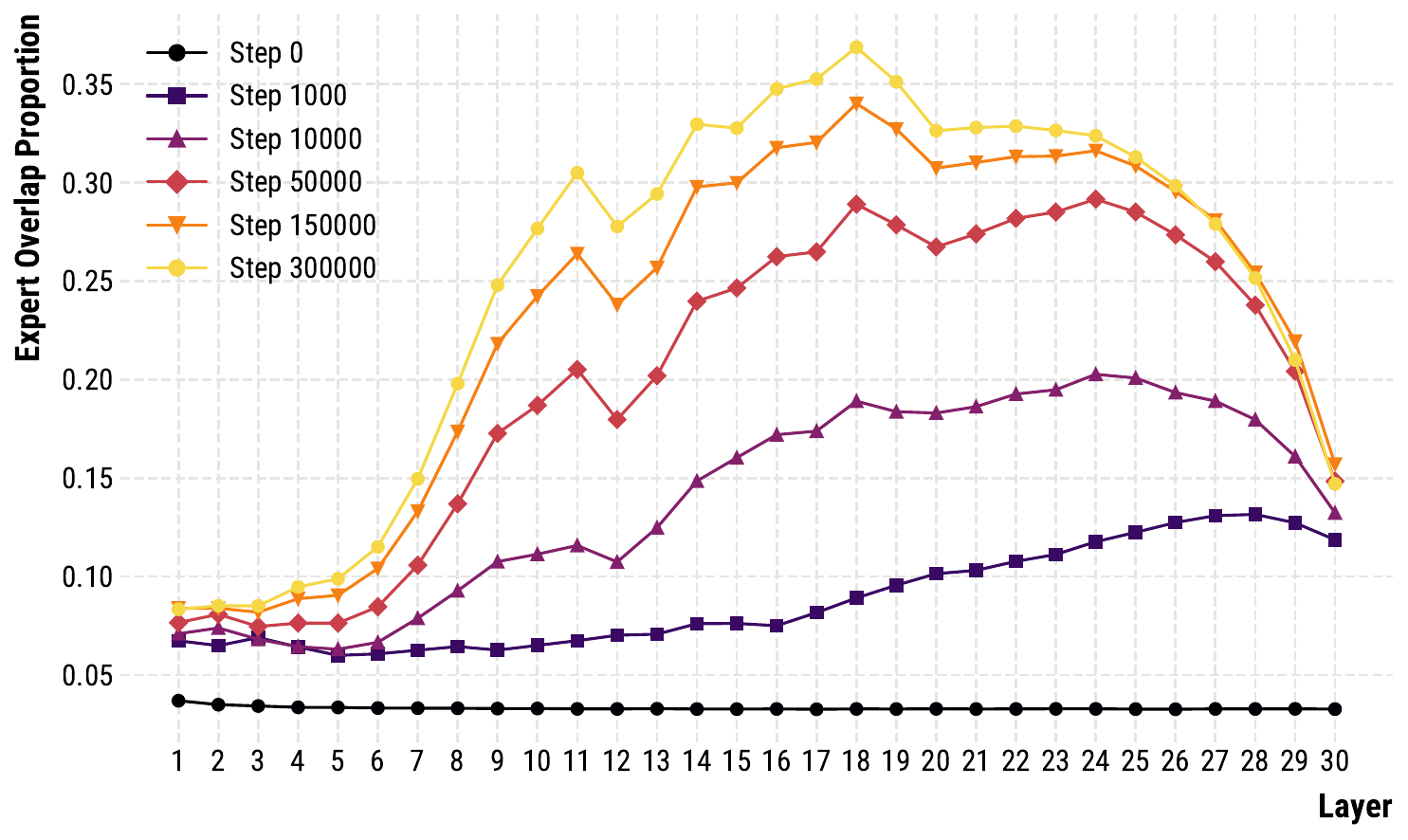}
    \caption{Cross-lingual overlap of \textsc{BLOOM-7b1}'s top 500 expert neurons per layer, showing the averaged proportion of shared neurons between language pairs.}
    \label{fig:semantics-layers_big}
\end{figure}

Beyond these detailed alignment analyses, we confirm that the alignment behavior during training (\Cref{fig:neuron_complete_training}) as well as the layer distributions (\Cref{fig:layer_distribution_comparison,fig:semantics-layers}) are consistently replicated in the larger \bloomb model (\Cref{fig:neuron_complete_training_big,fig:layer_distribution_comparison_big,fig:semantics-layers_big}).

\section{Text Generation Experiments}
\label{app:generation}

Following \citet{kojima-etal-2024-multilingual}, we generate 100 sentences per concept using different random seeds, with nucleus sampling (\(p=0.9\)), temperature (\(t=0.8\)), and a maximum sequence length of 64. For generation, we only manipulate the top 500 expert neurons by setting them to their concept-specific median values, and provide the model with the \texttt{</s>} token. 

\begin{figure*}[t!]
    \centering
    \begin{subfigure}[t]{0.45\textwidth}
        \centering
        \includegraphics[width=\linewidth,trim=0cm 0cm 0cm 0cm,clip]{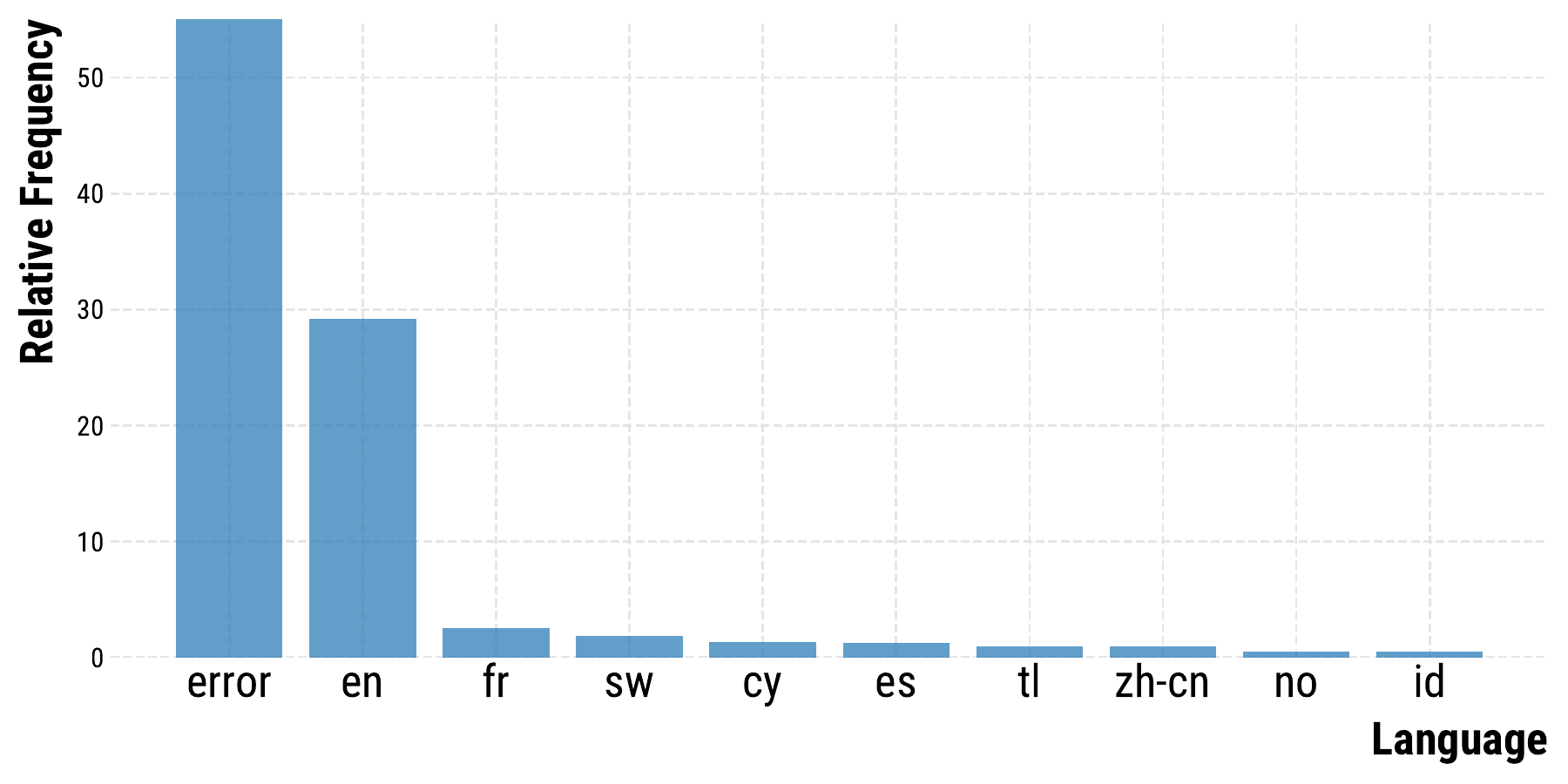}
        \caption{Step \num{1000}. The model generates incoherent text with excessive punctuation, which \langdetect cannot classify.}
    \end{subfigure}%
    ~ 
    \begin{subfigure}[t]{0.45\textwidth}
        \centering
        \includegraphics[width=\linewidth,trim=0cm 0cm 0cm 0cm,clip]{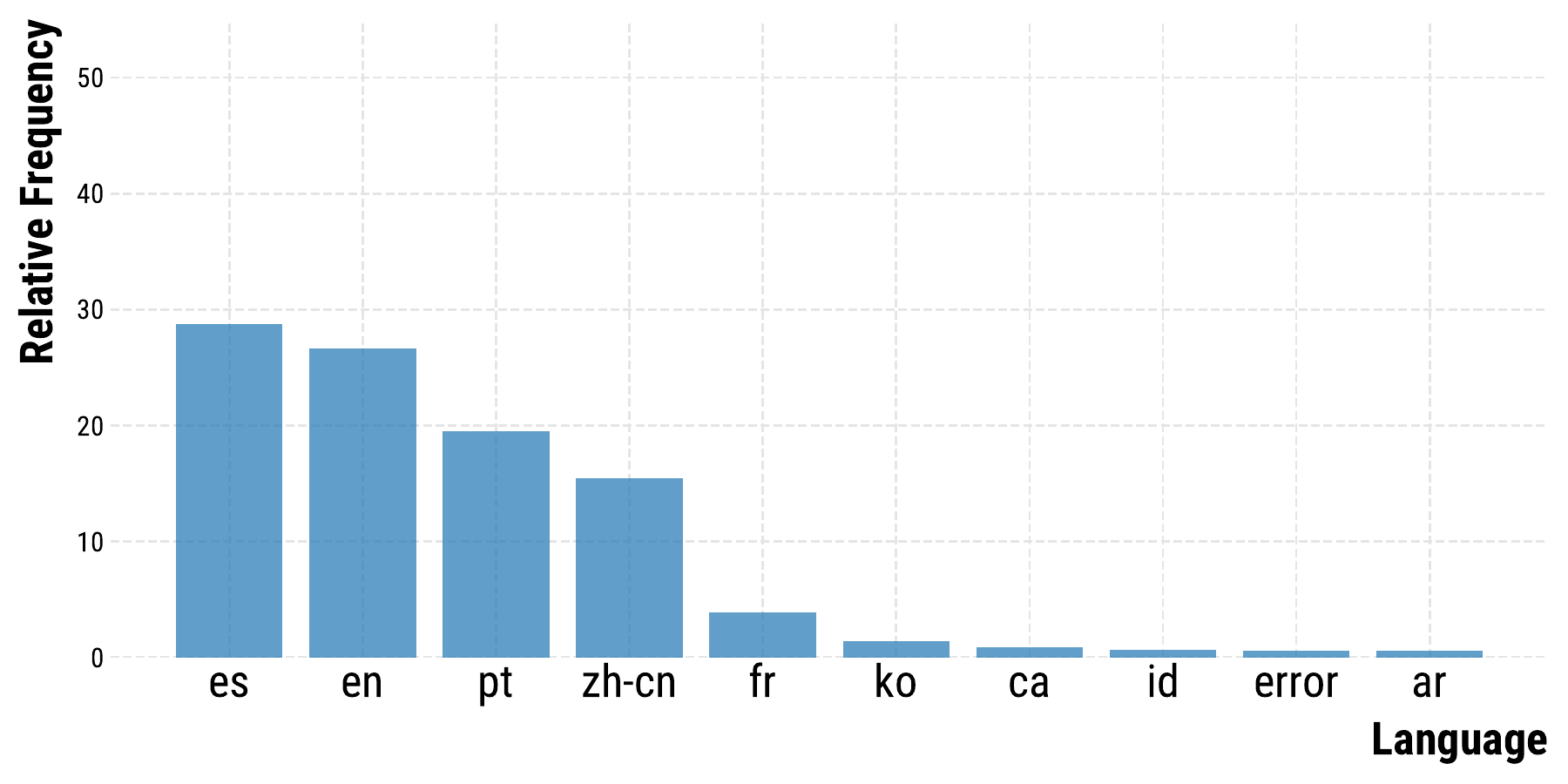}
        \caption{Step \num{10000}. The model primarily generates Spanish text, but we also observe a substantial presence of English, Portuguese, and Chinese  generations.}
    \end{subfigure}
    ~
    \begin{subfigure}[t]{0.45\textwidth}
        \centering
        \includegraphics[width=\linewidth,trim=0cm 0cm 0cm 0cm,clip]{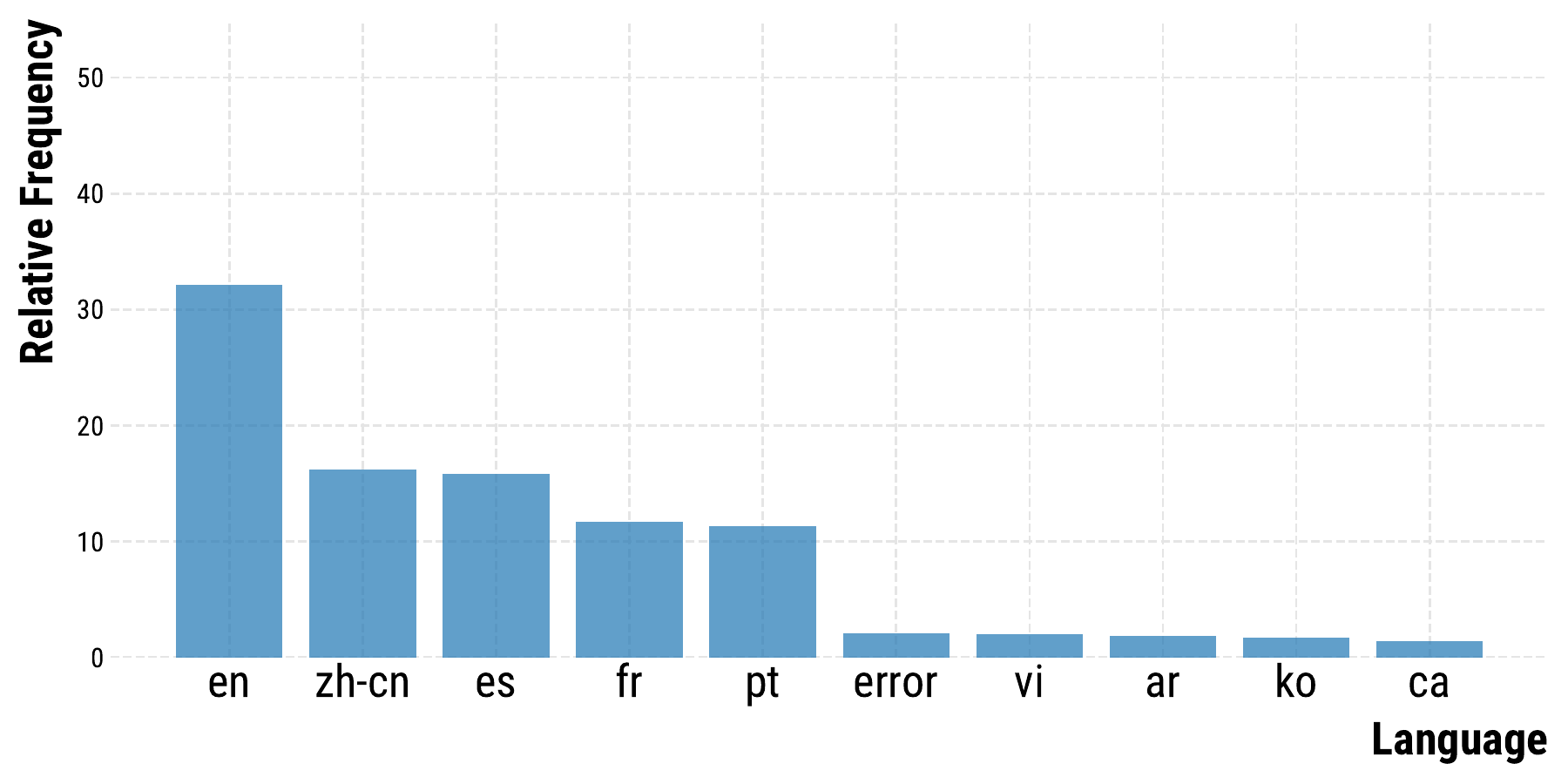}
        \caption{Step \num{100000}. The model now generates primarily English text, followed by Chinese and Spanish. Importantly, English and Chinese are the two most common languages in \textsc{BLOOM-560m}'s pre-training corpus.}
    \end{subfigure}%
    ~
    \begin{subfigure}[t]{0.45\textwidth}
        \centering
        \includegraphics[width=\linewidth,trim=0cm 0cm 0cm 0cm,clip]{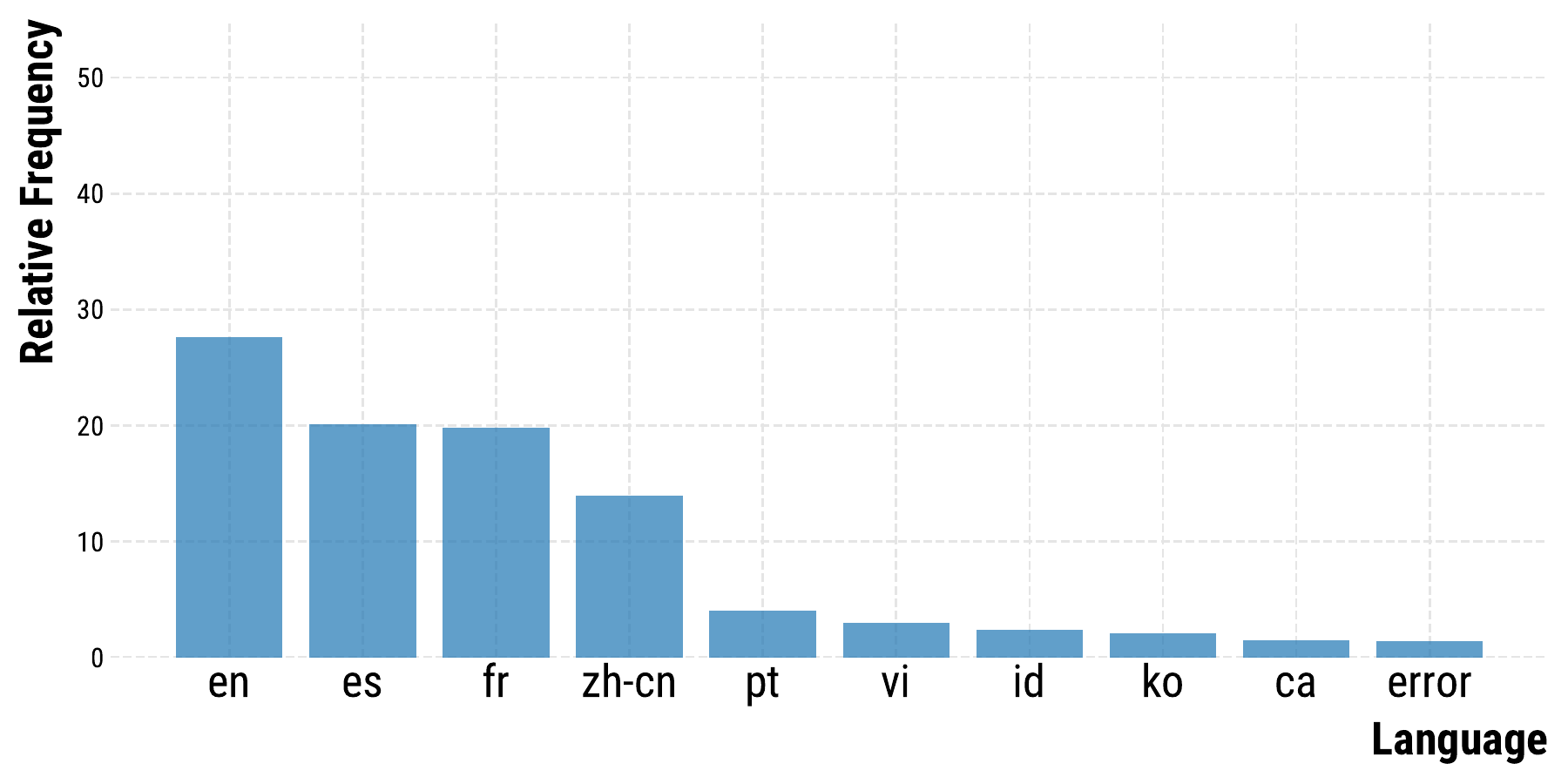}
        \caption{Step \num{200000}. French, a Romance language like Spanish, becomes more prominent. As with Chinese, its prominence reflects its status as a high-resource language in \textsc{BLOOM-560m}'s pre-training corpus.}
    \end{subfigure}
    ~
    \begin{subfigure}[t]{0.45\textwidth}
        \centering
        \includegraphics[width=\linewidth,trim=0cm 0cm 0cm 0cm,clip]{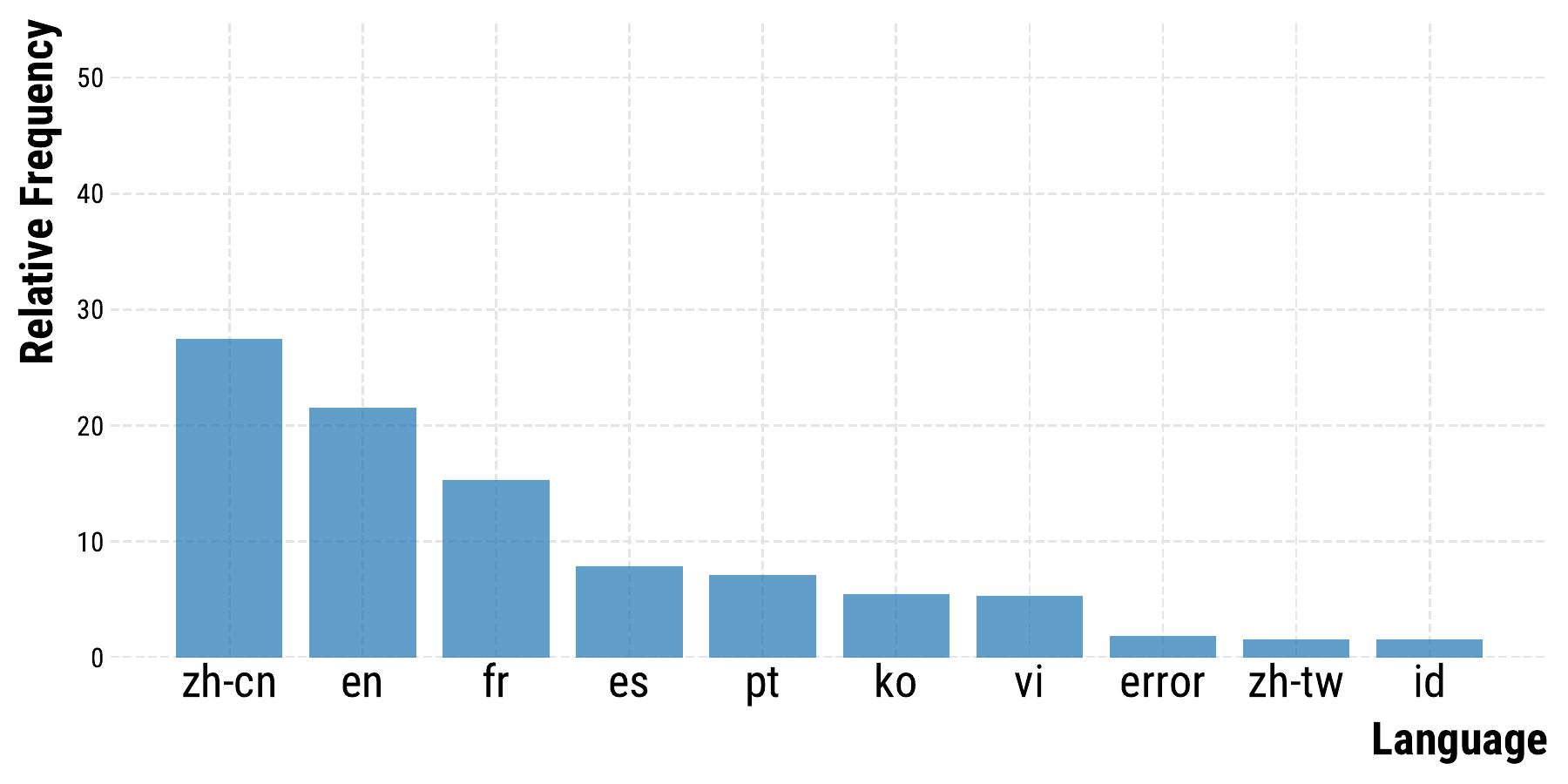}
        \caption{Step \num{300000}. The three high-resource languages--Chinese, English, and French--dominate, while Spanish becomes increasingly less present.}
    \end{subfigure}%
    ~
    \begin{subfigure}[t]{0.45\textwidth}
        \centering
        \includegraphics[width=\linewidth,trim=0cm 0cm 0cm 0cm,clip]{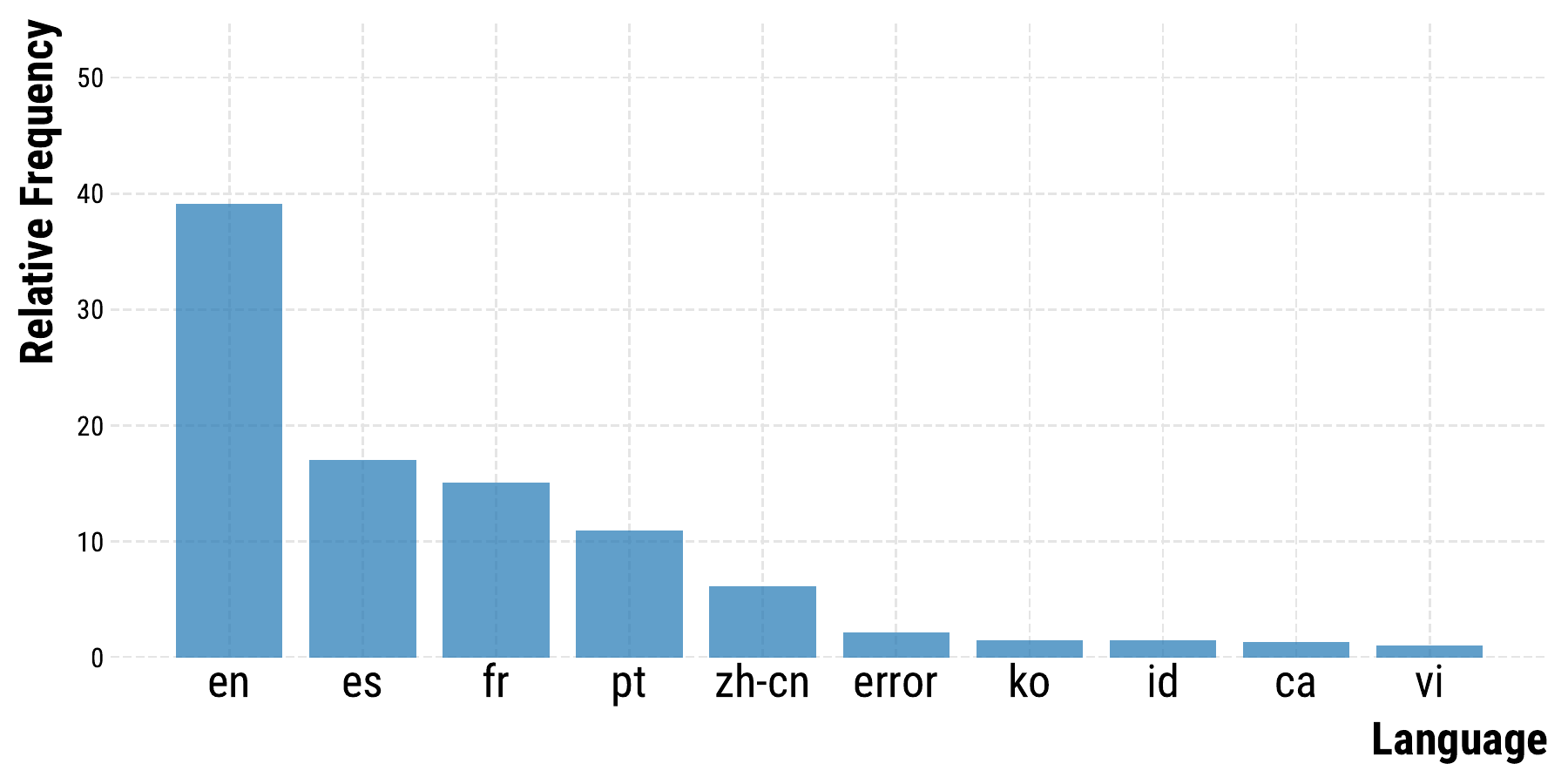}
        \caption{Step \num{400000}. The model generates a large amount of English text, with Spanish appearing to a much lesser extent.}
    \end{subfigure}
    \caption{Text generation experiments: Relative frequency distribution of the top 10 detected languages when manipulating neurons derived from Spanish data across all available \bloomm checkpoints, as classified using \langdetect.}
    \label{fig:langdetect-detailed-spanish-full}
\end{figure*}

Example generations for the senses \texttt{earthquake-1\_11\_00} and \texttt{joy-1\_12\_00}, derived from Spanish and Simplified Chinese data, are shown in \Cref{tab:generation}. To quantify the phenomenon of later \bloomm checkpoints favoring high-resource languages, we show the detailed development for Spanish in \Cref{fig:langdetect-detailed-spanish-full}. Here, we see that the model first creates language-specific concepts, generating text in Spanish, but in later checkpoints favors English. Alongside English, other high-resource languages such as Chinese and French \enquote{compete for dominance} as well (step \num{300000}).

\begin{figure*}[t!]
    \centering
    \begin{subfigure}[t]{0.45\textwidth}
        \centering
        \includegraphics[width=\linewidth,trim=0cm 0cm 0cm 0cm,clip]{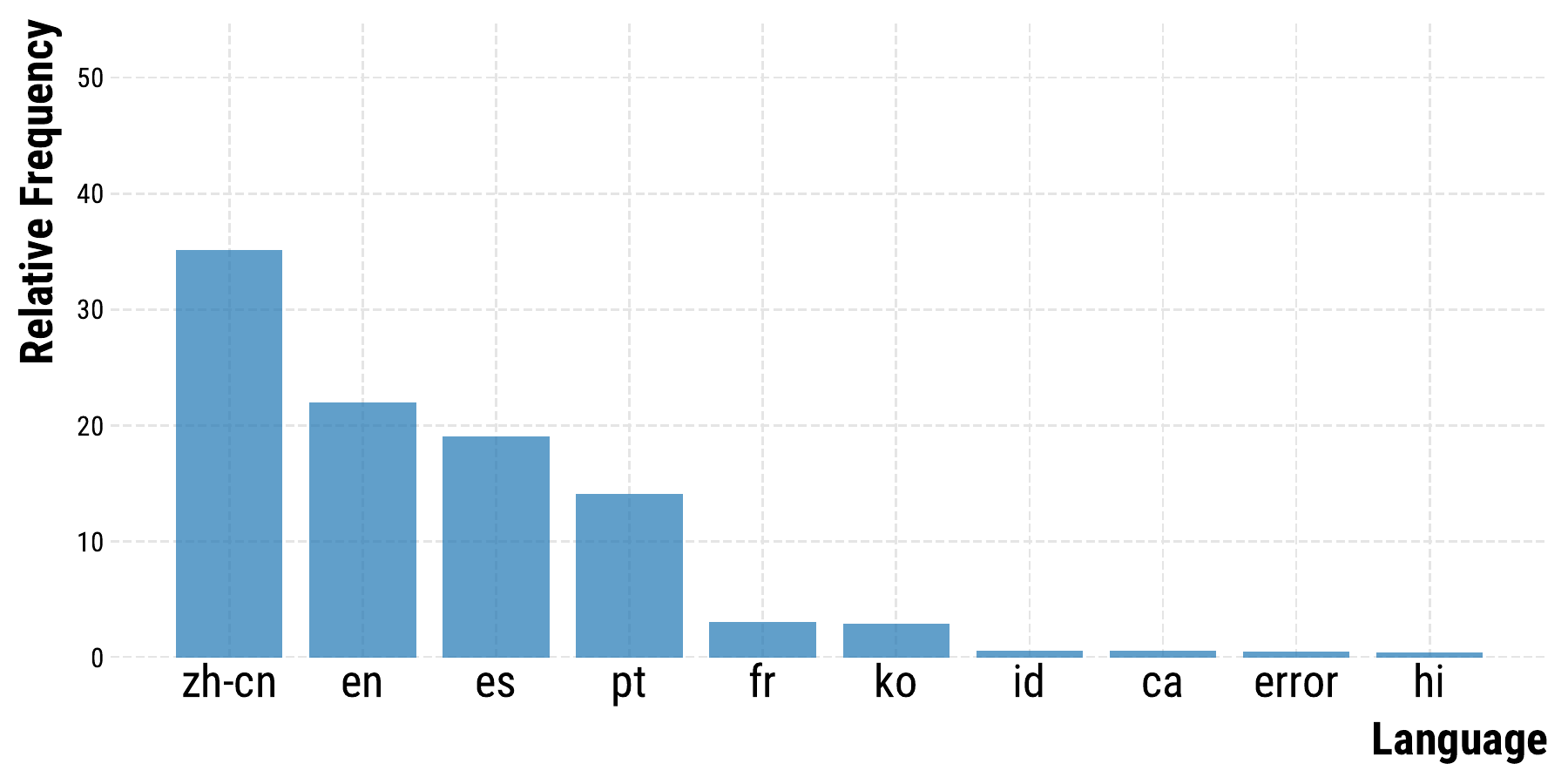}
        \caption{Early training stage (step \num{10000}).}
    \end{subfigure}%
    ~ 
    \begin{subfigure}[t]{0.45\textwidth}
        \centering
        \includegraphics[width=\linewidth,trim=0cm 0cm 0cm 0cm,clip]{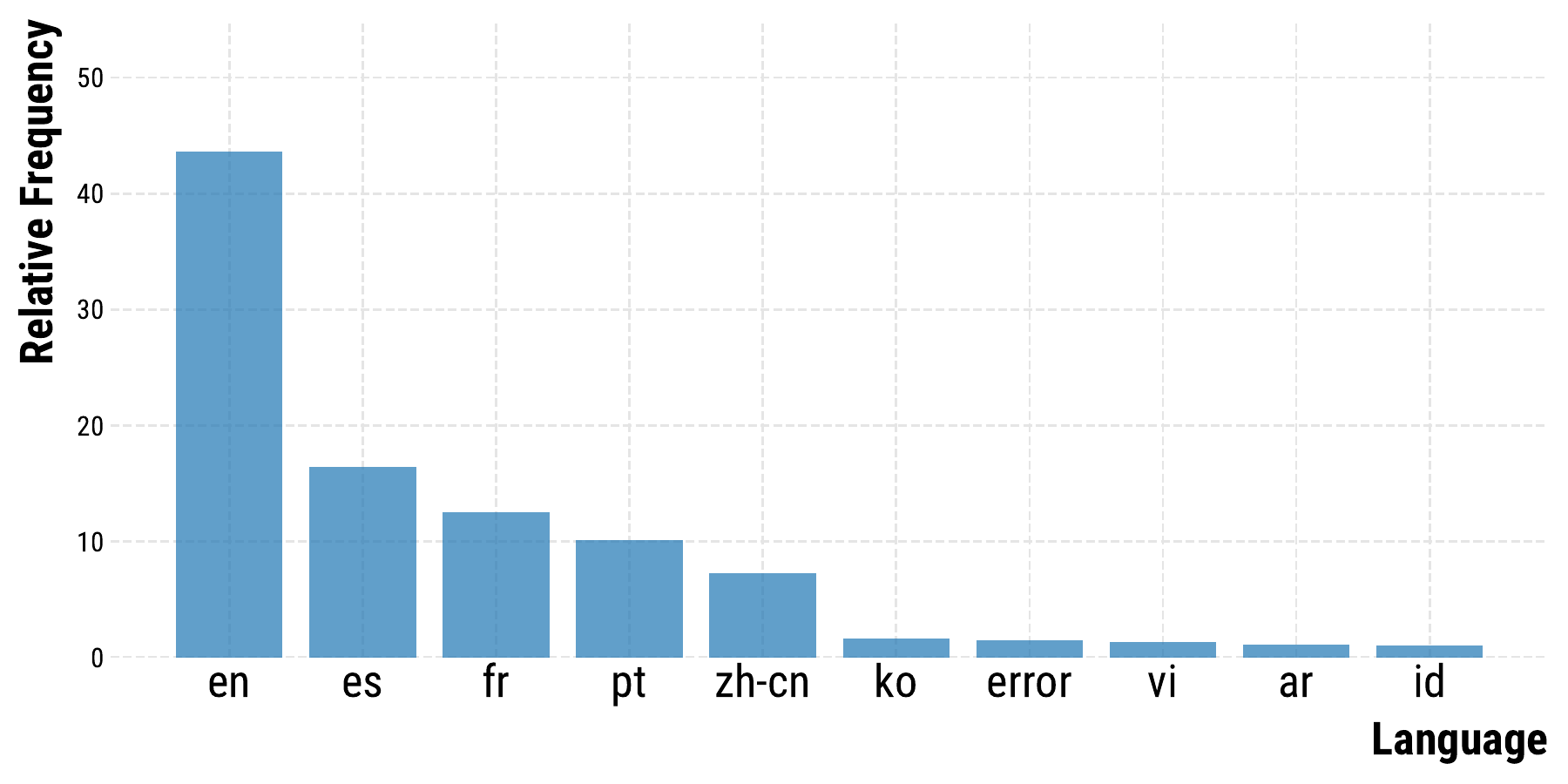}
        \caption{Late training stage (step \num{400000}).}
    \end{subfigure}
    \caption{Text generation experiments: Relative frequency distribution of the top 10 detected languages when manipulating neurons derived from Chinese data, as classified using \langdetect.}
    \label{fig:langdetect-detailed-chinese}
\end{figure*}

\begin{figure*}[t!]
    \centering
    \begin{subfigure}[t]{0.45\textwidth}
        \centering
        \includegraphics[width=\linewidth,trim=0cm 0cm 0cm 0cm,clip]{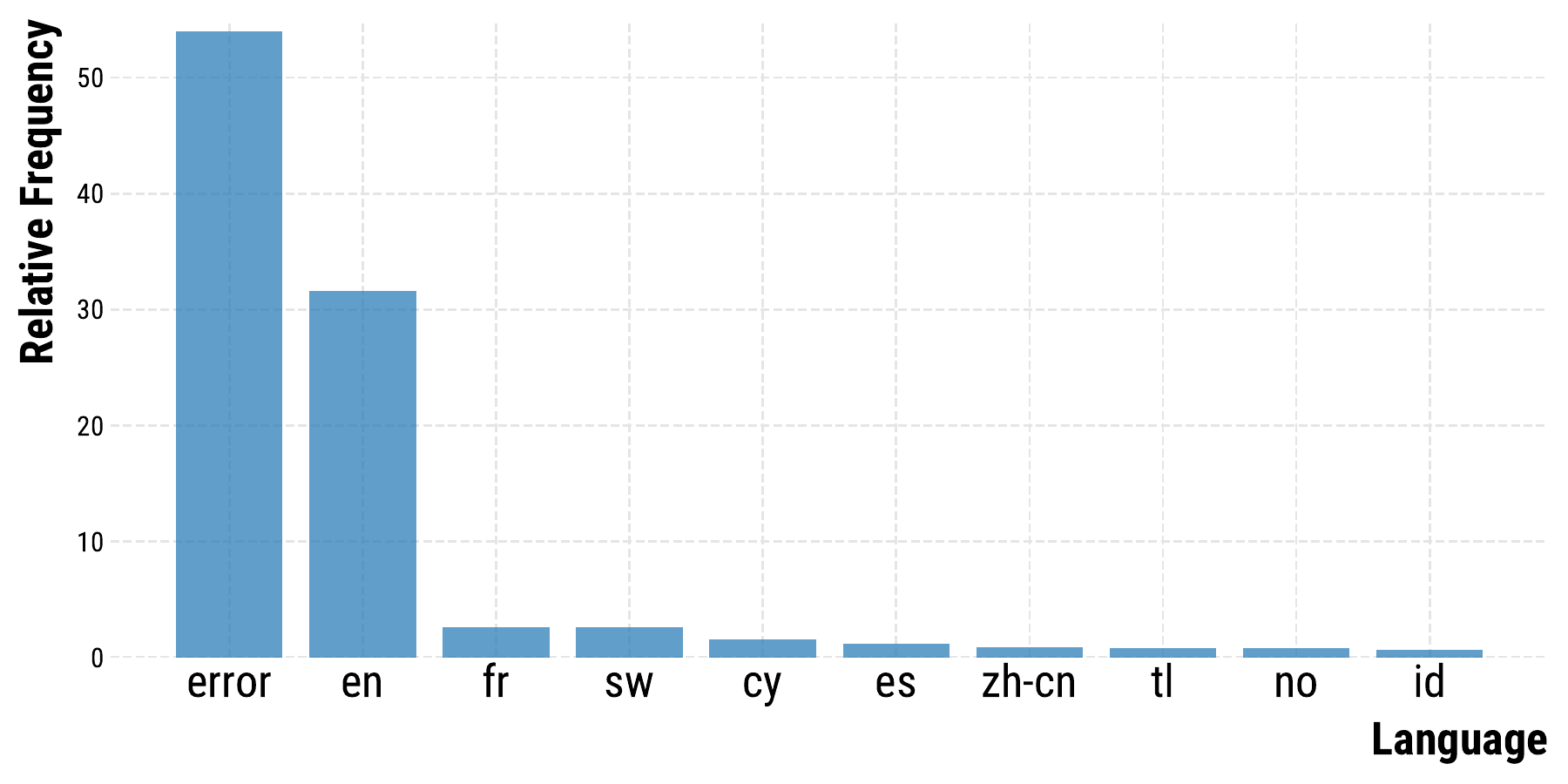}
        \caption{Step \num{1000}.}
    \end{subfigure}%
    ~ 
    \begin{subfigure}[t]{0.45\textwidth}
        \centering
        \includegraphics[width=\linewidth,trim=0cm 0cm 0cm 0cm,clip]{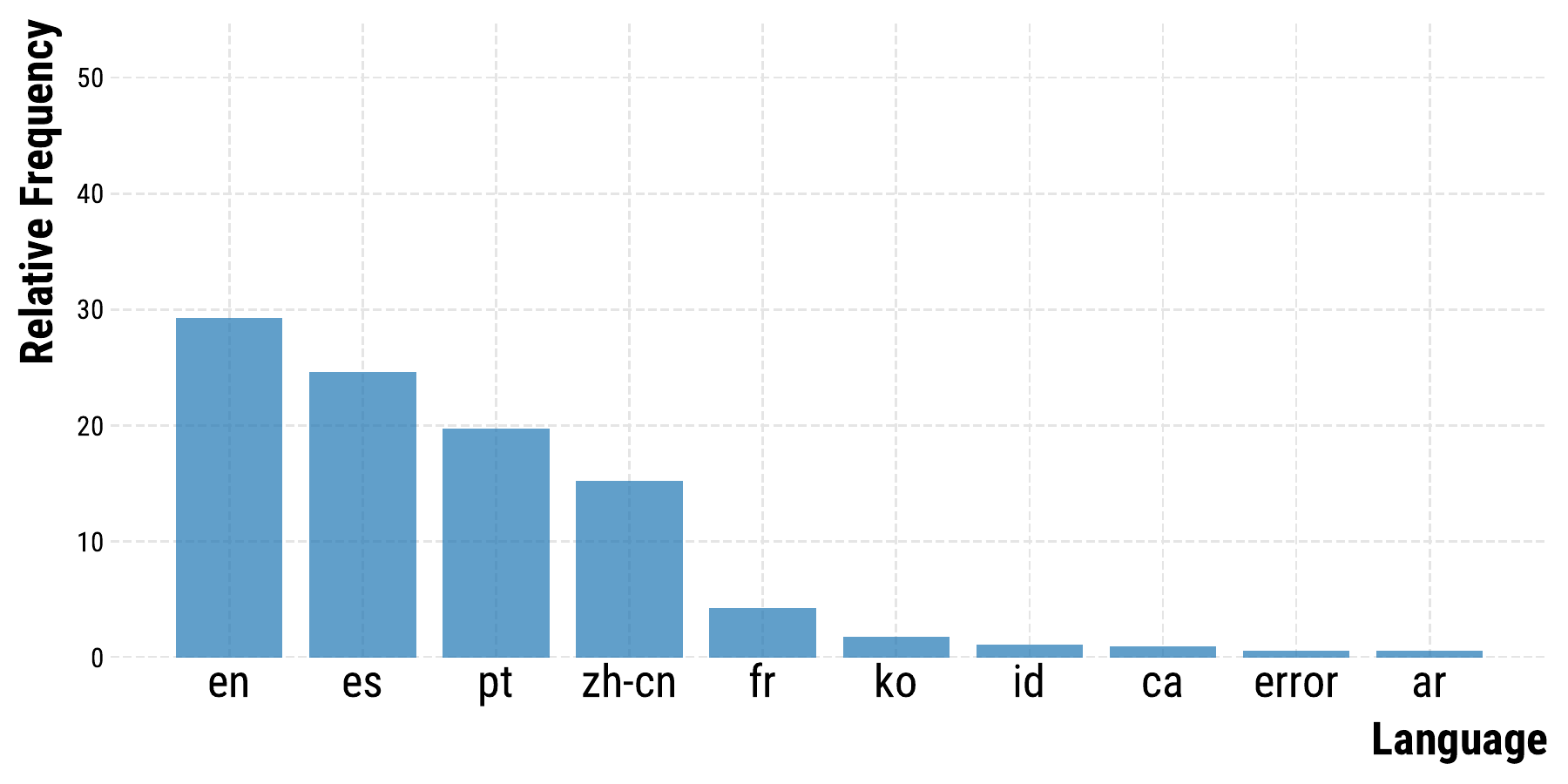}
        \caption{Step \num{10000}. The model already generates text in high-resource languages.}
    \end{subfigure}
    ~
    \begin{subfigure}[t]{0.45\textwidth}
        \centering
        \includegraphics[width=\linewidth,trim=0cm 0cm 0cm 0cm,clip]{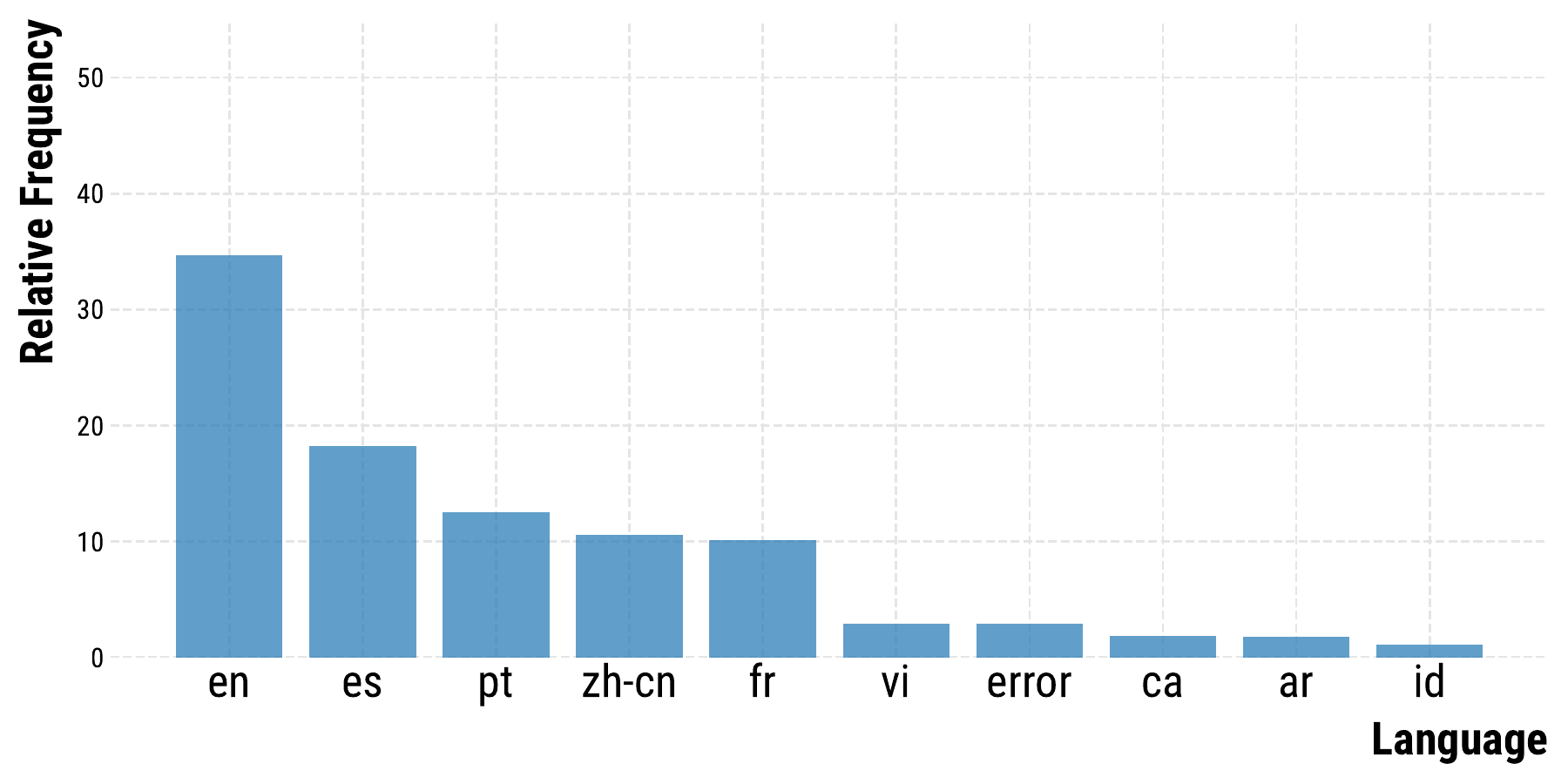}
        \caption{Step \num{100000}. }
    \end{subfigure}%
    ~
    \begin{subfigure}[t]{0.45\textwidth}
        \centering
        \includegraphics[width=\linewidth,trim=0cm 0cm 0cm 0cm,clip]{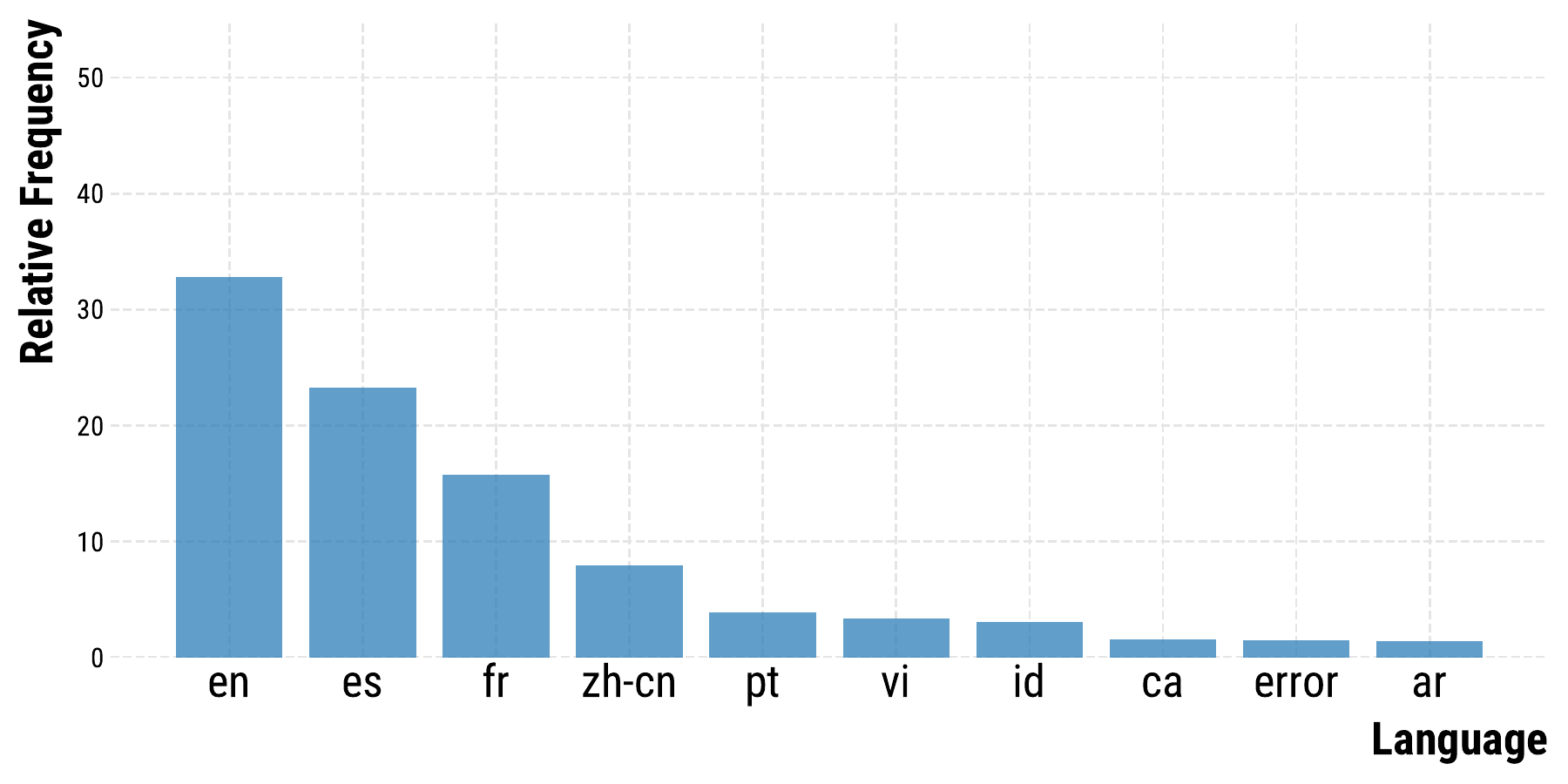}
        \caption{Step \num{200000}.}
    \end{subfigure}
    ~
    \begin{subfigure}[t]{0.45\textwidth}
        \centering
        \includegraphics[width=\linewidth,trim=0cm 0cm 0cm 0cm,clip]{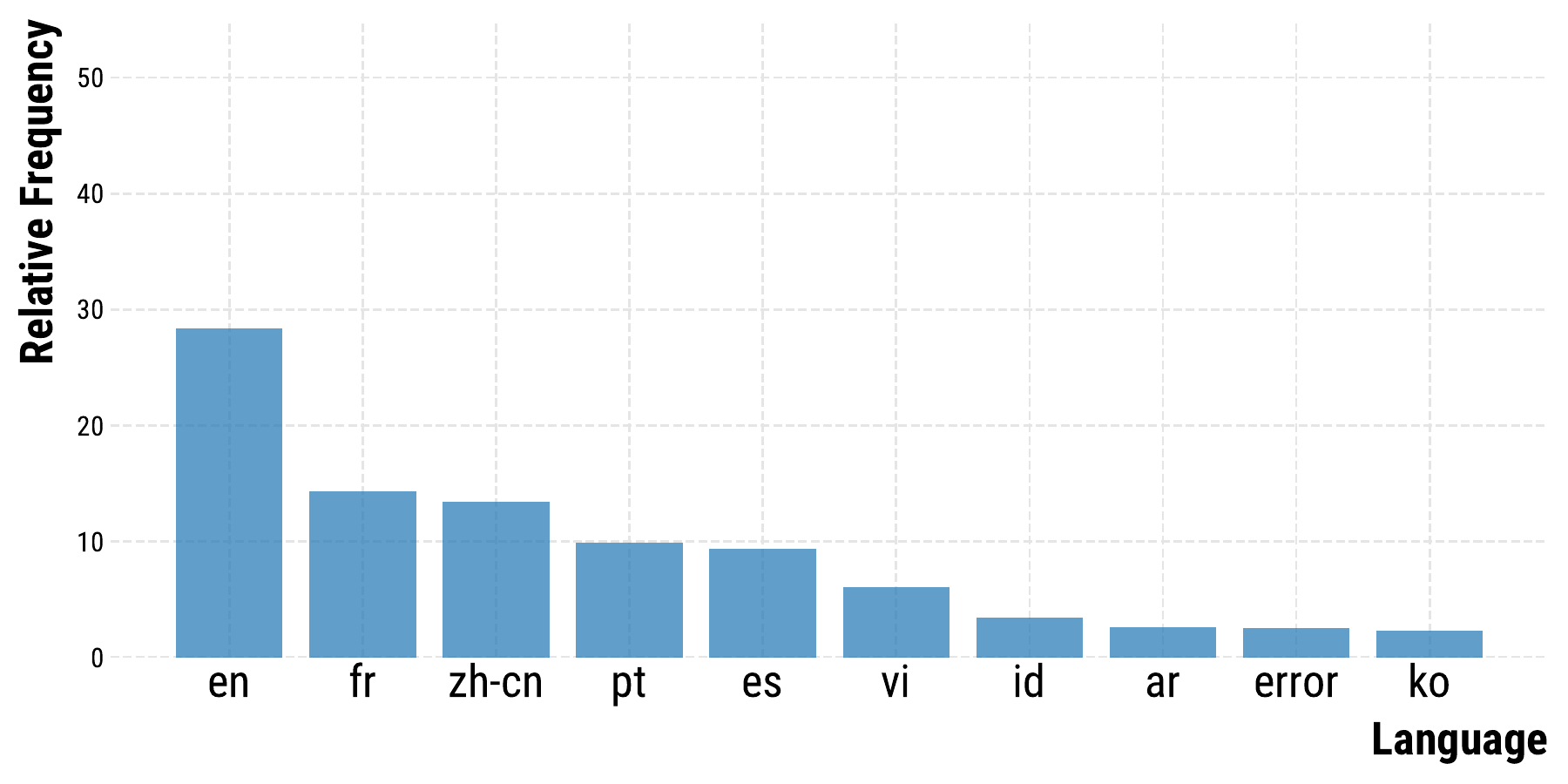}
        \caption{Step \num{300000}.}
    \end{subfigure}%
    ~
    \begin{subfigure}[t]{0.45\textwidth}
        \centering
        \includegraphics[width=\linewidth,trim=0cm 0cm 0cm 0cm,clip]{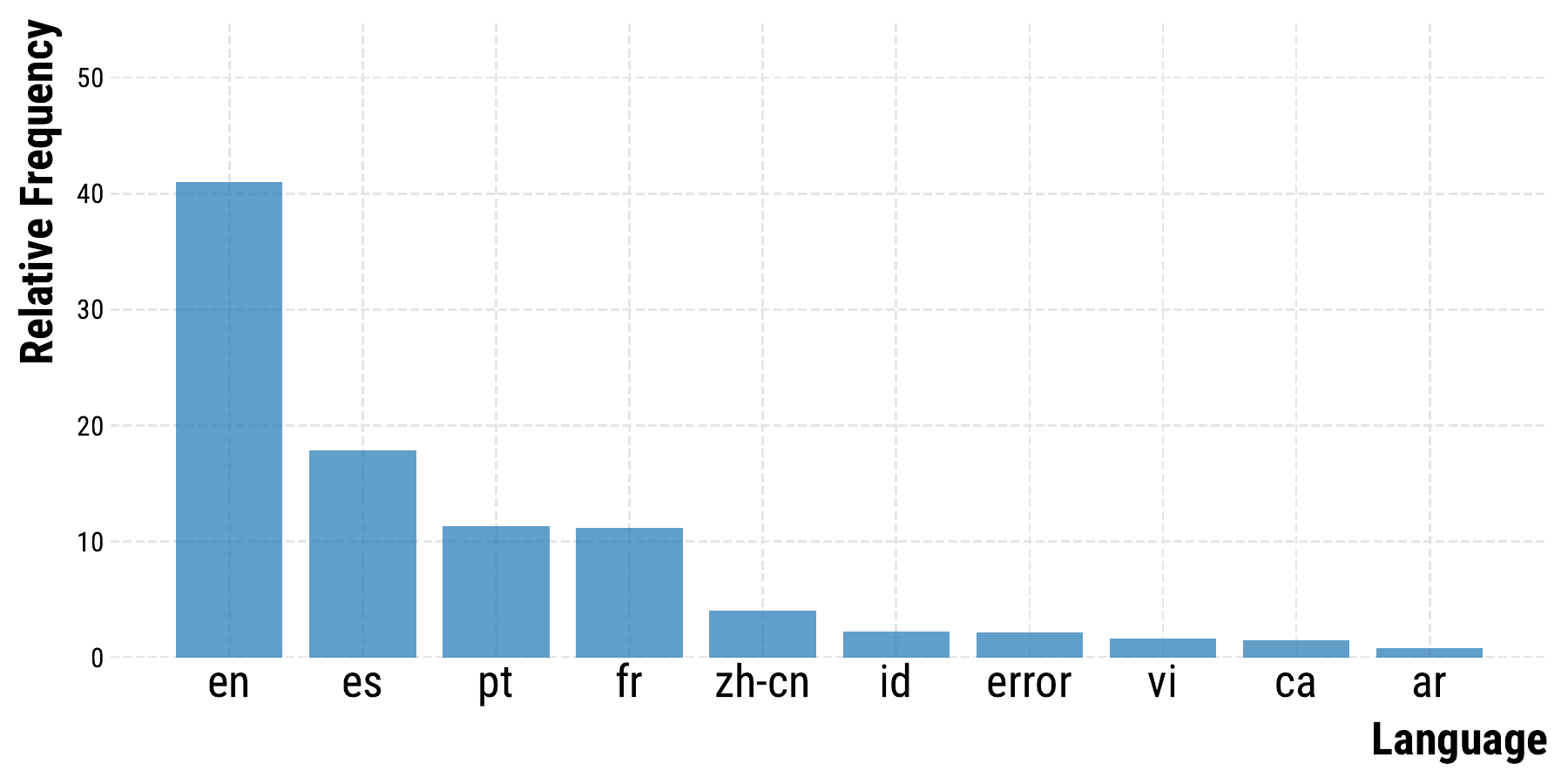}
        \caption{Step \num{400000}. }
    \end{subfigure}
    \caption{Text generation experiments: Relative frequency distribution of the top 10 detected languages when manipulating neurons derived from Swahili data across all available \bloomm checkpoints, as classified using \langdetect.}
    \label{fig:langdetect-detailed-swahili-full}
\end{figure*}

We demonstrate the same trend for Chinese in \Cref{fig:langdetect-detailed-chinese}. For lower-resourced languages like Swahili, however, the pattern differs: when deriving concept-specific neurons from Swahili data, no checkpoint generates a notable amount of Swahili text (\Cref{fig:langdetect-detailed-swahili-full}), suggesting these languages are underrepresented throughout training.

\end{document}